\documentclass[journal]{IEEEtran}
\IEEEoverridecommandlockouts

\usepackage{cite}
\usepackage{amsmath,amssymb,amsfonts}
\usepackage{algorithmic}
\usepackage{graphicx}
\usepackage{textcomp}
\usepackage{xcolor}
\usepackage{multirow}
\usepackage{booktabs}
\usepackage{array}
\usepackage{tabularx}
\usepackage{hyperref}
\usepackage{balance}
\usepackage{float}
\usepackage{enumitem}
\usepackage{pifont}
\usepackage{colortbl}
\usepackage{tikz}
\usepackage{adjustbox}
\usetikzlibrary{shapes.geometric, arrows, positioning, calc, mindmap, trees}
\usepackage{pgfplots}
\usepackage{orcidlink}
\pgfplotsset{compat=1.17}
\usepackage{subcaption}

\def\BibTeX{{\rm B\kern-.05em{\sc i\kern-.025em b}\kern-.08em
    T\kern-.1667em\lower.7ex\hbox{E}\kern-.125emX}}

\begin{document}

\title{Cybersecurity of Teleoperated Quadruped Robots: A Systematic Survey of Vulnerabilities, Threats, and Open Defense Gaps\\
{\footnotesize \textsuperscript{}}
}

\author{
\IEEEauthorblockN{Mohammad Sabouri\,\orcidlink{0000-0002-2568-3253}}\\
\IEEEauthorblockA{
    \textit{Department of Informatics, Bioengineering,}
    \textit{Robotics and Systems Engineering (DIBRIS)}\\
    University of Genoa\\
    Genoa, Italy\\
    S5659227@studenti.unige.it}
}

\maketitle

\begin{abstract}
Teleoperated quadruped robots are increasingly deployed in 
safety-critical missions---industrial inspection, military 
reconnaissance, and emergency response---yet the security of 
communication and control infrastructure linking operators to 
remote platforms remains insufficiently characterized. Quadrupeds 
present distinct security challenges arising from dynamic stability 
constraints, gait-dependent vulnerability windows, substantial 
kinetic energy, and elevated operator cognitive load that render 
existing robotic security frameworks inadequate.

This survey synthesizes peer-reviewed literature and vulnerability 
disclosures spanning 2019--2025 to provide a comprehensive analysis 
of cybersecurity threats, consequences, and countermeasures for 
teleoperated quadruped systems. We contribute: (i) a six-layer 
attack taxonomy spanning perception manipulation, VR/AR operator 
targeting, communication disruption, control signal attacks, 
localization spoofing, and network intrusion; (ii) systematic 
attack-to-consequence mapping with explicit timing characterization; 
(iii) Technology Readiness Level classification of defenses exposing 
a critical maturity gap between field-deployed communication 
protections (TRL 7--9) and largely experimental perception and 
operator-layer defenses (TRL 3--5); (iv) comparative security 
analysis of six commercial platforms based on publicly available 
documentation and disclosed vulnerabilities; (v) pragmatic deployment guidance stratified by implementation timeline and organizational capability; and (vi) eight prioritized research gaps with stratified implementation roadmaps addressing the most critical security deficiencies.

\textbf{Limitations:} Platform security assessments rely on publicly 
available information and may not reflect proprietary implementations. 
Quantitative attack success rates are derived from cited studies 
under controlled conditions and require domain-specific validation.
\end{abstract}


\section{Introduction}

\subsection{Motivation and Problem Statement}

\begin{figure*}[htbp]
    \centering
    
    \begin{subfigure}[b]{0.30\textwidth}
        \centering
        \includegraphics[width=\textwidth]{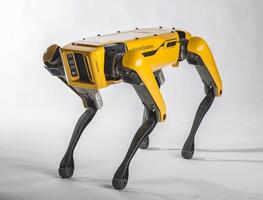}
        \caption{}
        \label{fig:baseline_a}
    \end{subfigure}
    \hfill
    \begin{subfigure}[b]{0.30\textwidth}
        \centering
        \includegraphics[width=\textwidth]{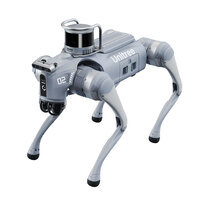}
        \caption{}
        \label{fig:baseline_b}
    \end{subfigure}
    \hfill
    \begin{subfigure}[b]{0.30\textwidth}
        \centering
        \includegraphics[width=\textwidth]{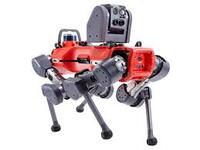}
        \caption{}
        \label{fig:baseline_c}
    \end{subfigure}
    
    \caption{Representative commercial quadruped platforms analyzed in this survey. These platforms exhibit diverse security architectures: Spot employs per-device cryptographic keying, Unitree platforms have exhibited fleet-wide credential vulnerabilities, and ANYmal targets industrial inspection with enterprise security requirements.
    (a)~Boston Dynamics Spot; 
    (b)~Unitree B2; 
    (c)~ANYbotics ANYmal.}
    \label{fig:quadruped_platforms}
\end{figure*}

Quadrupedal robots have emerged as capable platforms for industrial 
inspection, military reconnaissance, search and rescue, and 
hazardous material management \cite{boston2024security}. Platforms including Boston Dynamics Spot, Unitree Go1/Go2/B2, and ANYmal 
(Fig.~\ref{fig:quadruped_platforms}) exhibit locomotion capabilities 
enabling traversal of terrain inaccessible to wheeled or tracked vehicles. These systems 
fundamentally depend on reliable communication links between 
operators and remote platforms, implemented through handheld 
controllers, web-based interfaces, or immersive VR/AR systems 
\cite{zhou2024teleoperation}.

\begin{figure}[htbp]
    \centering
    \includegraphics[width=\columnwidth]{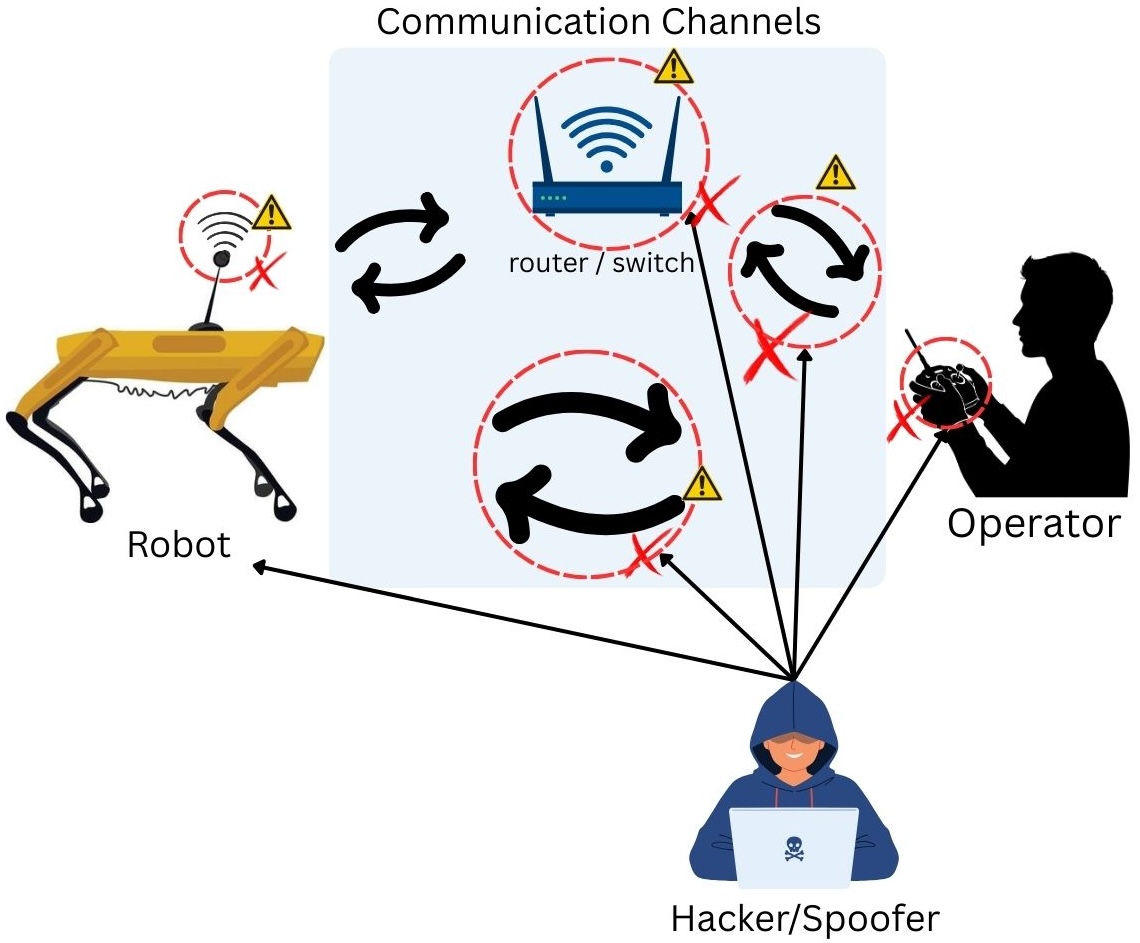}
    \caption{Conceptual illustration of communication channel attacks in quadruped robot teleoperation. An adversary can target the wireless link between the operator and robot to intercept, manipulate, or disrupt command and feedback transmissions.}
    \label{fig:comm_threat}
\end{figure}

The communication infrastructure supporting quadruped teleoperation 
spans multiple interdependent layers: wireless data links, video 
streaming pipelines, command transmission protocols, and state 
feedback channels. Each layer presents distinct attack surfaces 
where authentication enforced at one layer does not automatically 
extend to adjacent layers \cite{deng2022insecurity, ros2threatmodel2020}. 
Figure~\ref{fig:comm_threat} illustrates potential attack points 
along the communication channel. Unlike industrial robots operating in physically isolated cells, 
quadrupeds communicate over public wireless spectrum, exposing 
them to threats ranging from jamming and spoofing to perception 
manipulation and control hijacking \cite{botta2023survey}.

Quadruped platforms carry uniquely significant security implications 
arising from four distinguishing characteristics. First, operational 
kinetic energy is substantial: a 50~kg robot at 3~m/s carries 
approximately 225~J---sufficient to cause severe injury upon 
uncontrolled collision. Second, unlike statically stable wheeled 
robots, quadrupeds require continuous active balancing at 
100--1000~Hz; communication delays of 50~ms can cause missed 
swing-to-stance transitions during stair traversal. Third, 
deployment contexts frequently include critical infrastructure 
where malicious control can trigger cascading failures. Fourth, 
military and law enforcement applications introduce adversaries 
with advanced offensive capabilities \cite{neupane2023ai}. Recent 
security research demonstrated wormable exploits capable of 
propagating across robot fleets via Bluetooth Low Energy 
\cite{makris2025unipwn}.

\subsection{Scope and Contribution}

This survey systematically examines the communication, remote control,
and supervisory infrastructure connecting quadrupedal robots to human
operators, analyzing the attack surfaces of this critical link across
both conventional handheld controllers and immersive VR/AR interfaces.
Although existing reviews address general robotic cybersecurity or
autonomous vehicle security, no comprehensive survey focuses on the
distinct vulnerabilities arising from legged robot teleoperation.
This gap is consequential because quadrupedal robot platforms exhibit
fundamentally different attack--consequence profiles compared to
wheeled robots, UAVs, or industrial manipulators.

Our contributions are sixfold:

\begin{enumerate}[leftmargin=*]
    \item \textbf{Quadruped-Specific Threat Model}: This study investigates the significance of a security framework for quadrupedal robot teleoperation and evaluates its ramifications. We examine the distinct attack surfaces arising from dynamic stability requirements, operator cognitive burden, and gait-related vulnerabilities that may not be present in other robotic platforms or have not been addressed in this specific type of robot.
    
    \item \textbf{Attack Taxonomy with Consequence Mapping}: We introduce a six-tier taxonomy of cyber dangers, systematically correlating them with physical, human safety, mission-critical, infrastructure, and strategic repercussions, while offering quantitative metrics derived from recent literature.
    
    \item \textbf{Cascading Failure Analysis}: We present attack chain analysis that delineates how initial compromises disseminate throughout the cyber-physical system, leading to terminal effects, along with the identification of intervention points for defense prioritizing.
    
    \item \textbf{Maturity-Aware Defense Evaluation}: We methodically assess current countermeasures using a Technology Readiness Level (TRL) framework, distinctly categorizing field-deployed protections, laboratory-validated methods, and exploratory research concepts—addressing a significant deficiency in existing surveys that conflate speculative and production-ready defenses.
    
    \item \textbf{Practical Deployment Guidance}: We offer practical security advice categorized by implementation difficulty and organizational maturity, allowing practitioners to choose defenses according to feasible resource limitations.
    
    \item \textbf{Research Gap Identification}: We aim to identify significant unresolved issues pertaining to the security of quadruped robot remote operations and create a prioritized roadmap that differentiates between urgent practical advancements and long-term research objectives.
    
\end{enumerate}

Because quadruped-specific security research remains nascent---with only a handful of directly applicable empirical studies including Risiglione et al.'s TDPC validation~\cite{risiglione2021passivity}, the UniPwn BLE exploit~\cite{makris2025unipwn}, Ghost Robotics CVEs~\cite{incibe2025vision60}, and the RoboPAIR jailbreak~\cite{robson2024robopair}---this survey necessarily synthesizes cross-domain knowledge from autonomous vehicles, UAVs, and industrial robotics, reinterpreted through the lens of legged locomotion constraints and teleoperation-specific attack surfaces.

\subsection{Survey Methodology}
\label{sec:methodology}

To ensure thorough coverage while tackling the unique issues of controlling four-legged robots, we used a careful process for searching and choosing sources based on PRISMA-like guidelines. 

\textbf{Database Selection}: The systematic search encompassed five primary databases selected for their comprehensive coverage of robotics, cybersecurity, and control systems literature: IEEE Xplore, ACM Digital Library, Scopus, Web of Science, and arXiv. The search period spanned January 2019 through December 2025. Google Scholar served as a supplementary discovery tool for citation chaining but was not included in the primary database count due to its aggregator nature and potential for duplicate retrieval. Furthermore, we observed vulnerability advisories from INCIBE-CERT, CVE databases, and vendor security disclosures. Foundational works published before 2019 that established critical theoretical or empirical baselines for robotic cybersecurity—such as initial vulnerability demonstrations on teleoperated surgical robots \cite{bonaci2015surgical} and pioneering industrial robot security analyses \cite{cerrudo2017hacking, quarta2017robot}—were incorporated when they offer essential context for comprehending the current threat landscape.

\textbf{Search Terms}: Primary search queries combined robotic security terms (``robot
cybersecurity,'' ``ROS security,'' ``robotic intrusion detection'')
with platform-specific terms (``quadruped robot,'' ``legged robot,''
``legged locomotion'') and teleoperation terms (``teleoperation
security,'' ``VR teleoperation,'' ``remote robot control'').

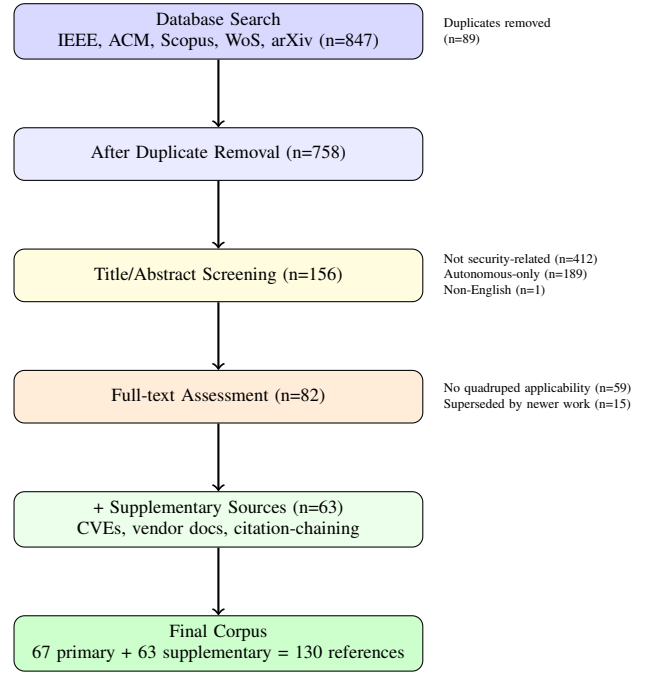
\begin{figure}[htbp]
\centering
\begin{tikzpicture}[node distance=0.9cm, 
    box/.style={rectangle, draw, rounded corners, text width=5.2cm, minimum height=0.7cm, align=center, font=\scriptsize},
    excl/.style={text width=3.0cm, font=\tiny, align=left},
    arrow/.style={->, thick}]
    
    \node[box, fill=blue!15] (id) {Database Search\\IEEE, ACM, Scopus, WoS, arXiv (n=847)};
    \node[box, fill=blue!8, below=of id] (dedup) {After Duplicate Removal (n=758)};
    \node[box, fill=yellow!15, below=of dedup] (screen) {Title/Abstract Screening (n=156)};
    \node[box, fill=orange!15, below=of screen] (elig) {Full-text Assessment (n=82)};
    \node[box, fill=green!8, below=of elig] (supp) {+ Supplementary Sources (n=63)\\CVEs, vendor docs, citation-chaining};
    \node[box, fill=green!20, below=of supp] (incl) {Final Corpus\\67 primary + 63 supplementary = 130 references};
    
    \node[excl, right=0.15cm of id] {Duplicates removed\\(n=89)};
    \node[excl, right=0.15cm of screen] {Not security-related (n=412)\\Autonomous-only (n=189)\\Non-English (n=1)};
    \node[excl, right=0.15cm of elig] {No quadruped applicability (n=59)\\Superseded by newer work (n=15)};
    
    \draw[arrow] (id) -- (dedup);
    \draw[arrow] (dedup) -- (screen);
    \draw[arrow] (screen) -- (elig);
    \draw[arrow] (elig) -- (supp);
    \draw[arrow] (supp) -- (incl);
\end{tikzpicture}
\caption{PRISMA-style flow diagram illustrating the systematic literature selection process. Preliminary database search across IEEE Xplore, ACM Digital Library, Scopus, Web of Science, and arXiv yielded 847 potentially relevant publications. Supplementary sources (n=63) include vendor security documentation, CVE databases, and publications identified through forward/backward citation tracking.}
\label{fig:prisma}
\end{figure}

\textbf{Selection Process}: The systematic search methodology is depicted in Fig.~\ref{fig:prisma}. Preliminary database search across IEEE Xplore, ACM Digital Library, Scopus, Web of Science, and arXiv yielded 847 potentially relevant publications. After removing 89 duplicates, the remaining
758 records underwent title and abstract screening. During this phase,
412 publications were excluded as not security-related, 189 addressed
only fully autonomous systems without teleoperation components, and
1 was excluded for non-English language, yielding 156 articles for
full-text review. Of these, 59 were excluded for lacking direct quadruped applicability, and 15 were superseded by more comprehensive subsequent publications from the same research groups, yielding 82 articles for quality assessment. During quality appraisal, 15 additional articles were reclassified as supplementary sources (providing contextual rather than primary evidence), resulting in 67 primary studies from database search. An additional 63 publications were identified through supplementary methods---including backward/forward citation tracking, manual searches of conference proceedings (ICRA, IROS, CCS, USENIX Security, S\&P), vendor security documentation, CVE database monitoring, and targeted queries for emerging topics such as LLM-based attacks and post-quantum cryptography---yielding a final corpus of 130 publications (67 from primary database search plus 63 from supplementary sources).

\begin{table}[htbp]
\caption{Literature Selection Summary}
\label{tab:selection_summary}
\centering
\footnotesize
\begin{adjustbox}{max width=\columnwidth}
\begin{tabular}{|l|r|l|}
\hline
\textbf{Stage} & \textbf{Count} & \textbf{Notes} \\
\hline
Database records identified & 847 & IEEE, ACM, Scopus, WoS, arXiv \\
After duplicate removal & 758 & 89 duplicates removed \\
Title/abstract screening & 156 & 602 excluded (not security-related, \\
 & & autonomous-only, non-English) \\
Full-text assessment & 82 & 74 excluded (no quadruped \\
 & & applicability, superseded) \\
Quality appraisal & 67 & 15 reclassified to supplementary \\
Supplementary sources & 63 & CVEs, vendor docs, citation chain \\
\hline
\textbf{Final corpus} & \textbf{130} & 67 primary + 63 supplementary \\
\hline
\end{tabular}
\end{adjustbox}
\end{table}

\subsubsection{Evidence Quality Appraisal}

To support transparent interpretation of findings, we classified all sources according to a three-tier evidence quality framework (Table~\ref{tab:evidence_quality}). Grade~A sources provide direct experimental validation on quadruped platforms; Grade~B sources offer peer-reviewed evidence from adjacent domains (autonomous vehicles, UAVs, general CPS) requiring cross-domain transfer assumptions; Grade~C sources include vendor documentation, CVE advisories, and gray literature that inform practitioner guidance but lack independent validation. Throughout this survey, claims derived primarily from Grade~B or Grade~C evidence are explicitly noted to distinguish direct quadruped evidence from cross-domain inference.

\begin{table}[htbp]
\caption{Evidence Quality Classification}
\label{tab:evidence_quality}
\centering
\footnotesize
\begin{adjustbox}{max width=\columnwidth}
\begin{tabular}{|c|p{3.5cm}|p{2.5cm}|c|}
\hline
\textbf{Grade} & \textbf{Criteria} & \textbf{Examples} & \textbf{Count} \\
\hline
A & Peer-reviewed with quadruped-specific validation & \cite{risiglione2021passivity, makris2025unipwn} & 15 \\
\hline
B & Peer-reviewed, cross-domain evidence & \cite{cao2019lidar, dong2020fdia, kundu2025cybersicknessattack} & 74 \\
\hline
C & Vendor docs, CVEs, gray literature & \cite{boston2024security, incibe2025vision60} & 42 \\
\hline
\end{tabular}
\end{adjustbox}

\vspace{2mm}
\scriptsize{\textbf{Note:} Grade reflects evidence directness for quadruped teleoperation, not intrinsic source quality. Grade~B sources may be high-quality research requiring domain transfer assumptions.}
\end{table}

\textbf{Inclusion Criteria}: Publications were included if they (1) addressed security aspects of robotic systems relevant to quadruped teleoperation, (2) presented novel attack methods, defense mechanisms, or vulnerability assessments, and (3) provided quantitative evaluations or formal analyses. We included important research on ROS/ROS2 security no matter the target platform because these frameworks are widely used in quadruped systems.

\textbf{Exclusion Criteria}:   We omitted publications that concentrate solely on autonomous navigation devoid of teleoperation elements, security assessments of fundamentally distinct robot morphologies (such as humanoid upper-body manipulation and underwater vehicles) lacking transferable insights, and studies that have been rendered obsolete by more recent works from the same research teams.

\textbf{Scope Justification}: The corpus of 130 publications provides comprehensive coverage of
quadruped-relevant security research and incorporates foundational studies from adjacent domains (bilateral teleoperation, industrial robotics, and autonomous vehicles) whose findings are transferable
to quadruped security. As demonstrated in our comparative analysis (Section V), quadruped-specific security research remains significantly less mature than UAV and autonomous vehicle domains. This survey explicitly addresses this gap by synthesizing cross-domain knowledge applicable to quadruped teleoperation while identifying where quadruped-specific research is urgently needed.

\textbf{Reproducibility and Data Availability:} To support reproducibility of the systematic review process, the complete search queries, inclusion/exclusion decision logs, and extracted data tables are available in the supplementary materials. The PRISMA checklist documenting adherence to systematic review guidelines is provided in Appendix~A. Platform security assessments are based exclusively on publicly available documentation, disclosed CVEs, and peer-reviewed publications; no proprietary or confidential information was utilized. The TRL classification rubric (Table~\ref{tab:trl_classification}) provides explicit criteria enabling independent replication of maturity assessments as new evidence emerges.

\subsection{Paper Organization}

The remainder of this paper is organized as follows. Section~II provides background on quadruped teleoperation architectures, establishes the security uniqueness of legged platforms, and reviews related work. Section~III presents the comprehensive attack taxonomy with consequence mapping and cascade failure analysis. Section~IV evaluates defense mechanisms with explicit maturity classification. Section~V provides comparative analysis across platforms and domains. Section~VI identifies open challenges and research gaps through analytical discourse. Section~VII delineates future research directions organized by temporal horizon. Section~VIII provides pragmatic implementation guidance for practitioners. Section~IX concludes the survey with principal findings, limitations, and outlook.

\section{Background and Related Work}

\subsection{Quadruped Robot Teleoperation Architectures}

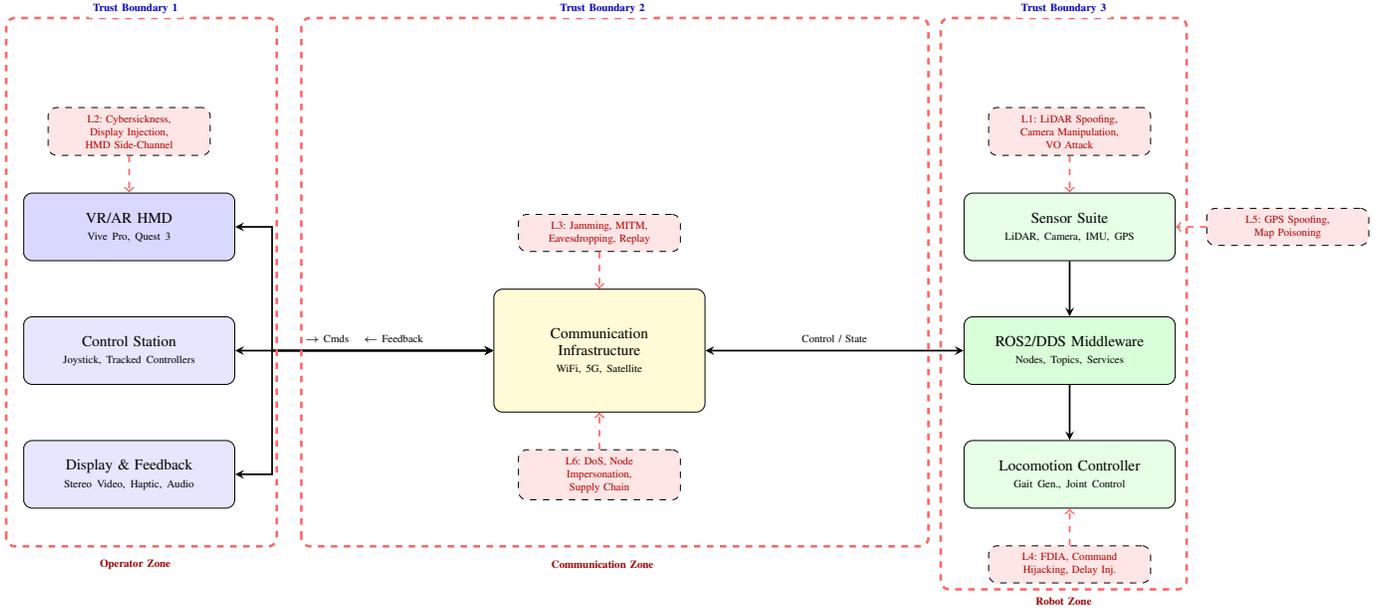
\begin{figure*}[htbp]
\centering
\resizebox{\textwidth}{!}{%
\begin{tikzpicture}[
    node distance=0.9cm,
    block/.style={rectangle, draw, rounded corners, minimum height=1.1cm, 
                  text width=3.2cm, align=center, font=\scriptsize},
    attack/.style={rectangle, draw, dashed, fill=red!10, rounded corners, 
                   text width=2.4cm, align=center, font=\tiny, text=red!70!black},
    trust/.style={draw=red!60, very thick, dashed, rounded corners=4pt},
    arrow/.style={->, thick, >=stealth},
    biarrow/.style={<->, thick, >=stealth}
]

\node[block, fill=blue!15] (hmd) 
    {VR/AR HMD\\{\tiny Vive Pro, Quest 3}};
\node[block, fill=blue!10, below=0.9cm of hmd] (control) 
    {Control Station\\{\tiny Joystick, Tracked Controllers}};
\node[block, fill=blue!10, below=0.9cm of control] (display) 
    {Display \& Feedback\\{\tiny Stereo Video, Haptic, Audio}};

\node[block, fill=yellow!20, right=4.2cm of control, minimum height=2.0cm] (comm) 
    {Communication\\Infrastructure\\{\tiny WiFi, 5G, Satellite}};

\node[block, fill=green!15, right=4.2cm of comm] (ros2) 
    {ROS2/DDS Middleware\\{\tiny Nodes, Topics, Services}};
\node[block, fill=green!10, above=0.9cm of ros2] (sensors) 
    {Sensor Suite\\{\tiny LiDAR, Camera, IMU, GPS}};
\node[block, fill=green!10, below=0.9cm of ros2] (actuators) 
    {Locomotion Controller\\{\tiny Gait Gen., Joint Control}};

\node[attack, above=0.6cm of hmd] (a2) 
    {L2: Cybersickness,\\Display Injection,\\HMD Side-Channel};

\node[attack, above=0.6cm of comm] (a3) 
    {L3: Jamming, MITM,\\Eavesdropping, Replay};

\node[attack, above=0.6cm of sensors] (a1) 
    {L1: LiDAR Spoofing,\\Camera Manipulation,\\VO Attack};

\node[attack, right=0.5cm of sensors] (a5) 
    {L5: GPS Spoofing,\\Map Poisoning};

\node[attack, below=0.6cm of comm] (a6) 
    {L6: DoS, Node\\Impersonation,\\Supply Chain};

\node[attack, below=0.6cm of actuators] (a4) 
    {L4: FDIA, Command\\Hijacking, Delay Inj.};

\draw[biarrow] (hmd.east) -- ++(0.6,0) 
    node[above, font=\tiny, midway]{} |- (comm.west);
\draw[biarrow] (control.east) -- 
    node[above, font=\tiny]{$\rightarrow$ Cmds \quad $\leftarrow$ Feedback} 
    (comm.west);
\draw[biarrow] (display.east) -- ++(0.6,0) |- (comm.west);

\draw[biarrow] (comm.east) -- 
    node[above, font=\tiny]{Control / State} 
    (ros2.west);
\draw[arrow] (sensors.south) -- (ros2.north);
\draw[arrow] (ros2.south) -- (actuators.north);

\draw[->, red!60, dashed, thick] (a2.south) -- (hmd.north);
\draw[->, red!60, dashed, thick] (a3.south) -- (comm.north);
\draw[->, red!60, dashed, thick] (a1.south) -- (sensors.north);
\draw[->, red!60, dashed, thick] (a4.north) -- (actuators.south);
\draw[->, red!60, dashed, thick] (a5.west) -- (sensors.east);
\draw[->, red!60, dashed, thick] (a6.north) -- (comm.south);

\draw[trust] (-2.0, -5.2) rectangle (2.4, 3.4);

\draw[trust] (2.8, -5.2) rectangle (13.0, 3.4);

\draw[trust] (13.2, -5.9) rectangle (17.2, 3.4);

\node[font=\tiny\bfseries, blue!70!black, fill=white, inner sep=2pt] 
    at (0.1, 3.55) {Trust Boundary 1};
\node[font=\tiny\bfseries, blue!70!black, fill=white, inner sep=2pt] 
    at (7.7, 3.55) {Trust Boundary 2};
\node[font=\tiny\bfseries, blue!70!black, fill=white, inner sep=2pt] 
    at (15.2, 3.55) {Trust Boundary 3};

\node[font=\tiny\bfseries, red!60!black] at (0.1, -5.5) {Operator Zone};
\node[font=\tiny\bfseries, red!60!black] at (7.7, -5.5) {Communication Zone};
\node[font=\tiny\bfseries, red!60!black] at (15.2, -6.1) {Robot Zone};

\end{tikzpicture}%
}
\caption{Quadruped teleoperation architecture with six-layer attack surface 
mapping. Red dashed boxes indicate attack vectors targeting each trust boundary 
zone. The three trust boundaries (Operator, Communication, Robot) correspond 
to the adversary model's attack access points. Each crossing between zones 
represents an exploitable interface where attacks from Layers 1--6 can 
intercept, manipulate, or disrupt the bidirectional command-feedback pipeline.}
\label{fig:architecture}
\end{figure*}

Contemporary quadruped teleoperation systems utilize stratified communication topologies linking human interfaces to robotic platforms. Figure~\ref{fig:architecture} depicts the standard architecture consisting of four fundamental levels.

\subsubsection{Traditional Controller-Based Teleoperation}
The most basic teleoperation method utilizes handheld devices (joysticks, gamepads) that communicate through WiFi or Bluetooth. Boston Dynamics Spot facilitates tablet-based control via WiFi, allowing the operator to see video feeds on the screen \cite{boston2024security}. Unitree platforms provide comparable capabilities via specialized controllers and mobile applications \cite{botta2023survey}. Communication latency needs are minimal (50-200 ms acceptable), although continual command streams and video feedback necessitate persistent bandwidth.

\subsubsection{VR/AR-Based Immersive Teleoperation}
Advanced teleoperation systems utilize head-mounted displays (HMDs) that deliver stereoscopic video feeds and facilitate intuitive control via tracked controllers or body movements \cite{zhou2024teleoperation}. Zhou et al. illustrated the HTC Vive-based teleoperation of Unitree Laikago for explosive ordnance disposal (EOD) operations, wherein operator immersion markedly enhanced manipulation precision \cite{zhou2024teleoperation}. The Open-TeleVision system facilitates coast-to-coast teleoperation at 60 Hz, illustrating the viability of long-distance VR control \cite{cheng2024opentv}. These systems enforce rigorous specifications: motion-to-photon latency must be under 20ms to avert cybersickness, high-resolution video streaming (2880$\times$1600 at 90Hz for Vive Pro), and low-latency bidirectional control channels. Cruz Ulloa et al. devised mixed-reality teleoperation techniques for superior control of legged-manipulator robots, demonstrating improved user experience and accuracy \cite{cruzulloa2024mr}.

\subsubsection{Web-Based and Cloud-Mediated Interfaces}
Cloud-connected operations provide remote access via web browsers, with video streams and commands transmitted through the manufacturer's infrastructure. The 2025 revelation of Unitree's CloudSail tunneling service indicated that robots sustained continuous connections to manufacturer servers, thereby establishing potential backdoor access without the operator's knowledge \cite{makris2025unipwn}. Such systems create supplementary attack vectors, encompassing cloud infrastructure breaches and supply chain weaknesses.

\subsection{The Uniqueness of Quadruped Teleoperation Security}
\label{sec:uniqueness}

Quadruped robots pose unique security challenges that cannot be mitigated by merely adapting methods from other robotic platforms. Four essential qualities distinguish quadruped teleoperation from UAVs, wheeled robots, and industrial manipulators:

\textbf{Dynamic Stability Requirements.} In contrast to wheeled robots that ensure static stability or UAVs capable of hovering, quadrupeds necessitate ongoing active balancing, employing control loops that function at frequencies between 100 and 1000 Hz. The stability margin during locomotion is fundamentally contingent upon the accurate timing of leg placement in relation to the trajectory of the center of mass. A communication delay of 50ms—imperceptible for a wheeled robot—can result in a quadruped missing the crucial swing-to-stance transition when ascending stairs, leading to a tumble fall. This establishes a fundamentally distinct threat model in which even transient communication interruptions during dynamic maneuvers are deemed safety-critical occurrences rather than simply performance declines.

\textbf{Significant Kinetic Energy with Ground Contact.} Table~\ref{tab:kinetic_comparison} compares kinetic energy and impact characteristics across robotic platform categories. A 50~kg quadruped traveling at 3~m/s carries 225~J of kinetic energy, equivalent to dropping a 15~kg mass from 1.5~m. Unlike UAVs that can hover to avoid collisions or industrial arms that work in controlled cells with safety enclosures, quadrupeds work near people in unstructured environments while keeping constant contact with the ground. The combination of mobility, mass, and operational environment leads to consequence severities similar to those of autonomous vehicles, rather than those of smaller mobile robots.

\begin{table}[htbp]
\caption{Platform-Specific Attack Surface Comparison}
\label{tab:kinetic_comparison}
\centering
\footnotesize
\begin{adjustbox}{max width=\columnwidth}
\begin{tabular}{|p{2.0cm}|c|c|c|c|}
\hline
\textbf{Characteristic} & \textbf{Quadruped} & \textbf{UAV} & \textbf{Wheeled} & \textbf{Ind. Arm} \\
\hline
Typical Mass & 12--70kg & 1--25kg & 10--500kg & 50--2000kg \\
\hline
Max Velocity & 3--6 m/s & 15--30 m/s & 1--5 m/s & 2--5 m/s \\
\hline
Kinetic Energy & 50--900J & 100--5000J & 5--6000J & N/A (fixed) \\
\hline
Impact Force & $>$3000N & Variable & $>$1000N & $>$10000N \\
\hline
Stability Type & Dynamic & Active & Static & Fixed \\
\hline
Sensor Attack Tolerance & Very Low & Medium & High & N/A \\
\hline
Recovery from Failure & Complex & Impossible & Trivial & N/A \\
\hline
Control Loop Criticality & $<$10ms & $<$100ms & $<$500ms & $<$1ms \\
\hline
Operator Cognitive Load & High & High & Low & Low \\
\hline
VR Teleoperation & Increasing & Common & Rare & Rare \\
\hline
\end{tabular}
\end{adjustbox}

\vspace{2mm}
\scriptsize{\textbf{Note:} KE calculated as $\frac{1}{2}mv^2$ at maximum operational velocity. Industrial arm KE listed as N/A since the base is fixed. \textbf{Qualitative Rating Definitions:} Sensor Attack Tolerance---Very Low: $<$50\,ms to destabilization under sensor corruption; Medium: 50--500\,ms tolerance; High: $>$500\,ms or graceful degradation possible. Recovery from Failure---Complex: requires active balancing and multi-step recovery sequence; Impossible: platform destruction likely (e.g., UAV crash); Trivial: passive stability enables stop-in-place. Control Loop Criticality---threshold latency beyond which stability margins are violated under nominal operating conditions.}
\end{table}

\textbf{Operator Cognitive Burden.} Teleoperating a legged robot requires the simultaneous management of 3D terrain evaluation, gait selection, balance preservation, and manipulation tasks. Research on operator cognitive load in teleoperation shows that controlling legged robots imposes a significantly higher mental workload compared to wheeled platforms \cite{zhou2024teleoperation}. This elevated cognitive demand introduces specific vulnerabilities:
factors that degrade operator performance---such as reduced video quality, communication latency, and cybersickness symptoms---have a disproportionately greater impact on quadruped teleoperation than on
wheeled-robot control. An operator experiencing mild symptoms of cybersickness may manage to continue teleoperating a wheeled robot but could unintentionally cause a quadruped to topple when navigating challenging terrain.

\textbf{Gait-Dependent Vulnerabilities.} The cyclical characteristics of legged mobility generate phase-dependent vulnerability intervals absent in other platforms. During the leg's swing phase, the robot functions within a diminished support polygon, resulting in a smaller stability margin. During stance shifts, the robot is susceptible to disturbances that may result in the newly positioned foot slipping. Shi et al. \cite{shi2024adversarial} quantitatively demonstrated at RSS 2024 that even state-of-the-art reinforcement learning-based quadrupedal locomotion controllers, including those utilized on ANYmal, can fail when subjected to low-magnitude adversarial sequences designed across multiple attack modalities (command, observation, and perturbation spaces concurrently). Their multi-modal adversarial framework successfully executed attacks when unimodal attacks were ineffective, demonstrating that gait phase-dependent vulnerabilities can be systematically exploited \cite{shi2024adversarial}. Attacks synchronized with these susceptible phases—through the analysis of stride patterns discernible in video feeds or predictable from terrain—could provide results with minimal necessary disruption. Such an approach presents chances for attackers as well as viable defensive strategies informed by gait-phase-aware security policies. Akter et al. \cite{akter2024legged} present an exhaustive overview of legged robot control techniques, encompassing gait generating approaches, the disruption of which comprises security vulnerabilities.

\subsection{Why Existing Solutions Fail for Quadrupeds}

Security solutions developed for other robotic domains often prove inadequate or inapplicable for quadruped teleoperation. Table~\ref{tab:solution_failures} summarizes key limitations.

\begin{table}[htbp]
\caption{Why Existing Robotic Security Solutions Fail for Quadrupeds}
\label{tab:solution_failures}
\centering
\footnotesize
\begin{tabular}{|p{2.0cm}|p{2.5cm}|p{2.8cm}|}
\hline
\textbf{Solution Domain} & \textbf{Approach} & \textbf{Quadruped Limitation} \\
 \hline
UAV Security & GPS-centric localization protection & Quadrupeds often operate GPS-denied (indoors, urban canyons) \\
\hline
UAV Security & Hover-and-wait on communication loss & Quadrupeds cannot hover; must maintain active balance \\
\hline
Industrial Robot & Safety caging and exclusion zones & Quadrupeds operate in unstructured environments with humans \\
\hline
Industrial Robot & Emergency stop as primary safety & E-stop during dynamic gait causes uncontrolled fall \\
\hline
Wheeled Robot & Graceful degradation under delay & 50ms delay causes instability during dynamic locomotion \\
\hline
Wheeled Robot & Simple odometry for localization & Leg slip, terrain variation require complex state estimation \\
\hline
Autonomous Vehicle & Extensive sensor redundancy & Weight/power constraints limit quadruped sensor suites \\
\hline
Autonomous Vehicle & V2X communication standards & No equivalent quadruped communication standards exist \\
\hline
\end{tabular}

\vspace{2mm}
\scriptsize{\textbf{Note:} Existing solutions assume either static stability (industrial arms, wheeled robots), hover capability (UAVs), or extensive onboard resources (autonomous vehicles)---none of which apply to quadrupeds operating in unstructured environments with dynamic balance requirements.}
\end{table}

The primary concern is that the majority of current solutions implicitly depend on assumptions that are inapplicable to quadrupeds: static stability (industrial arms, numerous wheeled platforms), the capacity to effortlessly disengage by hovering or flying away (UAVs), or ample onboard computational, sensing, and power resources (full-size autonomous vehicles).

Quadrupeds occupy a notable—and more perilous—intermediate position. They are dynamically stable, high-impulse devices that function within close proximity to individuals, generally under stringent payload and power limitations, and frequently in conditions lacking GPS or with compromised communication. In such context, a simplistic "emergency stop" does not inherently ensure safety; depending on the gait phase, terrain, and velocity, it may induce a loss of balance or uncontrolled descent, thereby exacerbating harm instead of mitigating it.

\subsection{Evolution of Robotic Cybersecurity}

The security of robotic systems has evolved through various stages. Before 2015, security was predominantly seen as a secondary consideration in robotics, with the majority of research focusing on functionality and performance \cite{cerrudo2017hacking}. Between 2015 and 2020, the domain transitioned towards systematic vulnerability identification and preliminary assessments, prompted by Cerrudo and Apa’s seminal 2017 IOActive study, which detailed exploitable flaws in many commercial robotic platforms \cite{cerrudo2017hacking}. In a seminal 2015 study, Bonaci et al. presented the inaugural empirical evidence of cyber attacks on teleoperated surgical robots (Raven II), demonstrating that hijacking, intention manipulation, and denial-of-service were achievable even in systems typically regarded as secure \cite{bonaci2015surgical}. Quarta et al. (2017) expanded these findings to industrial robot controls, illustrating that attackers could circumvent or compromise safety systems, resulting in physically detrimental behavior \cite{quarta2017robot}.

From 2018 to 2023, bilateral teleoperation security evolved into a discrete and progressively stringent research domain. Munteanu et al. experimentally categorized cyber-physical attacks on bilateral teleoperation systems based on target component, attacker knowledge, and adversarial objective, demonstrating that manipulation of the passivity layer can destabilize the slave side and induce unsafe behavior \cite{munteanu2018bilateral}. Alemzadeh et al. concurrently proposed dynamic, model-based detection and mitigation strategies for targeted assaults on surgical telerobots, achieving 90\% accuracy in distinguishing malicious orders from valid operator inputs \cite{alemzadeh2016targeted}. Yaacoub et al. (2022) expanded the framework with an extensive taxonomy of robotic vulnerabilities, assaults, and countermeasures across hardware, communication, and software layers, encompassing contemporary concerns such as software supply chain compromise \cite{yaacoub2022robotsecurity}. ACM Computing Surveys recently released a comprehensive examination of secure robotics, focusing on the interplay of safety, trust, and cybersecurity within cyber-physical systems \cite{secureroboticsacm2025}.

\begin{figure*}[htbp]
\centering
\begin{tikzpicture}[
    event/.style={rectangle, draw, rounded corners=2pt, text width=3.2cm, 
                  minimum height=0.5cm, align=left, font=\tiny, inner sep=3pt},
    eventN/.style={rectangle, draw, rounded corners=2pt, text width=2.8cm, 
                  minimum height=0.5cm, align=left, font=\tiny, inner sep=3pt},
    phase/.style={rectangle, draw, fill=#1, rounded corners=2pt, 
                  minimum height=0.45cm, text width=1.8cm, align=center, 
                  font=\scriptsize\bfseries, inner sep=2pt},
    yearl/.style={font=\footnotesize\bfseries, fill=white, inner sep=2pt},
    legend/.style={rectangle, draw, rounded corners=2pt, text width=1.6cm,
                   align=center, font=\tiny\bfseries, inner sep=3pt}
]

\draw[->, very thick, gray!60] (0,0) -- (18.0,0);

\foreach \x/\yr in {1.0/2015, 3.8/2017, 6.2/2019, 8.8/2021, 11.2/2023, 14.0/2025} {
    \draw[thick, gray!80] (\x,-0.15) -- (\x,0.15);
    \node[yearl] at (\x,-0.45) {\yr};
}

\node[phase=gray!20]   at (2.4, -1.0) {Pre-Awareness};
\node[phase=orange!20] at (5.0, -1.0) {Emergence};
\node[phase=yellow!20] at (7.5, -1.0) {Bilateral Teleop};
\node[phase=red!15]    at (10.0, -1.0) {Quadruped Era};
\node[phase=red!30]    at (12.6, -1.0) {Active Exploit};

\node[legend, fill=red!12] at (17.0, 0.8) {Attacks / Vulns.};
\node[legend, fill=green!12] at (17.0, -1.7) {Defenses};

\node[event, fill=red!8, anchor=south west] at (0.0, 0.6) 
    {Bonaci: First surgical robot cyber attack};
\draw[red!40, thin] (1.0, 0.6) -- (1.0, 0.15);

\node[event, fill=red!8, anchor=south west] at (4.4, 0.6) 
    {SROS2 DDS security specification};
\draw[red!40, thin] (6.2, 0.6) -- (6.2, 0.15);

\node[event, fill=red!8, anchor=south west] at (8.8, 0.6) 
    {RoboPAIR 100\% jailbreak on Go2};
\draw[red!40, thin] (11.2, 0.6) -- (11.2, 0.15);

\node[eventN, fill=red!15, anchor=south west] at (12.6, 0.6) 
    {Ghost V60 CVEs (CVSS 9.2)};
\draw[red!60, thin] (14.0, 0.6) -- (14.0, 0.15);

\node[event, fill=red!8, anchor=south west] at (0.0, 1.5) 
    {IOActive: Widespread robot vulnerabilities};
\draw[red!40, thin] (1.0, 1.5) -- (1.0, 0.6);

\node[event, fill=red!8, anchor=south west] at (4.4, 1.5) 
    {Cao: First LiDAR spoofing attack};
\draw[red!40, thin] (6.2, 1.5) -- (6.2, 0.6);

\node[event, fill=red!8, anchor=south west] at (8.8, 1.5) 
    {SROS2 V1--V3 flaws (ACM CCS)};
\draw[red!40, thin] (11.2, 1.5) -- (11.2, 0.6);

\node[eventN, fill=red!15, anchor=south west] at (12.6, 1.5) 
    {UniPwn wormable BLE exploit};
\draw[red!60, thin] (14.0, 1.5) -- (14.0, 0.6);

\node[event, fill=green!10, anchor=north west] at (0.0, -1.7) 
    {Quarta: Industrial robot security analysis};
\draw[green!40, thin] (1.0, -1.7) -- (1.0, -0.15);

\node[event, fill=green!10, anchor=north west] at (4.4, -1.7) 
    {CARLO LiDAR defense (2.3\%)};
\draw[green!40, thin] (6.2, -1.7) -- (6.2, -0.15);

\node[event, fill=green!10, anchor=north west] at (8.8, -1.7) 
    {ROSPaCe IDS dataset released};
\draw[green!40, thin] (11.2, -1.7) -- (11.2, -0.15);

\node[eventN, fill=green!10, anchor=north west] at (12.6, -1.7) 
    {ZTA and RV for ROS2 robotics};
\draw[green!40, thin] (14.0, -1.7) -- (14.0, -0.15);

\node[event, fill=green!10, anchor=north west] at (4.4, -2.5) 
    {TDPC quadruped teleoperation (ANA XPRIZE)};
\draw[green!40, thin] (6.2, -2.5) -- (6.2, -1.7);

\node[event, fill=green!10, anchor=north west] at (8.8, -2.5) 
    {DT-based CPS anomaly detection};
\draw[green!40, thin] (11.2, -2.5) -- (11.2, -1.7);

\end{tikzpicture}
\caption{Evolution of robotic cybersecurity research (2015--2025). Events above the timeline represent vulnerability discoveries and attacks (red); events below represent defensive advances (green). The accelerating density of events from 2023 onward reflects the transition from academic research to active exploitation of deployed quadruped platforms.}
\label{fig:timeline}
\end{figure*}
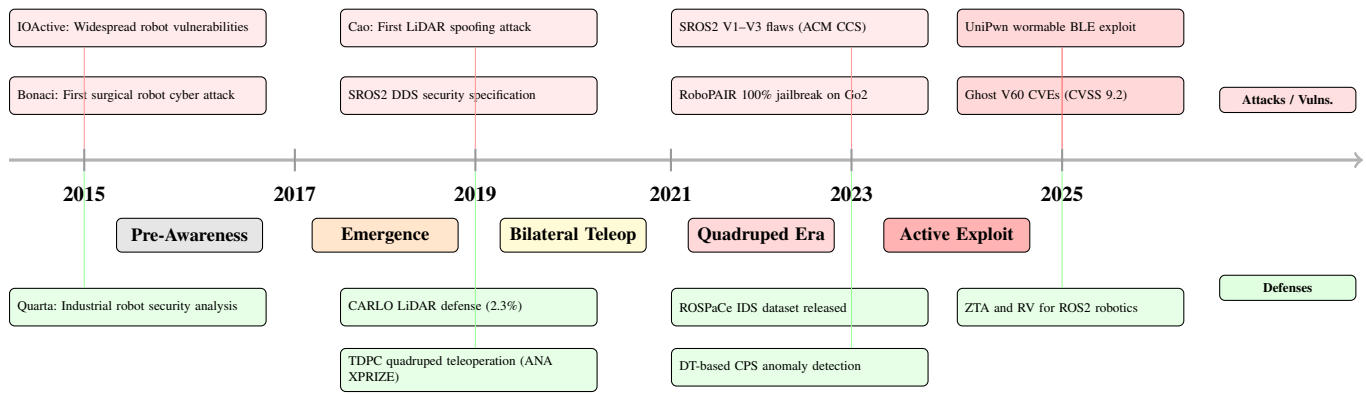

Between 2023 and 2025, the field has advanced due to two synergistic trends: the integration of AI-enabled robotic control into mainstream applications and the increasing deployment of systems in safety-critical, real-world environments. Takanashi et al. introduced encrypted four-channel bilateral control with homomorphic encryption, facilitating posture synchronization and force feedback while maintaining the confidentiality of control parameters \cite{takanashi2023encrypted}. During this timeframe, VR/AR interfaces have augmented the teleoperation framework and introduced novel attack surfaces, including the emergence of intentional "cybersickness attacks" aimed at the human operator as a component of the adversarial strategy \cite{yalcin2024cybersickness, kim2022cybersickness}. Khan et al. suggested security frameworks based on attack-tree analysis to systematically discover and prioritize exploitable vulnerabilities in cyber-physical robotic systems \cite{fortifyrobot2025}. Figure~\ref{fig:timeline} encapsulates these processes, emphasizing the accelerating interaction between vulnerability identification and the formulation of related responses.

\subsection{ROS and ROS2 Security Landscape}

The Robot Operating System (ROS) provides the dominant middleware framework for quadruped development. ROS1's architecture fundamentally lacked security provisions, relying on network isolation as the primary defense. The rosmaster node operated without authentication, and all topic communications occurred in plaintext, enabling trivial eavesdropping and injection attacks \cite{dieber2017security}.

ROS2, utilizing the Data Distribution Service (DDS) middleware, has created SROS2 (Secure ROS2), which inherently supports authentication, encryption, and access control in accordance with the DDS Security specification \cite{mayoral2022sros2}. The DDS Security specification (OMG DDS-SECURITY v1.1) delineates five security plugins: Authentication, Access Control, Cryptographic, Logging, and Data Tagging, which together provide identity-based authentication, granular topic-level access control, and message-specific encryption. The complexity of the specification presents considerable deployment obstacles: adequate configuration necessitates the generation of identity certificates for each participant, the formulation of governance and permissions documents in XML, and the distribution of Certificate Authority credentials—tasks that numerous research and operational deployments often overlook entirely \cite{diluoffo2019ros2}. Moreover, the DDS Quality of Service (QoS) policies governing reliability, durability, and deadline enforcement interact with security plugins in significant manners; implementing RELIABLE QoS alongside DDS Security elevates per-message overhead relative to BEST\_EFFORT, a tradeoff that is particularly critical for high-frequency quadruped control topics functioning at 100–1000 Hz. The DDS discovery protocol (Simple Participant Discovery Protocol, SPDP) disseminates participant information via well-known multicast addresses. Even with security measures in place, the discovery phase reveals network topology details, including participant GUIDs and accessible topics—data adequate for an adversary to delineate the robot's computational architecture.

Deng et al.'s seminal 2022 ACM CCS work revealed three significant design vulnerabilities (V1-V3) in SROS2 that permit unauthorized access retention despite the implementation of security measures \cite{deng2022insecurity}. The vulnerability V1 arises from inadequate permission revocation, permitting infected nodes to maintain access post-certificate revocation. V2 features inadequate namespace isolation, facilitating cross-domain information leaking. V3 reveals vulnerabilities in the discovery protocol that disclose network topology to passive attackers \cite{deng2022insecurity}. The official ROS 2 threat model \cite{ros2threatmodel2020} delineates entry points such as exposed debug ports, unsecured DDS domains, and firmware update channels, thereby affirming the extensive attack surface.

Yang et al. \cite{yang2024ros2formal} systematically characterized seven unique ROS2 communication vulnerabilities across subjects, services, and actions, demonstrating that each can compromise one or more confidentiality, integrity, and availability (CIA) features. Employing state-transition system models and Linear Temporal Logic (LTL), they identified failure modes in essential systems such as node discovery, topic subscription, and service invocation. These results are significant as they transition the discourse on ROS2 security from "misconfiguration anecdotes" to a meticulously delineated attack surface with technically provable breakpoints.

DiLuoffo et al. illustrated the security of ROS2 with DDS Security at the implementation level, although highlighted a persistent operational challenge: the burdensome configuration and intricate certificate governance frequently compel teams to forgo security entirely \cite{diluoffo2019ros2}. The disparity between "security is supported" and "security is actually deployed" is reflected in extensive ROS2 ecosystem surveys, indicating that numerous systems are only partially configured—thus facilitating opportunities for replay attacks and unauthorized subscriptions, despite developers' belief that they are "utilizing ROS2 security" \cite{ros2survey2025}.

The vulnerability is heightened by the reliance of ROS2’s security framework on both ROS tools and the foundational DDS implementations. Mayoral-Vilches et al. (Alias Robotics) documented approximately 15 significant vulnerabilities impacting DDS stacks and identified over 643 unique public-facing DDS services across 34 nations, demonstrating that DDS exposure is not merely theoretical—it is quantifiable, geographically dispersed, and currently observable from the open internet \cite{aliasrobotics2022dds}. The Robot Vulnerability Database (RVD) has documented more than 1,000 robotics-related vulnerabilities \cite{mayoral2023rvd}, while the Robot Vulnerability Scoring System (RVSS) enhances the Common Vulnerability Scoring System (CVSS) by incorporating robotics-specific factors to more accurately reflect cyber-physical ramifications and operational context \cite{vuln1000robots}.

A particularly concerning development is the 2025 demonstration of software supply chain attacks targeting SROS2. Severov et al. showed that trojanized sros2 CLI packages can exfiltrate keystore credentials during keystore/enclave creation—e.g., via covert DNS queries—enabling follow-on node impersonation within otherwise “secure” DDS graphs \cite{sros2supplychain2025}. The a significant observation is that this bypasses the usual network-facing threat model: you can do everything “right” at the DDS Security layer and still lose the keys if your tooling is poisoned at install-time.

This result also fits a wider pattern in modern software risk, where attackers compromise trusted build, packaging, or dependency channels rather than the runtime system directly. High-profile incidents such as the 2024 XZ Utils backdoor (CVE-2024-3094) illustrate how malicious logic can be inserted into widely trusted ecosystems and remain undetected long enough to threaten downstream users—exactly the kind of exposure that matters for ROS deployments that depend heavily on precompiled packages and third-party repositories \cite{sros2supplychain2025}.

\subsection{Commercial Platform Security Comparison}

The quadruped robot market shows stark security unevenness across manufacturers, deployment models, and target customers. Table~\ref{tab:quadruped_security} summarizes these differences across six major platforms, revealing that “quadruped security” is not a single maturity curve but a fragmented landscape shaped by cost targets, update pipelines, and whether the robot is designed for enterprise fleets or fast-moving consumer/prosumer adoption.

Boston Dynamics Spot is widely treated as a security benchmark in this space. Its stack includes TLS 1.2+ for encrypted communications, mutual authentication via X.509 client certificate verification, firmware integrity mechanisms based on cryptographically signed hash trees, JWT-based authentication using ES256, and per-device unique encryption keys—choices that collectively raise the bar for both remote compromise and supply-chain tampering \cite{boston2024security}.

On the other hand, Unitree platforms have always shown serious flaws that affect the entire fleet. The 2025 UniPwn leak showed that Unitree robots have a hardcoded AES key that is used when setting up BLE. This means that if one device is hacked, it might be possible to decrypt provisioning communication between several units \cite{unipwn2025spectrum}. The BLE provisioning workflow allegedly demonstrated insufficient input validation, allowing the injection of improperly formatted Wi-Fi credentials to enable root-level access; owing to the wireless characteristics of the vector, the attack could potentially become wormable, allowing a compromised robot to autonomously disseminate to other Unitree robots within BLE proximity \cite{unipwn2025spectrum}.

Mayoral-Vilches et al. reported data exfiltration behavior on the Unitree G1 humanoid, detailing the periodic transfer of telemetry (such as joint states, battery status, and IMU data) and recurrent audio capture to distant servers using TLS-encrypted channels \cite{mayoral2025unitreeg1}. Although the G1 is a humanoid platform, its shared firmware architecture, BLE provisioning stack, and cloud backend with Unitree's quadruped product line (Go2, B2) make its documented vulnerabilities directly relevant to quadruped deployments. RoboPAIR indicated a 100\% success rate in circumventing LLM safety guardrails on the Unitree Go2, highlighting that "cyber" and "behavioral" safety barriers may simultaneously fail when autonomy is influenced by language-model components \cite{robson2024robopair}.

These results are seen in official vulnerability monitoring. CVE-2025-35027 delineates a BLE command-injection vulnerability impacting various Unitree product lines (Go2/G1/H1/B2) \cite{unipwncve2025}, while CVE-2025-2894 identifies a backdoor-like state in the preceding Go1 series that facilitates remote surveillance without authentication \cite{unitreego1backdoor}. The Go1 issue predates UniPwn by several years, and subsequent revelations indicate that security enhancements have not been consistently implemented across generations \cite{makris2025unipwn}.

Ghost Robotics' Vision 60, extensively utilized by U.S. and allied military forces for perimeter security and reconnaissance, was revealed to possess several significant vulnerabilities disclosed by INCIBE-CERT in October 2025 \cite{incibe2025vision60}. The critical vulnerability, CVE-2025-41108 (CVSS 9.2), arises from an MAVLink-based control channel that allegedly lacked encryption and authentication, enabling a remote attacker on the same Wi-Fi network (or accessible via 4G/LTE, contingent on deployment) to impersonate the operator tablet and execute arbitrary commands \cite{incibe2025vision60}.

A second issue, CVE-2025-41109 (CVSS 8.7, High), pertains to exposed and unauthenticated physical interfaces, including three RJ-45 ports and a USB-C port. The robot’s internal router reportedly assigns an IP address to any connected device; coupled with a ROS 2 deployment that is often left unauthenticated by default, this could grant immediate access to control surfaces and data flows upon physical access \cite{incibe2025vision60}. CVE-2025-41110 (CVSS 7.0, High) has revealed hardcoded Wi-Fi and SSH credentials within the Android control application (APK), representing a design weakness that could compromise access control through a singular exposed binary artifact \cite{incibe2025vision60}.

ANYmal from Swiss company ANYbotics targets enterprise deployments in oil/gas facilities, power plants, and mining operations where safety certifications are mandatory \cite{hutter2017anymal}. However, the platform's security posture remains opaque: no independent security audit has been published, no CVE has been disclosed, and implementation details regarding encryption, authentication, and key management are not publicly documented. While the platform holds ATEX/IECEx certifications for operation in explosive atmospheres, these standards address physical safety rather than cybersecurity, and compliance with them cannot be interpreted as evidence of resilience against deliberate cyber attacks. The absence of public security documentation is itself a concern for organizations conducting threat assessments, as it precludes independent verification of the protections claimed by the manufacturer.

Xiaomi CyberDog, an open-source platform based on ROS2 Galactic with Cyclone DDS middleware, emphasizes accessibility and rapid experimentation rather than secure-by-default deployment \cite{xiaomi2021cyberdog}. Publicly reported default SSH credentials (username: \texttt{mi}, password: \texttt{123}) create an immediate remote-entry vector if the device is network-reachable, effectively reducing the system’s security to basic network perimeter assumptions. In addition, the platform’s integration with vendor cloud services can introduce data governance and sovereignty concerns in sensitive environments; however, the scope of this risk depends on verifiable details—what telemetry is transmitted, to which endpoints, under what authentication and opt-out controls—which should be documented or empirically measured before drawing definitive surveillance conclusions.

\begin{table*}[htbp]
\caption{Comprehensive Security Analysis of Commercial Quadruped Platforms}
\label{tab:quadruped_security}
\centering
\footnotesize
\begin{tabular}{|p{1.8cm}|p{2.0cm}|p{2.0cm}|p{2.0cm}|p{2.0cm}|p{2.0cm}|p{2.0cm}|}
\hline
\textbf{Aspect} & \textbf{Boston Dynamics Spot \cite{boston2024security}} & \textbf{Unitree Go1/Go2/G1 \cite{unipwn2025spectrum,mayoral2025unitreeg1}} & \textbf{Ghost Robotics Vision 60 \cite{incibe2025vision60}} & \textbf{ANYmal (ANYbotics) \cite{hutter2017anymal}} & \textbf{Xiaomi CyberDog \cite{xiaomi2021cyberdog}} & \textbf{DeepRobotics X30 \cite{deeprobotics2024x30}} \\
\hline
Encryption & TLS 1.2+/HTTP2 & AES (fleet-wide) & None (critical) & Enterprise TLS & Cyclone DDS & Not disclosed \\
\hline
Authentication & JWT/ES256, X.509 & BLE (vulnerable) & None in protocol & Enterprise PKI & Default mi/123 & Not disclosed \\
\hline
Firmware Security & Signed hash trees & Limited & Not documented & Industrial cert. & Open source & Proprietary \\
\hline
Key Management & Per-device unique & Hardcoded fleet & Hardcoded in APK & Per-device & Not documented & Not disclosed \\
\hline
Physical Ports & Restricted & Open & Unauth. (critical) & Industrial sealed & Open access & Standard \\
\hline
Cloud Backend & Optional, controlled & CloudSail backdoor & Military networks & Enterprise cloud & Xiaomi servers & Proprietary \\
\hline
Known CVEs & None public & Multiple \cite{robson2024robopair,unipwn2025spectrum} & CVE-2025-41108/09/10 \cite{incibe2025vision60} & None public & ROS2 inherited \cite{deng2022insecurity} & None public \\
\hline
ROS Version & Proprietary SDK & ROS1/ROS2 & Proprietary & ROS2 & ROS2 Galactic & ROS2 \\
\hline
Target Market & Enterprise/Military & Consumer/Research & Military/Defense & Industrial & Consumer/Research & Industrial \\
\hline
Security Maturity & High & Very Low & Low & Medium-High & Low & Unknown \\
\hline
\end{tabular}

\vspace{2mm}
\scriptsize{\textbf{Note:} Security maturity varies dramatically across platforms. Boston Dynamics represents industry best practice while consumer platforms exhibit critical vulnerabilities. Military platforms present concerning security gaps despite high-consequence deployment contexts.}
\end{table*}

Table~\ref{tab:scoring_rubric} defines the scoring rubric used 
for platform comparison. Based on this rubric, 
 Table~\ref{tab:platform_scores} assigns numerical scores (0--10) across five security dimensions using a structured evidence-based methodology. Each dimension is scored according to the rubric in Table~\ref{tab:scoring_rubric}, with scores derived from: (i) vendor documentation and published security whitepapers, (ii) independently disclosed CVEs weighted by CVSS severity (Critical CVE: $-3$ points; High: $-2$; Medium: $-1$), (iii) peer-reviewed security analyses, and (iv) empirical vulnerability demonstrations. Where multiple evidence sources conflict, the more conservative (lower) score is assigned. Confidence indicators (H/M/L) reflect the completeness of available evidence: High indicates $\geq$3 independent sources corroborating the assessment; Medium indicates 1--2 sources; Low indicates reliance on indirect inference or limited public documentation. Scores reflect independently verifiable characteristics: Encryption quality considers protocol strength and key management; Authentication evaluates mechanism robustness and default configuration; Firmware Security assesses signing, verification, and update integrity; Vulnerability History counts known CVEs weighted by CVSS severity; and Supply Chain Trust evaluates cloud backend transparency and data sovereignty practices.

\begin{table}[!t]
\caption{Security Scoring Rubric for Platform Comparison}
\label{tab:scoring_rubric}
\centering
\footnotesize
\begin{adjustbox}{max width=\columnwidth}
\begin{tabular}{c|p{6cm}}
\hline
\textbf{Score} & \textbf{Criteria} \\
\hline
9--10 & Documented implementation with per-device certificates, TLS~1.3+, 
        or equivalent; independent audit or CVE-free track record $\geq$2~years \\
7--8  & Enterprise-grade security (TLS, PKI) documented; minor gaps 
        (e.g., default credentials in early firmware) \\
5--6  & Basic encryption available but not enabled by default; 
        limited public documentation \\
3--4  & Minimal security features; known unpatched vulnerabilities \\
1--2  & No documented security measures; critical CVEs outstanding \\
0     & No information available or security explicitly absent \\
\hline
\end{tabular}
\end{adjustbox}

\vspace{1mm}
\begin{flushleft}
\footnotesize Scores reflect independently verifiable characteristics from 
vendor documentation, published CVEs, and preprint analyses. Where 
vendor information is unavailable, a confidence indicator (H/M/L) is 
appended to denote assessment reliability.
\end{flushleft}
\end{table}

\begin{table}[htbp]
\caption{Quantitative Security Scoring of Commercial Quadruped Platforms}
\label{tab:platform_scores}
\centering
\footnotesize
\begin{adjustbox}{max width=\columnwidth}
\begin{tabular}{|l|c|c|c|c|c|c|}
\hline
\textbf{Dimension} & \textbf{Spot} & \textbf{Unitree} & \textbf{Ghost} & \textbf{ANYmal} & \textbf{Xiaomi} & \textbf{Deep} \\
\hline
Encryption & 9 & 3 & 1 & 7 & 4 & -- \\
\hline
Authentication & 9 & 2 & 1 & 7 & 2 & -- \\
\hline
Firmware Sec. & 9 & 3 & 2 & 7 & 5 & -- \\
\hline
Vuln. History & 9 & 1 & 2 & 8 & 5 & -- \\
\hline
Supply Chain & 8 & 2 & 5 & 7 & 3 & -- \\
\hline
\textbf{Composite} & \textbf{8.8} & \textbf{2.2} & \textbf{2.2} & \textbf{7.2} & \textbf{3.8} & \textbf{N/A} \\
\hline
\end{tabular}
\end{adjustbox}

\vspace{2mm}
\scriptsize{\textbf{Note:} 0=absent/critical, 5=partial/adequate, 10=industry best practice. --: Insufficient public information. Composite: unweighted mean. Ghost Robotics and Unitree achieve identical composite scores despite different threat profiles: Ghost lacks basic protections in a military context, while Unitree has architectural vulnerabilities across a consumer product line.}

\end{table}

The composite scores in Table~\ref{tab:platform_scores} employ uniform weighting across the five security dimensions. However, operational contexts may warrant differential emphasis: safety-critical deployments prioritize authentication and firmware integrity, while mission-critical applications emphasize vulnerability history and supply chain trust. To assess scoring robustness, Table~\ref{tab:sensitivity_analysis} presents composite scores under three weighting schemes. The divergence between Ghost Robotics and Unitree under safety-weighted scoring is particularly instructive: despite identical equal-weighted composites (2.2), Ghost drops to 1.4 under safety weighting due to its critical authentication gap (score of 1), while Unitree's weaknesses are more uniformly distributed. This analysis confirms that aggregate scores can obscure operationally significant differences, reinforcing the importance of dimension-level assessment for deployment decisions.

\begin{table}[htbp]
\caption{Sensitivity Analysis: Platform Composite Scores Under Alternative Weighting Schemes}
\label{tab:sensitivity_analysis}
\centering
\footnotesize
\begin{adjustbox}{max width=\columnwidth}
\begin{tabular}{|l|c|c|c|c|}
\hline
\textbf{Platform} & \textbf{Equal} & \textbf{Safety} & \textbf{Mission} & \textbf{Conf.} \\
\hline
Boston Dynamics Spot & 8.8 & 9.0 & 8.6 & H \\
Unitree Go1/Go2/G1 & 2.2 & 2.0 & 1.8 & H \\
Ghost Robotics Vision 60 & 2.2 & 1.4 & 3.0 & H \\
ANYmal (ANYbotics) & 7.2 & 7.0 & 7.4 & L \\
Xiaomi CyberDog & 3.8 & 3.2 & 4.2 & M \\
DeepRobotics X30 & N/A & N/A & N/A & -- \\
\hline
\end{tabular}
\end{adjustbox}

\vspace{2mm}
\scriptsize{\textbf{Note:} Equal = uniform weights across five dimensions; Safety = 2$\times$ weight on Authentication and Firmware Security; Mission = 2$\times$ weight on Vulnerability History and Supply Chain Trust. \textbf{Conf.} = assessment confidence (H=High, M=Medium, L=Low based on public documentation availability).}
\end{table}

This security heterogeneity confirms that platform selection decisions carry significant security implications that practitioners must weigh alongside capability and cost factors.

Table~\ref{tab:data_availability} summarizes the availability of security-relevant documentation for each platform, which directly influences assessment confidence. Platforms with limited public documentation (ANYmal, DEEP Robotics X30) receive conservative security ratings that may not reflect actual implementation quality.

\begin{table}[htbp]
\caption{Platform Security Documentation Availability}
\label{tab:data_availability}
\centering
\footnotesize
\begin{adjustbox}{max width=\columnwidth}
\begin{tabular}{|l|c|c|c|c|}
\hline
\textbf{Platform} & \textbf{Public Docs} & \textbf{CVEs} & \textbf{Academic} & \textbf{Confidence} \\
\hline
Boston Dynamics Spot & High & Yes & Yes & High \\
Unitree Go1/Go2/B2 & Medium & Yes & Yes & High \\
Ghost Robotics Vision 60 & Low & Yes & Limited & Medium \\
ANYbotics ANYmal & Low & No & Limited & Low \\
Xiaomi CyberDog & Medium & No & Limited & Medium \\
DEEP Robotics X30 & Low & No & No & Low \\
\hline
\end{tabular}
\end{adjustbox}

\vspace{2mm}
\scriptsize{\textbf{Note:} Confidence reflects the reliability of security assessments based on available evidence. Platforms with ``Low'' confidence may have stronger (or weaker) security than documented; absence of disclosed CVEs does not imply absence of vulnerabilities.}
\end{table}

\subsection{Related Surveys}

Table~\ref{tab:survey_comparison} provides a systematic comparison of this work against existing surveys, highlighting the specific contributions and gaps each addresses.

\begin{table*}[htbp]
\caption{Comparison with Existing Surveys on Robotic Cybersecurity}
\label{tab:survey_comparison}
\centering
\footnotesize
\begin{tabular}{|p{2.8cm}|c|c|c|c|c|c|}
\hline
\textbf{Survey} & \textbf{Quadruped} & \textbf{Teleop.} & \textbf{TRL} & \textbf{Attack-Conseq.} & \textbf{VR/AR} & \textbf{Platform} \\
 & \textbf{Specific} & \textbf{Security} & \textbf{Assess.} & \textbf{Mapping} & \textbf{Attacks} & \textbf{Comparison} \\
\hline
Botta et al. (2023) \cite{botta2023survey} & \ding{55} & Partial & \ding{55} & \ding{55} & \ding{55} & Limited \\
\hline
Verma et al. (2025) \cite{verma2025survey} & \ding{55} & \ding{55} & \ding{55} & \ding{55} & \ding{55} & \ding{55} \\
\hline
Yaacoub et al. (2022) \cite{yaacoub2022robotsecurity} & \ding{55} & Partial & \ding{55} & Partial & \ding{55} & \ding{55} \\
\hline
Hamdan \& Mahmoud (2021) \cite{hamdan2021teleop} & \ding{55} & \ding{51} & \ding{55} & Partial & \ding{55} & \ding{55} \\
\hline
Li et al. (2025) \cite{ansnetwork2025} & \ding{55} & \ding{55} & \ding{55} & Partial & \ding{55} & \ding{55} \\
\hline
\textbf{This Work} & \ding{51} & \ding{51} & \ding{51} & \ding{51} & \ding{51} & \ding{51} \\
\hline
\end{tabular}

\vspace{2mm}
\scriptsize{\textbf{Note:}\ding{51}: Comprehensively addressed. \ding{55}: Not addressed. Partial: Tangentially discussed without systematic treatment. This work is the first to combine quadruped-specific threat analysis with TRL-based defense maturity assessment and VR/AR attack coverage.}
\end{table*}

To quantify the coverage differential, we define the \textit{domain specificity index} (DSI) as the fraction of surveyed papers directly addressing the target platform category (quadrupeds, UAVs, wheeled robots, or industrial arms). Existing surveys exhibit low quadruped DSI: Botta et al.~\cite{botta2023survey} achieve DSI $\approx$ 0.03 (estimated 4/127 papers with quadruped relevance); Verma et al.~\cite{verma2025survey} achieve DSI $\approx$ 0.02. In contrast, the present survey achieves DSI $\approx$ 0.52 (67/130 sources with direct or strongly transferable quadruped applicability), representing a 17--26$\times$ improvement in domain-specific coverage. This differential reflects our explicit focus on legged-platform security rather than generic robotic cybersecurity, enabling deeper treatment of quadruped-specific phenomena (gait-phase vulnerabilities, dynamic stability constraints, teleoperation-specific attack surfaces) that receive at most cursory mention in broader surveys.

Numerous surveys examine robotic cybersecurity; however, none offers an exhaustive analysis of quadruped teleoperation or the security problems specific to legged robots, which stem from dynamic stability, high-bandwidth control loops, and operation in close proximity to humans. Botta et al. (2023) offered a broad survey spanning industrial manipulators, mobile robots, and drones; however, coverage of legged platforms was limited, and the survey did not examine how stability margins and contact dynamics reshape attack impact and safety risk \cite{botta2023survey}. Verma et al. (2025) performed a systematic review of 127 publications and stressed the importance of domain-specific threat models, yet did not develop a quadruped-focused analysis or teleoperation-centered attacker models \cite{verma2025survey}. Neupane et al. (2023) surveyed the AI–robotics security nexus, including vulnerabilities in learning-based controllers that are relevant to quadrupeds using neural gait policies, but without tying these vulnerabilities to legged-specific failure modes such as loss of balance, impulsive ground-contact disturbances, or stability-constrained recovery behaviors \cite{neupane2023ai}.

Complementary surveys address significant subproblems that intersect with quadruped security, yet fail to bridge the quadruped-teleoperation gap.Pirayesh and Zeng (2022) performed a comprehensive analysis of jamming attacks and countermeasures pertinent to robotic communications; however, they neglected to investigate the interaction among teleoperation control authority, latency sensitivity, and stability constraints in contexts of denial or degradation \cite{pirayesh2022jamming}. Kim and Kaur (2024) investigated adversarial robustness in LiDAR, relevant to perception-dependent teleoperation and autonomy systems, concentrating mainly on the sensor rather than the platform \cite{kim2024lidar}. Chi et al. (2025) examined physical-world adversarial attacks in autonomous driving and created taxonomies that partially apply to quadrupeds utilizing cameras, LiDAR, and inertial sensing; however, the vehicular context fails to account for legged contact transitions and the heightened consequences of short-horizon destabilization \cite{chi2025adversarial}.Wang et al. (2024) investigated UAV swarm network security and offered insights relevant to multi-robot coordination; however, the assumptions concerning aerial mobility and lenient ground-contact requirements limit their direct applicability to legged teams \cite{wanguavswarm2024}. Holdbrook et al. (2024) performed a survey on network-based intrusion detection for industrial and robotic systems, emphasizing the lack of quadruped-specific datasets—an empirical shortcoming that hinders the evaluation of security monitoring in authentic locomotion and teleoperation contexts \cite{holdbrook2024ids}.

Overall, the literature provides substantial coverage of robot cybersecurity broadly and of individual attack surfaces (communications, perception, learning-based controllers, and IDS). However, it has not yet synthesized these threads into a quadruped-centered framework that connects cybersecurity objectives to stability-critical control, human-in-the-loop teleoperation, and legged mobility constraints.

Hamdan and Mahmoud (2021) conducted an extensive survey of control systems for bilateral teleoperation in the context of cyber assaults, examining detection and mitigation measures for DoS, FDIA, and replay attack categories \cite{hamdan2021teleop}. Zafir et al. (2024) investigated security improvements for the Internet of Robotic Things (IoRT), proposing integrated security frameworks while neglecting constraints particular to legged platforms \cite{zafir2024iorot}. Li et al. (2025) performed a systematic assessment of cyber assaults and defenses pertaining to autonomous navigation systems, offering a comprehensive categorization applicable to quadruped outdoor navigation \cite{ansnetwork2025}. Lee et al. (2024) conducted a survey on learning-based legged locomotion techniques, detailing the current advancements in reinforcement learning-based gait controllers, with a focus on the developing security issue of adversarial robustness \cite{lee2024learning}.

The lack of security assessments tailored for quadruped robots, despite their swift integration into critical applications, signifies a notable deficiency that this study seeks to rectify. By defining the distinct threat model for quadruped robots teleoperation and methodically examining attack-to-consequence routes, we lay the groundwork for focused security research and effective defensive implementation.

\textbf{Methodological Caveat on Cited Attack Effectiveness:} The attack success rates reported throughout this taxonomy derive from empirical studies with varying statistical rigor. Many foundational demonstrations---particularly perception-layer attacks and operator-targeting mechanisms---involve sample sizes below 50 trials, limiting statistical power for precise effect-size estimation. Studies reporting success rates $>$90\% with small samples may reflect optimistic bias; the true population success rate confidence intervals are substantially wider than point estimates suggest. Where possible, we cite studies with larger sample sizes or meta-analytic aggregation; where such evidence is unavailable, we note the preliminary nature of quantitative claims. Practitioners should interpret high success rates as indicating \textit{feasibility under favorable conditions} rather than \textit{guaranteed effectiveness} across operational variability.

\section{Attack Taxonomy and Consequence Analysis}

We provide a systematic classification of cyber risks aimed at the communication and control systems of teleoperated quadruped robots, along with a clear correlation to the resultant repercussions of successful attacks. The taxonomy consists of six assault tiers, each aimed at specific elements of the teleoperation infrastructure, followed by thorough consequence analysis and discovery of cascade failure pathways.

\subsection{Adversary Model and Threat Assumptions}

To ground the subsequent taxonomy in a formal framework, we define the adversary model using the STRIDE classification adapted for cyber-physical robotic systems. Formally, we model the adversary as a tuple $\mathcal{A} = \langle \mathcal{C}, \mathcal{K}, \mathcal{O}, \mathcal{R} \rangle$ where:

\begin{itemize}[leftmargin=*]
    \item $\mathcal{C} \in \{C_1, C_2, C_3\}$ denotes capability tier: $C_1$ (opportunistic, remote-only, public exploit knowledge), $C_2$ (sophisticated, proximity-based, protocol expertise), $C_3$ (APT, supply-chain access, platform-specific knowledge);
    \item $\mathcal{K} \subseteq \{k_\text{arch}, k_\text{dyn}, k_\text{gait}, k_\text{crypto}\}$ represents adversary knowledge: system architecture ($k_\text{arch}$), dynamic model ($k_\text{dyn}$), gait parameters ($k_\text{gait}$), and cryptographic keys ($k_\text{crypto}$);
    \item $\mathcal{O} \in \{\text{disrupt}, \text{degrade}, \text{deceive}, \text{destroy}, \text{exfiltrate}\}$ specifies primary objective;
    \item $\mathcal{R} \subseteq \{r_\text{net}, r_\text{RF}, r_\text{phys}, r_\text{supply}\}$ defines accessible attack resources: network access ($r_\text{net}$), RF equipment ($r_\text{RF}$), physical proximity ($r_\text{phys}$), and supply-chain position ($r_\text{supply}$).
\end{itemize}

This formalization enables systematic mapping between adversary profiles and feasible attack vectors, supporting risk-informed defense prioritization. We consider adversaries spanning three capability tiers: (1) \textit{Opportunistic attackers}, who operate remotely and exploit well-known CVEs, exposed services, and default credentials (e.g., misconfigured deployments, default passwords), requiring minimal robotics expertise. (2) \textit{Sophisticated attackers}, who have proximity-based access and can perform wireless injection, jamming, and sensor spoofing, with working knowledge of robotic protocols and stacks (e.g., ROS2/DDS, BLE provisioning, telemetry links). (3) \textit{Advanced persistent threats (APTs)}, who possess supply-chain or firmware-level access and detailed knowledge of target platform dynamics, representative of state-sponsored adversaries operating against military or critical-infrastructure deployments.

The trust-boundary model for quadruped teleoperation comprises four zones: (i) the operator environment (HMD, control station), (ii) the communication channel (wireless links, edge/cloud relay), (iii) the robot’s onboard computing (ROS2 graph, middleware, controllers), and (iv) the physical environment (sensors, actuators, terrain). Each boundary crossing constitutes an attack surface where cyber actions can induce physical consequences. We assume the adversary lacks direct physical access to the robot during operation, but can interact through wireless interfaces, cloud services, and the electromagnetic environment (e.g., RF injection and interference).

We then map STRIDE to the proposed six-layer taxonomy as follows. \textit{Spoofing} primarily corresponds to Layer 1 (sensor spoofing) and Layer 5 (localization spoofing). \textit{Tampering} aligns with Layer 4 (FDIA, command injection) and Layer 6 (firmware/stack modification). \textit{Repudiation} is captured under Layer 6 via logging and provenance manipulation. \textit{Information disclosure} maps to Layer 2 (operator/HMD side channels) and Layer 3 (eavesdropping over links). \textit{Denial of service} maps to Layer 3 (jamming and link-layer disruption) and Layer 6 (resource exhaustion and middleware-level disruption). Finally, \textit{Elevation of privilege} maps to Layer 4 (command-path hijacking) and Layer 6 (node impersonation and trust-anchor compromise).

\begin{figure*}[htbp]
\centering
\begin{tikzpicture}[
    mindmap,
    grow cyclic,
    every node/.style={concept, minimum size=0pt, text width=1.6cm, font=\scriptsize},
    root concept/.append style={concept color=red!50, fill=red!20, line width=1pt, minimum size=2.2cm, text width=2.2cm, font=\footnotesize\bfseries},
    level 1/.append style={level distance=3.2cm, sibling angle=60, concept color=blue!40, font=\scriptsize},
    level 2/.append style={level distance=2.2cm, sibling angle=45, concept color=green!30, font=\tiny, text width=1.3cm}
]
\node[root concept] {Quadruped Attack Taxonomy}
    child[concept color=purple!40] { node {L1: Visual/ Perception}
        child { node {LiDAR Spoofing} }
        child { node {Camera Manip.} }
        child { node {VO Attack} }
    }
    child[concept color=orange!40] { node {L2: VR/AR Specific}
        child { node {Cybersickness} }
        child { node {HMD Side-Ch.} }
        child { node {Display Inj.} }
    }
    child[concept color=cyan!40] { node {L3: Communi- cation}
        child { node {Jamming} }
        child { node {MITM} }
        child { node {Eavesdrop} }
    }
    child[concept color=yellow!60] { node {L4: Control Signal}
        child { node {FDIA} }
        child { node {Replay} }
        child { node {Hijacking} }
    }
    child[concept color=green!50] { node {L5: Localiz- ation}
        child { node {GPS Spoof} }
        child { node {Odometry} }
        child { node {Map Poison} }
    }
    child[concept color=red!40] { node {L6: Network/ System}
        child { node {DoS} }
        child { node {Node Spoof} }
        child { node {Supply Ch.} }
    };
\end{tikzpicture}
\caption{Six-layer attack taxonomy for quadruped robot teleoperation security. Each layer targets distinct system components with unique attack vectors. The taxonomy progresses from perception-level attacks (L1-L2) through communication and control (L3-L4) to infrastructure-level threats (L5-L6).}
\label{fig:attack_taxonomy}
\end{figure*}
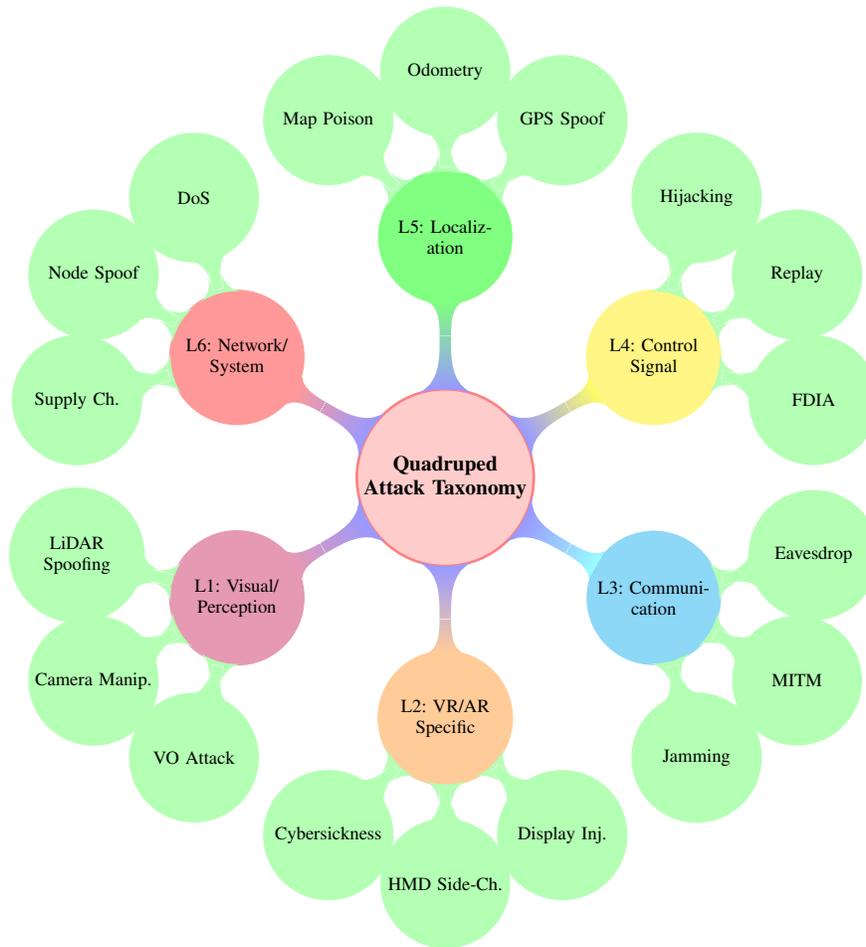

Table~\ref{tab:stride_mapping} formalizes this mapping. Fig.~\ref{fig:attack_taxonomy} provides a visual overview of the proposed attack taxonomy, summarizing the complete classification and its mapping across the identified layers.

\begin{table}[!t]
\caption{Mapping of STRIDE Threat Categories to the Six-Layer Attack Taxonomy}
\label{tab:stride_mapping}
\centering
\footnotesize
\begin{adjustbox}{max width=\columnwidth}
\begin{tabular}{l|cccccc}
\hline
\textbf{STRIDE Category} & \textbf{L1} & \textbf{L2} & \textbf{L3} & \textbf{L4} & \textbf{L5} & \textbf{L6} \\
 & \textbf{Percep.} & \textbf{Comms.} & \textbf{Comms.} & \textbf{Control} & \textbf{Operator} & \textbf{Platform} \\
\hline
Spoofing        & \checkmark &            &            &            & \checkmark &            \\
Tampering       &            &            &            & \checkmark &            & \checkmark \\
Repudiation     &            &            &            &            &            & \checkmark \\
Info.\ Disclosure &          & \checkmark & \checkmark &            &            &            \\
Denial of Service &          &            & \checkmark &            &            & \checkmark \\
Elev.\ of Privilege &        &            &            & \checkmark &            & \checkmark \\
\hline
\end{tabular}
\end{adjustbox}

\vspace{1mm}
\begin{flushleft}
\footnotesize L1: Perception Sensor Attacks; L2: Operator-Side Communication;
L3: Robot-Side Communication; L4: Control/Autonomy Manipulation;
L5: Operator Cognitive/Physiological; L6: Platform Infrastructure.
\end{flushleft}
\end{table}

\subsection{Layer 1: Visual and Perception Attacks}

Visual and perception attacks target a quadruped’s sensing pipeline—both onboard sensors and the video/telemetry streams delivered to the operator—degrading situational awareness, distorting intent, and ultimately steering human decision-making toward unsafe or suboptimal actions.

\subsubsection{LiDAR Spoofing and Manipulation}
LiDAR is a cornerstone sensor for quadruped navigation because it provides geometry that is largely invariant to lighting, enabling obstacle detection, free-space estimation, and support for SLAM/localization when GPS is unavailable. That same centrality makes LiDAR an unusually high-leverage attack surface: if an attacker can reliably create phantoms (false obstacles) or induce vanishing (remove real obstacles), they can shape both the robot’s autonomy stack and the operator’s decisions downstream.

Early demonstrations established feasibility with off-the-shelf hardware. Cao et al. reported that LiDAR spoofing on autonomous systems could succeed 75\% of the time using commercially available laser equipment, underscoring that practical adversaries do not necessarily require specialized lab infrastructure to cause meaningful perception failures \cite{cao2019lidar}. More recent work shows the attack space is not limited to adding artifacts—attackers can also subtract evidence. The Physical Removal Attack (PRA) achieved 92.7\% success in removing 90\% of obstacle point clouds from perception frameworks including Apollo and Autoware, illustrating that point-cloud sparsification can be engineered to look like a plausible sensing artifact rather than an overt anomaly \cite{cao2023pra}.

A parallel line of work attacks perception through adversarial physical objects rather than signal injection. MSF-ADV proposes 3D-printed objects engineered to be simultaneously inconspicuous to both camera and LiDAR, reporting 90\%+ attack success and 100\% collision rates in simulation—an especially concerning combination for legged robots operating near humans or in cluttered environments \cite{msfadv2020}. Importantly, defensive upgrades in newer LiDAR designs do not eliminate the problem. Sato et al. showed that even next-generation LiDAR mechanisms using timing randomization and pulse fingerprinting can still be coerced into object-vanishing behavior, successfully concealing five real vehicles from the sensor \cite{sato2024nextgenlidar}.

Not all effective LiDAR attacks are “active” in the electromagnetic sense. A particularly stealthy category is passive optical manipulation: strategically placed planar mirrors can inject phantom obstacles or conceal real objects without emitting electronic signatures, which makes detection difficult for conventional monitoring approaches that look for anomalous transmissions or spectrum activity \cite{mirrorlidar2025}. This matters operationally because passive attacks shift the defender’s job from detecting an emitter to detecting subtle environmental changes—often in exactly the kinds of public or semi-public spaces where quadrupeds are deployed.

Sensor fusion helps, but it is not a cure-all. Hallyburton et al. found that camera–LiDAR fusion can be inherently resilient to naive spoofing (with attack success below 5\%), yet sophisticated long-range tracking attacks remain effective, suggesting that attackers can adapt by targeting the fusion logic, temporal consistency, or tracking priors rather than the raw sensor stream alone \cite{hallyburton2022cameralidar}. Broad surveys reinforce a deeper, uncomfortable pattern: as LiDAR detection models become more capable, they may also become more fragile under adversarial pressure—high-performing models can be paradoxically more vulnerable to spoofing, raising the risk that “better perception” increases the payoff of targeted attacks \cite{aldairi2024lidarsurvey, aldairi2024dlattacks}.

For quadruped robots, LiDAR attacks become disproportionately dangerous during stair climbing and obstacle traversal, where centimeter-level depth accuracy directly governs foot placement and balance margins. A wheeled platform that encounters a perception inconsistency can often degrade gracefully—slow down, stop, or re-plan with relatively low immediate consequence. A quadruped, by contrast, is frequently committed to a dynamically stable motion: mid-stride on stairs, it has limited time and limited contact options to recover if the perceived terrain geometry is wrong.

Crucially, the failure mode is not merely “autonomy gets fooled.” In many real deployments, quadrupeds operate under shared control: semi-autonomous locomotion policies and foothold planners rely on a terrain model, while the operator relies on the same perception stack (or its derived video/overlays) to make supervisory decisions. A LiDAR attack therefore succeeds by corrupting the terrain representation that both the robot and the human implicitly trust—inducing a misstep, a slip, or a destabilizing foothold sequence. In this regime, the robot’s safety mechanisms can become counterproductive: an emergency stop during a stair ascent may arrest actuators at exactly the wrong phase of the gait, increasing the probability of a fall rather than preventing harm.

\subsubsection{Camera Feed Manipulation}
Video streams connecting robot cameras to operator displays are vulnerable to several high-impact attack vectors. Fake object insertion injects non-existent obstacles or targets into the stream, potentially provoking evasive operator commands that destabilize the robot. Obstacle removal selectively edits out real hazards, increasing collision risk. Video replay replaces the live feed with pre-recorded footage, allowing attackers to manipulate the robot or its surroundings while the operator observes benign scenes. Temporal desynchronization introduces variable delay and jitter, degrading operator perception–action coupling and inducing disorientation—especially when combined with rapid robot motion and head-tracked displays \cite{xu2025voattack}.

The DoubleStar attack (USENIX Security 2022) fabricates depth perception at ranges up to 15 meters at night and 8 meters in daytime by exploiting weaknesses in stereo matching for depth cameras \cite{doublestar2022}. This threat is particularly relevant to quadruped teleoperation, where operators often rely on stereoscopic feeds (and their derived depth cues) to judge clearance and traverse complex terrain. Corrupted depth cues do not merely reduce situational awareness; they introduce systematic distance-estimation bias, increasing the probability of misjudged footholds, mistimed commands, and unsafe body posture selection.

\subsubsection{Visual Odometry Corruption}
Visual odometry (VO) is frequently used to estimate ego-motion when GPS is unreliable or unavailable, and its integrity can be undermined through keypoint manipulation that disrupts feature tracking and degrades pose estimation. This failure mode is particularly consequential for legged systems because locomotion control depends on accurate short-horizon motion estimates to maintain stability margins during dynamic maneuvers. Xu et al. (ICRA 2025) demonstrated both attacks and defenses for VO, achieving 99.5\% detection success with sub-100ms latency via inconsistency analysis \cite{xu2025voattack}. Zhang et al. proposed complementary detection signals based on reprojection error statistics, feature descriptor stability, and motion prediction residuals \cite{zhang2025voattack}.

More broadly, physical-world adversarial attacks on camera-based perception have been extensively characterized in autonomous driving, spanning adversarial patches, scene manipulation, and camera-level interference \cite{chi2025adversarial}. Many of these techniques transfer directly to quadruped platforms. The critical difference is not feasibility but consequence severity: for a dynamically stable legged robot, small perception biases can cascade into balance loss or hazardous recovery behavior, whereas many wheeled platforms can often fail more gracefully by stopping or slowing down.

\subsection{Layer 2: VR/AR-Specific Attacks}

When teleoperation relies on immersive VR/AR interfaces, the human operator becomes part of the control loop—and therefore a legitimate target. These attacks exploit the unusually tight coupling between visual stimuli, vestibular perception (balance sense), and motor response in immersive environments. In other words: if an attacker can shape what the operator sees (and when they see it), they can shape what the operator does—even without touching the robot’s onboard stack.

\subsubsection{Cybersickness Induction Attacks}

Cybersickness induction attacks weaponize a fragile part of immersive teleoperation: the operator’s sensorimotor system. Cybersickness is a well-documented physiological response characterized by nausea, disorientation, and discomfort, with measurable correlates in prefrontal cortex activity and head-movement dynamics \cite{yalcin2024cybersickness, kourtesis2024csqvr}. A large ACM Computing Surveys synthesis of 223 studies identifies the most consistent inducing factors as latency and latency jitter, prolonged exposure, and rapid or unnatural visual-field motion; across this literature, sensory conflict theory remains the dominant explanatory framework \cite{biswas2024cybersicknessreview}. Because these symptoms translate directly into impaired decision-making and degraded manual control, NATO STO has issued mitigation guidance for military VR systems—most notably recommending refresh rates of at least 90 Hz and motion-to-photon latency below 20 ms \cite{nato2023cybersickness}.

Recent work shows that even defenses can be attacked. Kundu et al. (2025) demonstrated adversarial examples (FGSM, PGD, and C\&W) against deep-learning cybersickness detectors, causing a 4.65$\times$ drop in LSTM-based detection accuracy and a 5.94$\times$ drop in Transformer-based detection \cite{kundu2025cybersicknessattack}. This issue matters operationally: if an attacker can both induce symptoms and blind the monitoring layer, cybersickness becomes not merely a human-factors problem but a cyber-physical vulnerability.

Mechanistically, cybersickness can be induced through multiple deliberate vectors. Flicker injection uses high-frequency oscillations (8–25 Hz) to provoke adverse responses in susceptible individuals; vection manipulation introduces artificial optic-flow patterns that generate false self-motion perception; vestibular conflict is created by engineering a mismatch between visual cues and proprioceptive feedback; and field-of-view (FOV) perturbation dynamically changes perceived visual extent to destabilize spatial orientation \cite{stauffert2020latency}. Empirical evidence consistently indicates that longer exposure, higher apparent motion speeds, and joystick-based locomotion increase symptom severity, making many standard teleoperation interface choices risk-amplifying by default \cite{duzmanska2018factors, kim2022cybersickness, rebenitsch2021cybersickness}. To support controlled evaluation, De Pace et al. developed a standardized cybersickness testbed with four scenarios designed to elicit symptoms predictably, enabling more rigorous comparisons of mitigation strategies \cite{csqvrtestbed2024}.

The consequence pathway differs fundamentally from conventional technical attacks. Rather than corrupting the robot’s software stack, cybersickness attacks degrade the operator’s perceptual integrity and motor performance. An impaired operator tends to generate delayed, noisy, or overcorrective inputs—effects that are especially hazardous for quadruped teleoperation, where stability often depends on rapid micro-adjustments during contact transitions and terrain changes. In this regime, destabilization can occur even when the robot’s autonomy and communication channel remain uncompromised: the “attack surface” is the human-in-the-loop controller.

Formal threat modeling has begun to treat this class explicitly. Valluripally et al. (2021) introduced attack-fault tree formulations for cybersickness-inducing threats, providing a structured basis for analyzing operator-targeted mechanisms and their causal chains. \cite{valluripally2021cybersickness}. On the defensive side, Yalcin et al. (2024) reported real-time cybersickness detection achieving a 91.1\% F1-score using bidirectional LSTM models with data augmentation, illustrating both the practicality of automated monitoring and the extent to which cybersickness can be operationalized as a detectable safety-and-security hazard \cite{yalcin2024cybersickness}.

\subsubsection{HMD Side-Channel Attacks}
Zhang et al. (USENIX Security 2023) shown that head-mounted displays (HMDs) can inadvertently function as sophisticated side-channel sensors, disclosing sensitive information across many quantifiable channels \cite{zhang2023vr}. Specifically, they demonstrated (i) keystroke inference through head-motion dynamics, attaining 82\% top-5 word accuracy; (ii) application identification via head-movement “fingerprints”; (iii) bystander distance estimation with a mean absolute error of 10.3 cm; and (iv) hand-gesture recognition surpassing 90\% accuracy. In teleoperation scenarios, these leakage channels extend beyond privacy issues; they can disclose mission-critical context and operational intentions, as well as expose authentication credentials input during control sessions—data that facilitates subsequent compromise. 

\subsubsection{Display Content Injection}

Display content injection attacks distort the operator's perception of reality. The attacker, rather than manipulating the robot's controls, focuses on disrupting the human decision-making process by introducing overlays that mimic authentic telemetry, such spurious environmental annotations, counterfeit status indications, or misleading navigation cues \cite{casey2021immersive}. The threat is epistemological rather than computational—operators may struggle to differentiate between malicious overlays and genuine UI elements, especially when under time constraints, cognitive burden, or compromised network conditions.

In quadruped teleoperation, the consequence profile is intensified by the platform's dynamic stability limitations. Injected waypoints might direct the robot into terrain transitions necessitating meticulous foot placement; erroneous battery or heat alerts may provoke hasty withdrawals or perilous "rush" actions; and altered pose/velocity readings can skew operator control towards destabilizing directives. In summary, content injection is effective by undermining situational awareness, and quadrupeds rectify awareness problems more swiftly than numerous other robotic categories.

\subsection{Layer 3: Communication Layer Attacks}

Communication-layer attacks aim at the wireless and network connections linking the operator, the robot, and any relay infrastructure, posing the greatest immediate threat to teleoperation continuity. In contrast to perceptual or UI attacks, which diminish decision quality, communication attacks compromise the control channel, facilitating interruption, deterioration, or usurpation of command-and-feedback loops.

\subsubsection{Jamming Attack Variants}
Pirayesh and Zeng’s comprehensive 2022 survey reports that constant jamming can drive packet errors to effectively 100\% when the signal-to-jamming ratio (SJR) drops below 4 dB for WiFi 18 Mbps transmissions. \cite{pirayesh2022jamming}. Building on Diller et al.’s taxonomy \cite{diller2023jamming}, the main jamming variants can be structured as follows: constant jamming, where continuous wideband noise saturates the channel; reactive jamming, where interference is triggered only when legitimate transmissions are detected to conserve attacker power and reduce detectability; deceptive jamming, which injects spoofed or malformed packets that consume airtime while appearing protocol-compliant; frequency-sweep jamming, which scans across channels to create broadband disruption; and pilot jamming, which targets reference signals used for channel estimation in OFDM systems, corrupting demodulation even when raw energy is not maximized.

For quadruped teleoperation, the hazard profile of jamming is uniquely severe because communication loss occurs mid-dynamics. A wheeled platform often fails “gracefully” by braking to a halt; a quadruped on stairs, rubble, or a narrow ledge may require continuous closed-loop correction to preserve balance and foot placement. Consequently, the impact of jamming is not only a function of channel metrics (SJR, duty cycle, bandwidth) but also of the robot’s locomotion phase at attack onset: the same dropout that is survivable during stance can be catastrophic during swing or during a COM transition. This implies that classical jamming literature—often developed around statically stable or quasi-static platforms—misses a key state-dependent coupling between cyber disruption and physical stability.

Recent anti-jamming defenses increasingly move beyond fixed hopping rules toward learning-based adaptation, particularly deep reinforcement learning (DRL) for dynamic frequency selection. Multi-agent RL approaches allow terminals to sense spectrum conditions and coordinate channel choices using Q-learning-style policies with primary/backup allocation rules, improving transmission success across diverse jamming strategies \cite{antijarl2024}. In parallel, UAV and ground-robot studies have explored two-level anti-jamming architectures that combine higher-level coordination (e.g., group-level instructions) with local optimization such as beamforming or link adaptation \cite{wanguavswarm2024}. For quadrupeds, the research gap is translating these methods into locomotion-aware communication policies—where the radio adapts not just to spectrum conditions but to stability-critical moments (e.g., prioritizing ultra-reliable low-latency control packets during gait transitions and deferring nonessential payloads).

\subsubsection{Man-in-the-Middle (MITM) Attacks}
Man-in-the-middle attacks facilitate command alteration, data injection, and session takeover. Finn and Santoso (IEEE TDSC 2023) achieved a 99\% detection success rate with deep learning neural networks on US Army GVT-BOT replicas; nevertheless, unprotected systems experienced total compromise \cite{finn2023mitm}. Man-in-the-middle attacks pose significant risks to teleoperation by covertly altering commands while preserving the illusion of communication integrity.

\subsubsection{Eavesdropping and Data Exfiltration}
Unencrypted ROS topics expose sensitive data including camera feeds, localization information, and control commands. Abeykoon et al. developed ROSploit demonstrating practical exploitation of ROS1 vulnerabilities \cite{abeykoon2018rosploit}. The Unitree vulnerability disclosure revealed that telemetry data could be intercepted and decoded using fleet-wide hardcoded keys, enabling passive surveillance of any deployed robot \cite{unipwn2025spectrum}. The Alias Robotics research identified over 643 distinct public-facing DDS services exposed to the internet, many leaking private IP addresses and internal network architecture details \cite{aliasrobotics2022dds}.

\subsection{Layer 4: Control Signal Attacks}

Control signal attacks immediately alter the command stream between the operator and the robot, transforming the control channel into the weapon. 

\subsubsection{False Data Injection Attacks (FDIA)}
Dong et al. (IEEE TCST 2020) presented a destabilizing FDIA framework demonstrating that an assailant may disrupt robot joint velocity regulation via meticulously designed signal injection \cite{dong2020fdia}. Instead of overtly "breaking" the controller, the approach manipulates the feedback loop's structure to induce error amplification while maintaining sufficient consistency to bypass conventional residual-based detection methods. More sophisticated variants utilize Lie group formulations to establish conditions under which a False Data Injection Attack (FDIA) can modify follower motion while the operator observes nominal behavior, including situations that exploit malleability characteristics in encrypted control pipelines \cite{liegroup2024fdia}. Kwon and Ueda \cite{kwon2024fdia} expanded this research to bilateral teleoperation including second-order nonlinear manipulator dynamics, establishing Lie-group-based criteria for \textit{perfectly undetectable} FDIA. By experimentally confirming feasibility across a US--Japan teleoperation link, it is demonstrated that these attacks can endure under realistic networking conditions.

The repercussions for quadrupeds are exacerbated by the close relationship between command time and stability. An FDIA that implements minor, structured alterations in gait-phase signals (e.g., swing timing, touchdown events, or inter-leg coordination) may appear visually inconspicuous while progressively leading to balance impairment—particularly during transitions such as stair climbing, turning, or recovering from slight slips. 

\subsubsection{Replay Attacks}
Replay attacks, as described by Mo and Sinopoli \cite{mo2009replay}, involve the recording of authentic control and measurement sequences, which are subsequently retransmitted, enabling an adversary to "freeze" the operator's perception of the system while the physical robot deviates. Empirical evidence indicates that standard detection methods based on $\chi^2$ residual tests achieve only 35--75\% detection rates at a 5\% false-alarm rate in typical robotic environments \cite{dieber2020penetration}, highlighting significant vulnerability. In teleoperation, replaying previous operator commands may lead to the repeated execution of maneuvers that were safe in the recorded context but perilous in the current environment, or it may inhibit urgent corrective actions by obscuring real-time deviations.

\subsubsection{Command Injection and Hijacking} Command injection and hijacking involve the unauthorized alteration, insertion, or substitution of operator commands. Consequences vary from minor disturbances, such as enforced emergency halts due to denial-of-service attacks, to critical safety manipulations, including trajectory redirection into perilous zones, joint velocity saturation that destabilizes movement, and parameter corruption that alters locomotion dynamics beyond validated operational limits. At the AI-control interface, RoboPAIR attained a 100\% success rate in jailbreaking Unitree Go2 systems, facilitating detrimental actions via adversarial urging \cite{robson2024robopair}. The Ghost Robotics CVE-2025-41108 disclosure reveals that the control protocol at the protocol layer lacks authentication, allowing for direct command hijacking without necessitating cryptographic compromise \cite{incibe2025vision60}.

\subsubsection{Delay Injection Attacks}
The deliberate introduction of latency diminishes teleoperation quality and may provoke operator cybersickness. Research demonstrates that delays over 200ms substantially hinder operator performance, whilst delays exceeding 500ms render precise manipulation unfeasible \cite{risiglione2021passivity, hamdan2021teleop}. Stauffert et al. demonstrated that latency jitter, despite an acceptable mean latency, induces considerable cybersickness in virtual reality environments \cite{stauffert2020latency}.

\subsection{Layer 5: Localization and Navigation Attacks}

\subsubsection{GPS/GNSS Spoofing}
Four advanced attack categories have been exhibited against GPS-dependent robots: turn-by-turn manipulation causing gradual trajectory deviation, overshoot attacks that propel the platform beyond intended waypoints, wrong-turn injection that redirects it onto erroneous paths, and stop spoofing that inaccurately indicates arrival at the destination. Bianchin et al. rigorously established conditions for mathematically imperceptible GNSS spoofing through a game-theoretic framework.\cite{bianchin2019undetectable}. Zhang et al., using the Motion-Sensitive Analysis Framework (MSAF), showed that the target’s dynamic state—especially acceleration and high-speed cruising—can substantially amplify spoofing effectiveness, reporting 82\% success for off-road deviation attacks \cite{ghostnavigator2025}. More recent results indicate that even IMU-based spoofing detectors can be neutralized: external IMU sensors and feedback-based spoofing can be used to track and counteract the target’s internal EKF innovations, effectively “shadowing” the estimator \cite{gps2024neutralization}. On the defense side, the GPS-IDS framework combines physics-based vehicle models with EKF sensor fusion for anomaly-based detection \cite{gpsids2024}, complemented by earlier sensor fusion-based GNSS spoofing detection approaches for autonomous vehicles \cite{gpsfusion2021}, while multi-channel GNSS receivers paired with machine-learning classifiers provide emerging detection capability in UAV contexts \cite{mouzai2025gpsuavml, infocom2024gpsuav}. Recent work further demonstrates that GPS spoofing can degrade AI-based navigation with obstacle avoidance in UAV systems, confirming the transferability of this threat to autonomous robotic platforms \cite{gpsspoofuavdrl2025}. A broad mitigation survey identifies multi-sensor fusion (GNSS/IMU/LiDAR) as the most promising direction, but notes the absence of quadruped-specific implementations \cite{gps2024pmc}.

\subsubsection{Odometry Corruption}
Attacks on proprioceptive sensing degrade dead-reckoning performance and are especially consequential in GNSS-denied environments. Unlike wheeled robots—where wheel encoders dominate odometry—quadrupeds rely on leg odometry (contact timing, kinematic constraints, joint sensing) tightly integrated with IMU-based state estimation. This shifts the attack surface toward proprioceptive channels: perturbing IMU measurements, biasing joint encoders, or manipulating contact detection can induce systematic drift in the estimated base pose and velocity, degrading both navigation and balance-critical control.

\subsubsection{Map and Landmark Poisoning}
Map poisoning targets the integrity of SLAM back-ends and long-lived navigation priors. By corrupting stored maps or injecting false landmarks, an attacker can induce persistent localization and planning errors that outlive the initial intrusion. The risk compounds in fleet settings: poisoned artifacts may propagate through shared map repositories, causing repeated navigation failures across missions and platforms operating in the compromised region.

\subsection{Layer 6: Network and System Attacks}

\subsubsection{Denial of Service (DoS)}
De Persis and Tesi demonstrated via Input-to-State Stability (ISS) analysis  that control systems may endure maximum Denial of Service (DoS) duty cycles  below a system-dependent threshold (illustratively on the order of a few  percent for certain controller configurations) while preserving stability  \cite{depersis2015dos}.

\subsubsection{ROS Topic Flooding}
Overwhelming specific topics (e.g., /cmd\_vel, /joint\_states) with spurious messages degrades system performance and can trigger safety shutdowns. The ROSPaCe dataset provides labeled attack scenarios including topic flooding, enabling training of detection systems \cite{puccetti2024rospace}.

\subsubsection{Node Impersonation}
Utilizing ROS2 DDS discovery mechanisms to introduce rogue nodes that impersonate genuine system components facilitates command injection and data exfiltration. The flaws in the discovery protocol of SROS2 (V3) reveal network topology to passive attackers, enabling subsequent impersonation assaults \cite{deng2022insecurity}.

\subsubsection{Firmware and Supply Chain Attacks}
Compromising firmware update methods or development toolchains creates enduring backdoors. The UniPwn attack revealed that the BLE provisioning daemon of Unitree robots operates with root privileges and receives arbitrary orders over the WiFi credential injection channel, facilitating a persistent compromise that can autonomously disseminate across adjacent robots \cite{unipwn2025spectrum}. Mayoral-Vilches et al. reported that the over-the-air update procedures in the Unitree G1 are susceptible to manipulation by an attacker with network access \cite{mayoral2025unitreeg1}. The 2025 SROS2 supply chain attack demonstration revealed that compromised packages disseminated via conventional channels could exfiltrate credentials undetected \cite{sros2supplychain2025}. Espionage activities by nation-states aimed at the robotics sector have been monitored since late 2024, with Russia's DcRAT, AsyncRAT, and XWorm malware identified as targeting robotics manufacturers and their supply chains \cite{recordedfuture2025robotics}.

\begin{table*}[htbp]
\caption{Comprehensive Attack Taxonomy for Quadruped Robot Teleoperation Communication Systems}
\label{tab:attack_taxonomy}
\centering
\footnotesize
\begin{tabular}{|p{2.0cm}|p{2.5cm}|p{3.5cm}|p{2.0cm}|p{2.0cm}|p{2.5cm}|}
\hline
\textbf{Attack Layer} & \textbf{Attack Type} & \textbf{Mechanism} & \textbf{Target Component} & \textbf{Success Rate} & \textbf{Key References} \\
\hline
\multirow{4}{*}{Visual/Perception} 
& LiDAR Spoofing & Laser injection/relay & 3D perception & 75--92\% & \cite{cao2019lidar, cao2023pra} \\
& Camera Manipulation & Video injection/editing & Operator view & High & \cite{doublestar2022} \\
& Visual Odometry Attack & Keypoint manipulation & Ego-motion estimation & Variable & \cite{zhang2025voattack} \\
& Depth Perception Attack & Stereo matching exploit & Distance estimation & 80\%+ & \cite{doublestar2022, msfadv2020} \\
\hline
\multirow{4}{*}{VR/AR-Specific}
& Cybersickness Induction & Flicker/vection manipulation & Human operator 
& High$^\dagger$ & \cite{kundu2025cybersicknessattack} \\
& HMD Side-Channel & Motion/gesture analysis & Operator privacy & 82--90\% & \cite{zhang2023vr} \\
& Display Injection & Content overlay & Operator perception & High & \cite{zhang2023vr, casey2021immersive} \\
& Latency Manipulation & Delay injection & Motion-to-photon & 100\% & \cite{stauffert2020latency} \\
\hline
\multirow{4}{*}{Communication}
& Constant Jamming & Continuous noise & Wireless channel & 100\% (SJR$<$4dB) & \cite{pirayesh2022jamming, diller2023jamming} \\
& Reactive Jamming & Triggered interference & Active transmissions & High & \cite{diller2023jamming} \\
& MITM Attack & Session interception & Data/commands & High (undefended) & \cite{finn2023mitm} \\
& Eavesdropping & Passive monitoring & Confidentiality & 100\% (unencrypted) & \cite{abeykoon2018rosploit} \\
\hline
\multirow{4}{*}{Control Signal}
& False Data Injection & Sensor/actuator manipulation & Control loop & Variable & \cite{dong2020fdia, liegroup2024fdia} \\
& Replay Attack & Signal recording/replay & Command sequence & 65\% evade det. & \cite{mo2009replay, dieber2020penetration} \\
& Command Hijacking & Unauthorized injection & Actuators & 100\% (RoboPAIR) & \cite{robson2024robopair} \\
& Delay Injection & Latency introduction & Real-time control & 100\% & \cite{risiglione2021passivity, hamdan2021teleop} \\
\hline
\multirow{3}{*}{Localization}
& GPS Spoofing & False satellite signals & Position estimation & High & \cite{bianchin2019undetectable} \\
& Odometry Corruption & Encoder/IMU manipulation & Dead reckoning & Variable & -- \\
& Map Poisoning & SLAM database corruption & Navigation & Persistent & \cite{xu2025voattack} \\
\hline
\multirow{4}{*}{Network/System}
& DoS Attack & Resource exhaustion & System availability & $>$3\% duty cycle & \cite{depersis2015dos} \\
& Topic Flooding & Message spam & ROS communication & High & \cite{deng2022insecurity, puccetti2024rospace} \\
& Node Impersonation & DDS discovery exploit & System integrity & Medium & \cite{deng2022insecurity} \\
& Supply Chain & Firmware/infrastructure & Persistent access & Demonstrated & \cite{sros2supplychain2025, unipwn2025spectrum} \\
\hline
\end{tabular}

\vspace{2mm}
\scriptsize{\textbf{Note:} Success rates indicate demonstrated effectiveness under study-specific experimental conditions and should not be interpreted as operational guarantees. Key methodological variations across cited studies include: (i) target platform heterogeneity (automotive LiDAR vs. robotic sensors), (ii) environmental conditions (laboratory vs. outdoor, static vs. dynamic targets), (iii) attacker positioning and equipment constraints, and (iv) detection/defense baseline configurations. Rates marked ``High'' without quantification indicate consistent success across multiple experimental trials without systematic failure modes; ``Variable'' indicates sensitivity to experimental parameters that may differ substantially in quadruped deployment contexts. Transfer of these rates to quadruped teleoperation requires domain-specific validation accounting for gait-induced sensor motion, platform-specific sensor configurations, and operational environment characteristics. ``High (undefended)'' indicates near-certain success against systems without specific countermeasures. ``Variable'' indicates context-dependent success rates. \\
$^\dagger$~Success rate depends on individual susceptibility and exposure duration.}
\end{table*}

To facilitate interpretation of reported success rates, Table~\ref{tab:attack_transferability} summarizes the experimental conditions under which key attack metrics were obtained and provides preliminary assessment of transferability to quadruped teleoperation contexts.

\begin{table}[htbp]
\caption{Experimental Conditions for Reported Attack Success Rates}
\label{tab:attack_transferability}
\centering
\footnotesize
\begin{adjustbox}{max width=\columnwidth}
\begin{tabular}{|p{2.0cm}|p{1.3cm}|p{1.8cm}|p{2.2cm}|p{2.2cm}|}
\hline
\textbf{Attack Type} & \textbf{Rate} & \textbf{Platform} & \textbf{Conditions} & \textbf{Quadruped Transfer} \\
\hline
LiDAR Spoofing & 75--92\% & Automotive & Static target, lab & Likely lower (gait motion) \\
\hline
Depth Camera & 80\%+ & Stereo camera & 8--15m, night & Similar expected \\
\hline
VO Detection & 99.5\% & Generic VO & Indoor, stable & Unknown (vibration) \\
\hline
Cybersickness & High & VR headsets & Lab, 10--30 min & Similar expected \\
\hline
FDIA Evasion & 65\% & Simulation & Known model & Context-dependent \\
\hline
GPS Spoofing & 82\% & Ground vehicle & Outdoor, dynamic & Similar expected \\
\hline
\end{tabular}
\end{adjustbox}

\vspace{2mm}
\scriptsize{\textbf{Note:} ``Quadruped Transfer'' indicates expected change when applied to quadruped teleoperation. Gait-induced motion and vibration may increase or decrease attack effectiveness.}
\end{table}

\subsection{Attack-to-Consequence Mapping}
\label{sec:impact_analysis}

The aforementioned taxonomy delineates the technical methods by which attackers can infiltrate quadruped teleoperation systems. A thorough security assessment necessitates a methodical evaluation of the repercussions that arise when these attacks are successful. In contrast to solely digital systems, where security breaches lead to data loss or service interruptions, attacks on teleoperated quadruped robots have tangible repercussions that can severely impact human safety, mission integrity, infrastructure, and strategic outcomes.

\subsubsection{Physical Impacts on Robot Platform}

Quadruped robots are most vulnerable during dynamic locomotion phases where stability depends on tight timing margins and high-bandwidth coordination across legs. Zhang et al. demonstrated fall prediction with 95\% accuracy and 395 ms advance warning, and further showed that unmitigated falls can produce impact velocities roughly 3× higher than controlled descent \cite{zhang2024fall}. This matters for security because many cyber-physical attacks manifest as timing faults: for example, a delay-injection attack that introduces ~200 ms latency during the swing-to-stance transition can shift touchdown timing enough to cause a missed foothold, triggering a cascading loss of balance. For a ~32 kg-class quadruped such as Boston Dynamics Spot operating on industrial steel stairs, the resulting impacts can reach the kilonewton range (order-of-magnitude), which is sufficient to crack sensor housings, deform leg linkages, and induce secondary failures in onboard computing or connectors \cite{boston2024security}.

Perception attacks carry particularly severe physical consequences because they corrupt the robot’s environmental model without necessarily producing an immediate “something is wrong” signal for the operator or autonomy stack. LiDAR spoofing and perception-removal attacks demonstrated by Cao et al. achieve 75–92\% success in suppressing obstacle representations within common perception pipelines \cite{cao2019lidar, cao2023pra}. In dense industrial environments—pipes, guardrails, rotating machinery, cable trays—removing even a single salient obstacle from the occupancy grid can produce a “clean” path in planning, resulting in a direct collision that is not a simple inconvenience but a high-energy contact event with compounding effects (damage → degraded sensing → worse state estimation → higher fall likelihood).

\subsubsection{Human Safety Consequences}

The most immediate safety concern is direct physical harm from a compromised quadruped operating near people. Military and law-enforcement deployments increasingly involve payloads such as manipulator arms and inspection kits, and in some configurations, weaponized or weapons-adjacent systems \cite{boston2024security}. Under that operational reality, the critical vulnerability described in CVE-2025-41108—complete absence of authentication in the control protocol—has a straightforward implication: an adversary who can assume control can convert the platform from a sensing asset into a kinetic hazard.

Operator-targeted assaults raise the damage potential significantly. Cybersickness induction assaults do not require the penetration of the robot's control loop; rather, they impair the human's capacity to securely close the loop. Systematic reviews indicate that 60–80\% of virtual reality users encounter varying degrees of cybersickness, with symptoms potentially lasting 1–5 hours post-exposure in clinical settings \cite{kim2022cybersickness, rebenitsch2021cybersickness, duzmanska2018factors}. In mission profiles necessitating prolonged focus—such as multi-hour reconnaissance, inspection, or remote intervention—a brief hostile stimulus at the onset of the operation can result in a persistent performance detriment over the balance of the job duration.

\subsubsection{Mission-Critical and Infrastructure Consequences}

Quadruped deployments often sit inside time-critical, high-consequence workflows—EOD support, search-and-rescue, hazardous inspection, perimeter reconnaissance—where mission failure costs can exceed the platform’s price by orders of magnitude. In these settings, the attacker’s goal need not be dramatic; small degradations in reliability (intermittent delay, periodic sensor corruption, selective link disruption) can be sufficient to force aborts, delay response, or create unsafe uncertainty.

Finally, the threat model escalates from single-robot compromise to fleet-level systemic risk when propagation is possible. The botnet formation capability demonstrated at GEEKCon 2025—where compromised Unitree robots spread attacks to nearby units via Bluetooth—introduces an operationally distinct failure mode: a facility deploying multiple quadrupeds can experience a “contagion” dynamic in which compromise probability increases with density and proximity, turning local wireless exposure into a swarm-scale hazard \cite{unipwn2025spectrum}.

\subsection{Cascading Failure Pathways}

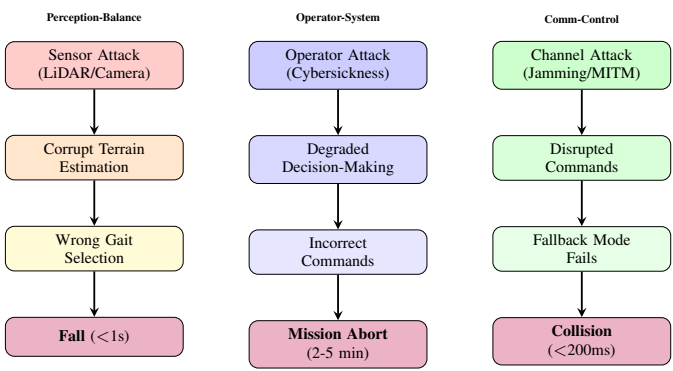
\begin{figure}[htbp]
\centering
\begin{adjustbox}{max width=\columnwidth}
\begin{tikzpicture}[
    node distance=0.7cm,
    box/.style={rectangle, draw, rounded corners, text width=2.5cm, minimum height=0.6cm, align=center, font=\scriptsize},
    arrow/.style={->, thick, >=stealth}
]
\node[box, fill=red!20] (p1) {Sensor Attack\\(LiDAR/Camera)};
\node[box, fill=orange!20, below=of p1] (p2) {Corrupt Terrain\\Estimation};
\node[box, fill=yellow!20, below=of p2] (p3) {Wrong Gait\\Selection};
\node[box, fill=purple!30, below=of p3] (p4) {\textbf{Fall} ($<$1s)};

\draw[arrow] (p1) -- (p2);
\draw[arrow] (p2) -- (p3);
\draw[arrow] (p3) -- (p4);

\node[box, fill=blue!20, right=1.0cm of p1] (o1) {Operator Attack\\(Cybersickness)};
\node[box, fill=blue!15, below=of o1] (o2) {Degraded\\Decision-Making};
\node[box, fill=blue!10, below=of o2] (o3) {Incorrect\\Commands};
\node[box, fill=purple!30, below=of o3] (o4) {\textbf{Mission Abort}\\(2-5 min)};

\draw[arrow] (o1) -- (o2);
\draw[arrow] (o2) -- (o3);
\draw[arrow] (o3) -- (o4);

\node[box, fill=green!20, right=1.0cm of o1] (c1) {Channel Attack\\(Jamming/MITM)};
\node[box, fill=green!15, below=of c1] (c2) {Disrupted\\Commands};
\node[box, fill=green!10, below=of c2] (c3) {Fallback Mode\\Fails};
\node[box, fill=purple!30, below=of c3] (c4) {\textbf{Collision}\\($<$200ms)};

\draw[arrow] (c1) -- (c2);
\draw[arrow] (c2) -- (c3);
\draw[arrow] (c3) -- (c4);

\node[above=0.15cm of p1, font=\tiny\bfseries] {Perception-Balance};
\node[above=0.15cm of o1, font=\tiny\bfseries] {Operator-System};
\node[above=0.15cm of c1, font=\tiny\bfseries] {Comm-Control};
\end{tikzpicture}
\end{adjustbox}
\caption{Three primary cascading failure pathways in quadruped teleoperation. Perception-Balance cascades operate in sub-second timeframes, Operator-System cascades develop over minutes, and Communication-Control cascades can be near-instantaneous during dynamic locomotion.}
\label{fig:cascade}
\end{figure}

Security incidents in quadruped teleoperation rarely manifest as isolated events. Our analysis identifies three primary cascading failure pathways. Fig.~\ref{fig:cascade} illustrates these pathways visually, and Table~\ref{tab:attack_chains} provides concrete examples with quantitative timing.

\begin{table*}[htbp]
\caption{Attack Chain Examples: From Initial Attack to Terminal Consequence}
\label{tab:attack_chains}
\centering
\footnotesize
\begin{tabular}{|p{2.2cm}|p{2.5cm}|p{2.5cm}|p{3.0cm}|p{1.8cm}|p{2.5cm}|}
\hline
\textbf{Initial Attack} & \textbf{Primary Effect} & \textbf{Secondary Effect} & \textbf{Terminal Consequence} & \textbf{Time-to-Conseq.} & \textbf{Intervention Points} \\
\hline
LiDAR Spoofing & Phantom obstacle or removal & Emergency stop on stairs / collision & Fall + platform damage & $<$500ms & Multi-sensor fusion, gait-phase protection \\
\hline
Cybersickness Induction & Operator disorientation & Delayed/wrong commands & Mission abort + operator incapacitation & 2--5 min & Real-time symptom detection, FOV adaptation \\
\hline
GPS Spoofing & Wrong position estimate & Path deviation & Entry to hazardous zone / mission failure & 10--60s & IMU/GPS consistency, VO fallback \\
\hline
MITM on Video & Delayed visual feedback & Operator overcorrection & Collision with obstacle & 1--3s & End-to-end encryption, latency monitoring \\
\hline
Control Command Injection & Forced gait transition & Loss of balance & Fall + potential human injury & $<$100ms & Command authentication, bounded control \\
\hline
Jamming during stairs & Communication loss & Autonomous fallback fails & Uncontrolled fall & $<$200ms & Multi-link redundancy, safe-stance mode \\
\hline
FDIA on joint velocity & Subtle timing errors & Gradual drift to instability & Fall several cycles later & 2--5s & State estimator residual monitoring \\
\hline
Supply chain backdoor & Persistent access & Data exfiltration + command capability & Technology transfer / coordinated attack & Days--months & Firmware verification, network isolation \\
\hline
\end{tabular}

\vspace{2mm}
\scriptsize{\textbf{Note:} Immediate (<1s): physical stability cascades; Short-term (1s--5min): operator/mission degradation; Persistent (hours--months): strategic/supply-chain effects. \textbf{Estimate Types:} (E) evidence-derived from cited studies; (I) author-inferred from engineering analysis; (S) speculative extrapolation.}
\end{table*}

\textbf{Perception–Balance Cascade.} Sensor-layer attacks compromise terrain and obstacle estimate, therefore skewing foothold selection and gait parameterization. The discrepancy between the presumed contact geometry and the actual surface dynamics destabilizes the locomotion controller, elevating the likelihood of slippage and precipitating loss-of-balance incidents, potentially culminating in a complete fall. This cascade occurs on sub-second timeframes (typically $<1$ s), allowing minimal opportunity for human intervention. Key intervention points are included (i) in the sensor-fusion and state-estimation framework (consistency verification, multi-modal cross-validation, fault isolation) and (ii) in the gait controller (resilient contact management, conservative gait fallback, swift disturbance rejection).

\textbf{Operator–System Cascade.}  Operator-targeted assaults impair perception, cognition, or physiological preparedness (e.g., disorientation, attentional capture), resulting in delayed, erratic, or systematically distorted directives. In contrast to solely technical cascades, the failure mechanism disseminates through the human-in-the-loop control pathway: impaired situational awareness $\rightarrow$ diminished decision quality $\rightarrow$ hazardous command sequences $\rightarrow$ heightened likelihood of destabilizing maneuvers. This cascade generally develops over extended periods (seconds to minutes), generating significant opportunities for automated detection (operator-state monitoring, telemetry plausibility assessments, intent-consistency evaluations) and protective measures (rate limiting, command validation, shared autonomy restrictions, enforced stabilization modes).

\textbf{Communication–Control Cascade.} Communication-layer assaults (jamming, delay, packet manipulation, replay) compromise the integrity and promptness of command/telemetry transmission. The immediate consequence is not solely the forfeiture of operator authority but also the activation of autonomous fallback behaviors (such as hold position, return-to-home, local navigation, and “safe stop”) that may be incompatible with the existing contact state or terrain (e.g., stair traversal, narrow footholds, compliant ground). The sequence is as follows: channel disruption leads to diminished closed-loop observability and controllability. Inappropriate mode transition may lead to instability or mission failure. Critical intervention points are located (i) at the communications boundary (link-quality attestation, anti-jam adaptation, authenticated sequencing, delay/replay detection) and (ii) within the robot controller (state-aware fallback selection, stability-preserving degraded modes, safe recovery primitives instead of generic stop behaviors).

\subsubsection{Cross-Layer Coordinated Attack Scenario}

While the preceding cascading pathways describe how single attack vectors propagate through layered subsystems, a capable adversary can do something nastier: coordinate multi-layer actions that exploit inter-layer coupling to generate outcomes no single vector could reliably produce. In other words, the attacker stops “poking one component” and starts conducting the system—timing small perturbations across perception, human-in-the-loop control, and actuation to force a convergent failure.

Consider a representative scenario targeting a Unitree Go2 used for industrial inspection. The adversary first leverages the BLE provisioning weakness (Layer 6) to gain persistent access to the robot’s internal network. Rather than acting immediately, the attacker deploys a dormant payload that passively monitors operational state and waits for a high-leverage moment—e.g., stair ascent, where stability margins are thin and recovery options are constrained. At the critical phase of locomotion, the adversary triggers a coordinated three-part maneuver: (1) a subtle LiDAR point-cloud manipulation that removes the representation of a stairway railing (Layer 1), biasing the perceived traversable corridor; (2) injection of ~150 ms additional latency into the video feed delivered to the operator’s VR interface (Layer 2), degrading situational awareness and closed-loop reaction time; and (3) a low-magnitude FDIA on joint-velocity commands (Layer 4), introducing timing perturbations at the swing-to-stance transition that might be tolerable in isolation but become destabilizing when combined with corrupted terrain cues.

The key insight is that each component is intentionally engineered to remain below the alert thresholds of single-layer defenses. The LiDAR manipulation targets a peripheral structure rather than the primary stair geometry; the latency increase can masquerade as benign network jitter; and the FDIA magnitude can be chosen to sit within the envelope of nominal disturbances and sensor noise. Yet their joint effect is multiplicative: the robot’s world model becomes slightly overconfident, the operator becomes slightly slower and less certain, and the gait controller is nudged toward a marginally mistimed contact sequence—together producing a high-probability fall precisely when consequence severity is maximal.

This motivates a defense posture that is explicitly cross-layer: not merely better detectors within each layer, but correlation mechanisms that treat weak signals as meaningful when they co-occur with the right timing and dependency structure. Concretely, this points toward cross-layer anomaly correlation systems that fuse sub-threshold deviations—perception inconsistencies, operator-interface latency shifts, and control-loop residual anomalies—into a joint statistical test (or graph-based dependency model) capable of flagging coordinated attacks that would remain invisible to siloed monitoring.

\subsection{Consequence Severity Framework}

The severity ratings in Table~\ref{tab:consequence_severity} are assigned based on a structured multi-criteria framework integrating four analytical dimensions. First, \textit{reversibility} quantifies whether the consequence can be undone once the attack ceases: Low indicates full reversibility within minutes, Medium indicates reversibility requiring hours to days of remediation, High indicates partial irreversibility with permanent degradation, and Critical indicates complete irreversibility (e.g., human injury, platform destruction). Second, \textit{scope of propagation} measures whether impact remains confined to the immediate target (Low), extends to adjacent systems or personnel in the operational area (Medium/High), or propagates to infrastructure, strategic assets, or fleet-wide compromise (Critical). Third, \textit{temporal persistence} distinguishes immediate resolution upon detection (Low), resolution within the mission window (Medium), persistence beyond the mission requiring explicit remediation (High), and indefinite persistence enabling follow-on attacks (Critical). Fourth, \textit{detection difficulty} captures whether the attack produces obvious symptoms (Low), subtle indicators requiring trained observation (Medium), minimal observable signatures (High), or is effectively undetectable without specialized monitoring (Critical).

The composite severity rating for each attack-consequence pair reflects the maximum severity across applicable dimensions, weighted by consequence category. Human safety consequences receive elevated weighting ($1.5\times$) relative to purely economic or mission impacts, reflecting the irreversibility and ethical primacy of personnel harm. Each attack category is assessed across six impact dimensions: Robot (physical platform damage), Human (operator and bystander safety), Mission (operational objective completion), Infrastructure (collateral damage to the operating environment), Economic (direct and indirect financial losses), and Strategic (intelligence compromise, adversary capability gain, or reputational damage).

\begin{table}[htbp]

\caption{Consequence Severity Framework for Quadruped Teleoperation Security Incidents}
\label{tab:consequence_severity}
\centering
\footnotesize
\begin{adjustbox}{max width=\columnwidth}
\begin{tabular}{|p{2.4cm}|c|c|c|c|c|c|}
\hline
\textbf{Attack Category} & \textbf{Robot} & \textbf{Human} & \textbf{Mission} & \textbf{Infra.} & \textbf{Econ.} & \textbf{Strategic} \\
\hline
Video Feed Manipulation & L & H & C & M & M & H \\
\hline
Display Injection & -- & H & H & L & L & M \\
\hline
Replay Attack & M & M & H & M & M & H \\
\hline
Control Command Hijacking & C & C & C & C & H & C \\
\hline
Delay Injection & M & H & H & M & M & H \\
\hline
Camera/VO Attack & H & M & H & H & H & M \\
\hline
LiDAR Spoofing & H & M & H & H & H & M \\
\hline
Cybersickness Induction & -- & H & H & L & M & L \\
\hline
Constant Jamming & L & M & C & M & M & H \\
\hline
MITM Attack & V & H & C & H & H & C \\
\hline
FDIA & H & H & H & H & H & H \\
\hline
GPS Spoofing & M & M & H & M & M & H \\
\hline
RoboPAIR Jailbreak & V & C & C & C & C & C \\
\hline
Supply Chain Attack & L & M & H & H & C & C \\
\hline
\end{tabular}
\end{adjustbox}

\vspace{2mm}
\scriptsize{\textbf{Note:} L=Low (minor/recoverable), M=Medium (significant but contained), H=High (severe, substantial response needed), C=Critical (catastrophic/irreversible), V=Variable (context-dependent), --=None.}

\end{table}

This examination reveals several tendencies. Initially, control command hijacking and RoboPAIR jailbreak attacks exhibit the most significant overall risk profile, attaining Critical severity across almost all effect dimensions. This illustrates the inherent risk of an adversary obtaining direct command authority over a kinetically capable platform operating in proximity to personnel—a consequence category without parallel in purely digital systems. Secondly, assaults aimed at the human operator—such as cybersickness induction, display injection, and delay injection—demonstrate a notable asymmetry: they entail significant human and mission repercussions while inflicting minimal direct harm to the robot or equipment. This asymmetry has significant consequences for defense priority, since these attacks may be deprioritized by frameworks that concentrate solely on platform security while neglecting the operator as an essential system component. Third, supply chain attacks exhibit a reversed severity profile relative to most other categories: their immediate impact at the operational level is minimal, although their strategic and economic ramifications are significant due to the enduring, clandestine access they facilitate. An adversary possessing supply chain access can determine the timing and manner of escalation, resulting in an unlimited range of potential consequences. The frequency of Variable (V) ratings for MITM and RoboPAIR indicates authentic context-dependence rather than analytical ambiguity: an MITM attack during a patrol mission in an unoccupied warehouse yields fundamentally different outcomes compared to the same attack occurring during explosive ordnance disposal near civilian infrastructure.
The framework demonstrates that no attack category exhibits uniformly low severity across all dimensions, confirming that a comprehensive defense strategy cannot afford to neglect any layer of the attack taxonomy. Moreover, the prevalence of Critical ratings in the Human, Mission, and Strategic columns—rather than solely in Robot—highlights that the most significant risks associated with quadruped teleoperation security failures transcend mere platform replacement expenses.

\section{Defense Mechanisms Analysis: A Maturity-Aware Assessment}

The prior attack taxonomy and consequence analysis outline a detailed threat landscape of six layers, featuring cascading failure pathways that may occur within sub-second intervals. This prompts a practical inquiry: which hazards may be alleviated with existing technology, and where do significant protection deficiencies remain?

To address this, we conduct a systematic analysis of defense mechanisms documented in the literature, categorizing them by deployment readiness rather than only by technical classification. A significant drawback of several current surveys is their tendency to conflate speculative research concepts with production-level controls without clearly differentiating maturity, thereby leading to an exaggerated perception of the security posture that practitioners can feasibly attain. We mitigate this constraint by employing a Technology Readiness Level (TRL) framework, categorizing each defense mechanism according to its present maturity to facilitate risk-informed, implementation-focused defensive planning.

Table~\ref{tab:trl_classification} presents the complete TRL classification of all defense mechanisms examined in this section. We utilize a streamlined three-tier classification corresponding to realistic implementation choices: \textit{Field-Deployed} (TRL 7--9), \textit{Laboratory-Validated} (TRL 4--6), and \textit{Exploratory/Conceptual} (TRL 1--3).

\textbf{Methodological Note on TRL Assessment:} The Technology Readiness Level classifications presented in Table~\ref{tab:trl_classification} represent single-assessor evaluations based on documented validation evidence from cited literature. While independent multi-rater assessment with Cohen's $\kappa$ reliability measurement would strengthen these classifications, resource constraints precluded formal inter-rater validation. To mitigate potential bias, TRL assignments adhere strictly to the NASA/DoD TRL definitions~\cite{mankins1995trl}, with explicit justification documented for each mechanism. Readers should interpret TRL values as informed estimates subject to revision as additional validation evidence emerges. For mechanisms where assessment confidence is limited (e.g., ANYmal security due to proprietary documentation), we assign conservative lower-bound TRL values and note uncertainty explicitly.

\begin{table*}[htbp]
\caption{Technology Readiness Level Classification of Defense Mechanisms for Quadruped Teleoperation}
\label{tab:trl_classification}
\centering
\footnotesize
\begin{tabular}{|p{2.8cm}|c|p{3.0cm}|p{3.0cm}|p{2.5cm}|p{2.0cm}|}
\hline
\textbf{Defense Mechanism} & \textbf{TRL} & \textbf{Validation Status} & \textbf{Quadruped-Specific Validation} & \textbf{Deployment Barrier} & \textbf{Key Ref.} \\
\hline
\multicolumn{6}{|c|}{\cellcolor{green!15}\textbf{Tier 1: Field-Deployed (TRL 7--9) --- Immediately Deployable}} \\
\hline
TLS 1.2+/DTLS Encryption & 9 & Industry standard, deployed on Spot & Yes (Boston Dynamics Spot) & None; 1--3ms overhead & \cite{boston2024security} \\
\hline
X.509/JWT Authentication & 9 & Industry standard, deployed on Spot & Yes (Boston Dynamics Spot) & Certificate management complexity & \cite{boston2024security} \\
\hline
Firmware Signing (Hash Trees) & 8 & Deployed on Spot, standard practice & Yes (Boston Dynamics Spot) & Requires secure boot chain & \cite{boston2024security} \\
\hline
FHSS Anti-Jamming & 8 & Mature military/WiFi technology & No quadruped-specific testing & Synchronization during dynamic gait & \cite{pirayesh2022jamming} \\
\hline
Multi-Link Redundancy & 7 & Commercial UAV/military systems & No quadruped failover time validation & Sub-200ms failover not yet proven & \cite{diller2023jamming} \\
\hline
Network Segmentation/Firewall & 9 & Standard IT practice & Applicable without modification & Onboard resource constraints & \cite{botta2023survey} \\
\hline
\multicolumn{6}{|c|}{\cellcolor{yellow!15}\textbf{Tier 2: Laboratory-Validated (TRL 4--6) --- Promising but Requires Engineering}} \\
\hline
ML-Based IDS (CNN+LSTM) & 5 & 98--99\% accuracy on generic datasets & No quadruped-specific datasets exist & Dataset gap, false positive rates & \cite{holdbrook2024ids, martin2025rips} \\
\hline
CARLO LiDAR Defense & 5 & Reduces spoofing to 2.3\% on AV data & Not tested on quadruped point clouds & Compute overhead (10--50ms) & \cite{sun2020carlo} \\
\hline
Passivity-Based Control (TDPC) & 6 & Demonstrated at ANA XPRIZE on quadruped & Yes (Risiglione et al., 2021) & Conservative behavior limits agility & \cite{risiglione2021passivity} \\
\hline
VO Inconsistency Detection & 5 & 99.5\% detection, <100ms, no GPU & Tested on generic VO pipelines & Not validated on legged platforms & \cite{xu2025voattack} \\
\hline
Cybersickness Detection (Bi-LSTM) & 4 & 91.1\% F1 in lab VR studies & Not tested during teleoperation & Sensor requirements (eye tracking) & \cite{yalcin2024cybersickness} \\
\hline
RIDPS (ROS2 IDS) & 5 & 97.1\% on ROS2 traffic & Generic ROS2, not gait-specific & Limited attack scenario coverage & \cite{martin2025rips} \\
\hline
Lightweight Crypto (ASCON) & 6 & NIST standardized, IoT benchmarks & Not tested on robotic workloads & Integration with ROS2/DDS needed & \cite{radhakrishnan2024lightweight} \\
\hline
Multi-Sensor Fusion Defense & 5 & Demonstrated in AV context & Not validated for legged kinematics & Computational overhead on embedded & \cite{sun2020carlo} \\
\hline
DRL Anti-Jamming & 4 & Simulation results promising & No physical robot testing & Sim-to-real transfer unknown & \cite{antijarl2024} \\
\hline
\multicolumn{6}{|c|}{\cellcolor{red!10}\textbf{Tier 3: Exploratory/Conceptual (TRL 1--3) --- Long-Term Research Directions}} \\
\hline
Digital Twin Anomaly Detection & 3 & CPS framework validated on industrial systems & No legged robot implementation & Real-time sync latency, physics model fidelity & \cite{xu2021digitaltwin, erceylan2025dtcps} \\
\hline
Zero-Trust Architecture (ROS2) & 3 & Robot Immune System EPP demonstrated & Limited to industrial arms & ROS2 node-level enforcement complexity & \cite{aliasrobotics2020zt, deng2022insecurity} \\
\hline
Moving Target Defense & 3 & CPS control framework with entropy optimization & No robotic validation & Switching overhead, stability guarantees & \cite{kanellopoulos2020mtd, griffioen2021mtd} \\
\hline
Runtime Verification (ROS) & 4 & ROSRV and TeSSLa-ROS demonstrated & Limited attack scenario coverage & Monitoring overhead, specification completeness & \cite{huang2014rosrv, celik2023rv} \\
\hline
Federated Learning IDS & 2 & Conceptual framework only & No robotic datasets available & Privacy-utility tradeoff undefined & -- \\
\hline
Gait Signature Authentication & 1 & Proposed concept, no implementation & Theoretical only & Fundamental feasibility unproven & -- \\
\hline
Locomotion-Aware Security & 1 & Proposed concept, no implementation & Theoretical only & Requires gait-phase detection integration & -- \\
\hline
XAI-Guided Cybersickness Defense & 2 & SHAP analysis on detection models & Lab setting only & Real-time XAI overhead unknown & \cite{kundu2025cybersicknessattack} \\
\hline
\end{tabular}

\vspace{2mm}
\scriptsize{\textbf{Note:} 1=Basic principles observed, 2=Technology concept formulated, 3=Proof of concept, 4=Lab validation, 5=Validation in relevant environment, 6=Demonstration in relevant environment, 7=System prototype in operational environment, 8=System complete and qualified, 9=Proven in operational environment. \textbf{Critical Observation:} Only communication-layer defenses (TLS/DTLS, authentication, FHSS) have achieved field deployment. No quadruped-specific defense mechanism beyond passivity-based control has been validated on an actual legged platform.}
\end{table*}

\subsection{Tier 1: Field-Deployed Defenses (TRL 7–9)}

These mechanisms are \emph{deployable today} and have demonstrated operational effectiveness in real systems. For organizations fielding quadruped robots, they should be treated as \textit{baseline controls}, not optional hardening.

\subsubsection{TLS/DTLS Encrypted Communication}
Transport Layer Security (TLS) and its datagram-oriented variant (DTLS) provide the foundational cryptographic layer for teleoperation links, protecting confidentiality and integrity against passive interception and active manipulation. Boston Dynamics, for example, enforces TLS 1.2+ for API traffic and prioritizes cipher suites that provide forward secrecy \cite{boston2024security}. For latency-sensitive control paths that rely on UDP, DTLS 1.2 offers comparable security guarantees while preserving datagram semantics, making it a practical fit for constrained or lossy networks.

\textbf{Performance Impact:} In typical embedded deployments, per-packet DTLS processing introduces on the order of 1–3,ms additional latency, which is generally tolerable even for high-rate command/telemetry streams (100–1000,Hz) when appropriately engineered (e.g., avoiding oversized records, minimizing copies, leveraging hardware crypto where available). The dominant cost is often session establishment: a full TLS handshake can add roughly 100–150,ms at connection setup, which is acceptable for initial session creation but unsafe as a \emph{recovery path} during dynamic locomotion. A quadruped that loses link mid-stair ascent cannot afford to “pause and handshake,” motivating \emph{fast re-keying and reconnection} via session resumption and connection continuity features (e.g., TLS session tickets, DTLS connection IDs), which can reduce re-establishment latency to sub-10,ms in practice \cite{radhakrishnan2024lightweight}.

Fully deployed on Boston Dynamics Spot, these protections are \emph{available} across modern stacks but are frequently left unused in practice. In ROS2/DDS ecosystems, DDS Security can be enabled but is often disabled by default or omitted at deployment time because the configuration burden is non-trivial \cite{diluoffo2019ros2}. The binding constraint is rarely cryptography or protocol capability; it is operational discipline. Enabling DDS Security typically requires standing up a local public key infrastructure (PKI), generating and distributing per-participant certificates, authoring governance and permissions artifacts, and establishing processes for key rotation and revocation. In many research labs and small organizations, these lifecycle tasks are treated as “later,” and “later” becomes “never,” leaving secure-by-design middleware running in effectively open-network mode.

\subsubsection{Authentication and Access Control}
Authentication is the gatekeeper control: without it, encryption and integrity mechanisms cannot reliably prevent impersonation or unauthorized command issuance. Boston Dynamics Spot illustrates a defensible baseline by combining modern application-layer authentication (JWTs signed with ES256, i.e., ECDSA over P-256) with mutual authentication using X.509 client certificates and per-device unique cryptographic keys \cite{boston2024security}. This design supports both scalable identity management (token-based authorization) and strong device binding (certificate-based mutual TLS), reducing the feasibility of trivial replay, credential stuffing, and “anyone on the network can be the operator” failure modes.

The disparity with platforms devoid of authentication highlights a disconcerting truth: the primary security deficiency in numerous quadruped implementations is not the lack of innovative procedures, but rather the deficiency in \emph{deployment hygiene}. The CVE-2025-41108 vulnerability in Ghost Robotics Vision 60 pertains to an unauthenticated control channel; Xiaomi CyberDog's default credentials (e.g., \texttt{mi}/\texttt{123}) undermine access control by rendering it a public secret; Unitree's BLE provisioning vulnerabilities facilitate wireless access that can circumvent higher-layer controls. These examples collectively suggest that the most significant immediate task is not the invention of novel authentication methods, but rather the constant enforcement of established identification, authorization, and credential-management standards throughout the whole teleoperation framework.

\textbf{Practical Recommendation:} For organizations deploying any quadruped platform, the single highest-impact security improvement is enabling and properly configuring existing authentication mechanisms. This requires no novel technology, only systematic implementation of established practices.

\subsubsection{Frequency Hopping Spread Spectrum (FHSS)}

Frequency Hopping Spread Spectrum (FHSS) mitigates \emph{narrowband} jamming by rapidly switching carrier frequencies according to a shared hopping sequence, forcing the attacker to either (i) jam a wide band (higher power, higher detectability) or (ii) predict the hop pattern (cryptographic problem if implemented correctly). A typical keyed hopping update (Eq.~\ref{eq:fhss}) can be expressed as
\begin{equation}
\label{eq:fhss}
ch_{next} = (ch_{current} + HMAC(K_{hop}, t)) \bmod N_{ch}
\end{equation}
where $K_{hop}$ is the shared hopping key, $t$ is the current time slot (or hop index), and $N_{ch}$ is the number of available channels \cite{pirayesh2022jamming}. FHSS is operationally mature: it has been used in military radios for decades and is foundational in short-range standards (e.g., Bluetooth) precisely because it increases resilience against interference and certain classes of opportunistic jamming.

\textbf{Quadruped-Specific Consideration:} The engineering risk with FHSS is not whether hopping “works,” but whether \emph{coordination failure} creates a worse outcome than the attack it is meant to mitigate. Quadruped teleoperation is uniquely sensitive to short communication discontinuities during phases such as swing-to-stance transitions, stair ascent, or obstacle traversal. A hop schedule that is too aggressive increases the probability of transient desynchronization under multipath fading, packet loss, or clock drift—precisely the conditions that often co-occur with contested RF environments. Conversely, a hop schedule that is too slow reduces jamming resistance and can be tracked by reactive or adaptive jammers.

Practically, FHSS must be designed with a “fail-soft” synchronization strategy: bounded rejoin time, explicit resynchronization beacons, and pre-planned fallback behavior on the robot that preserves balance without relying on immediate operator input. The literature has established FHSS as a mature anti-jam primitive, but (to the best of the cited surveys) there is no quadruped-specific evaluation that quantifies how hop rate, resync latency, and packet loss interact with locomotion stability margins during dynamic terrain transitions. That gap is best treated as a validation and certification problem—hardware-in-the-loop plus field tests under controlled jamming—rather than as a requirement for new anti-jamming theory.

\subsubsection{Multi-Link Communication Redundancy}
Maintaining redundant communication paths is one of the most effective ways to blunt \emph{link-specific} attacks (or failures) in teleoperation. The basic idea is simple: do not bet the robot’s balance on a single RF technology. A practical stack often includes a primary high-throughput link (e.g., WiFi 5,GHz), a secondary link with better penetration characteristics (e.g., WiFi 2.4,GHz), a wide-area cellular path (4G/LTE), and—when missions justify the cost—an emergency satellite channel \cite{diller2023jamming}. Redundancy increases attacker workload (they must degrade multiple links) and reduces the probability that a single-point DoS, localized interference source, or protocol-specific exploit produces total loss of operator connectivity.

What matters operationally is not just \emph{having} multiple links, but how traffic is carried across them. Classic network “failover” is typically \emph{reactive}: the system detects primary-link degradation, triggers a handover, re-establishes sessions, and then resumes streaming. In many mobile robotics and industrial IoT deployments, a failover time on the order of seconds is tolerated because the platform can pause safely and wait. Quadrupeds are different: during dynamic locomotion, seconds are an eternity. If control or high-fidelity state feedback drops mid-stride on stairs, the robot may enter a fallback mode that is physically incompatible with the instantaneous contact state, making the “recovery” itself a source of harm.

\textbf{Critical Gap:} Reactive failover times of $\sim$2,s may be acceptable for wheeled robots, but they are incompatible with quadruped dynamic stability requirements in challenging terrain. Quadruped-grade redundancy should target \emph{sub-200,ms} effective continuity, which strongly favors \emph{proactive} multi-link strategies over reactive ones. The most robust pattern is to transmit over multiple paths concurrently (or at least maintain warm standby with continuous keep-alives and cryptographic session resumption) and perform \emph{receiver-side selection} (or aggregation) based on measured latency, jitter, and loss—rather than waiting for a link to “fail” before switching.

Concretely, this implies three engineering shifts. First, treat the communication layer as a \emph{real-time subsystem}: monitor latency/jitter at high rate, not just connectivity status. Second, minimize reconnection overhead via pre-established secure contexts (e.g., session resumption, connection identifiers) so that changing paths does not trigger full handshakes. Third, design the robot’s controller to degrade gracefully under partial information: if video degrades but command uplink remains, prioritize balance-critical state channels; if uplink degrades, prioritize a locally safe locomotion policy consistent with the current gait phase. The core technologies for multi-link redundancy already exist; the unmet need is quadruped-specific timing validation and integration so that redundancy improves safety instead of introducing new discontinuities.

\subsection{Tier 2: Laboratory-Validated Defenses (TRL 4–6)}
These mechanisms have demonstrated effectiveness in controlled environments but require further engineering, adaptation, or validation before deployment on quadruped platforms. They represent the most productive near-term research investment.

\subsubsection{Machine Learning-Based Intrusion Detection}
Table~\ref{tab:ids_performance} summarizes representative ML-based IDS approaches along with their reported performance and evaluation context. Because the underlying studies use heterogeneous metrics (e.g., accuracy, AUC, and F1-score), the table reports the primary metric as stated in the corresponding reference.

\begin{table}[htbp]
\caption{ML-Based Intrusion Detection: Reported Performance and Validation Context}
\label{tab:ids_performance}
\centering
\footnotesize
\begin{adjustbox}{max width=\columnwidth}
\begin{tabular}{|l|c|c|c|c|}
\hline
\textbf{Approach} & \textbf{Primary Metric} & \textbf{FPR} & \textbf{Validation Context} & \textbf{Ref.} \\
\hline
CNN+LSTM Hybrid  & Acc.\ 98.2\% & 1.1\%  & Generic network traffic & \cite{holdbrook2024ids} \\
DeepIIoT        & Acc.\ 99.0\% & $<$1\% & IoT benchmark traffic   & \cite{zafir2024iorot} \\
MVT-Flow        & AUC 94.5\%   & 2.3\%  & ROS2 dataset (ROSPaCe / voraus-AD) & \cite{puccetti2024rospace} \\
RIDPS           & Acc.\ 97.1\% & 1.8\%  & ROS2 traffic (generic scenarios)  & \cite{martin2025rips} \\
Bi-LSTM+CGAN    & F1 91.1\%    & 2.1\%  & VR lab study (operator-state)     & \cite{yalcin2024cybersickness} \\
\hline
\end{tabular}
\end{adjustbox}
\vspace{2mm}

\scriptsize{\textbf{Note:} Reported metrics are conditional on the specific datasets and threat models used in each study; cross-domain generalization is not established. In particular, no quadruped-specific IDS dataset exists, and transfer performance to legged-robot deployments is unknown. Quadruped operation induces structured and nonstationary communication patterns (e.g., gait-cycle periodicity, terrain-transition bursts, load-dependent timing jitter, and mode-switch signatures) that are largely absent from current ROS/IoT corpora. Consequently, published performance should be interpreted as a dataset-bound estimate rather than an operational guarantee for quadruped teleoperation.}
\end{table}

Martin et al. (2025) proposed RIDPS explicitly tailored to ROS2 deployments, reporting 97.1\% detection accuracy and a “graduated response” design that can escalate from logging to active mitigation rather than relying on a single blunt reaction \cite{martin2025rips}. Complementing this, the ROSPaCe dataset provides labeled ROS2-specific attack scenarios intended to support comparable evaluation of ROS2 intrusion-detection methods \cite{puccetti2024rospace}. Nguyen et al. \cite{nguyen2025cyberdefense} introduced a compact testbed for assessing cyber attacks and defenses on autonomous mobile robots, pairing lightweight countermeasures—such as dynamic watermarking for camera streams and velocity-consistency checks for LiDAR validation—with an experimental workflow that is feasible for resource-constrained platforms. Moving beyond signature-style IDS assumptions, the ROBOCOP framework \cite{robocop2024iros} proposed a zero-day cyber-physical detection approach using zero-shot learning, directly targeting the core weakness of many IDS pipelines: brittle performance when confronted with previously unseen attack patterns. Earlier foundational work by Guerrero-Higueras et al. \cite{guerrero2018cyber} demonstrated attack detection for indoor real-time localization systems, a setting that transfers naturally to quadruped indoor missions where GPS denial is the norm and localization faults can propagate rapidly into unsafe locomotion decisions.

A critical caveat is that existing IDS datasets and evaluation protocols remain dominated by \emph{generic} ROS/ROS2 traffic and platform behaviors. Quadruped operation introduces distinctive dynamics—periodic gait cycles, mode-dependent message bursts during terrain transitions, load-dependent timing jitter, and tight inter-leg coordination constraints—that are largely absent from current training corpora. As a result, reported accuracy values should be interpreted conservatively: they are best viewed as \emph{upper bounds} under simplified conditions, not as validated operational metrics for legged systems. Until IDS models are trained and calibrated on representative quadruped mission traces (including stair ascent/descent, slip recovery, and autonomy/teleop mode switching), performance claims cannot be assumed to transfer without substantial degradation in false positives, missed detections, or both.

\textbf{Path to Deployment:} The critical enabling step is a quadruped-specific IDS dataset that captures \emph{normal} operation across gait modes (walk/trot/bound), terrain classes (flat, stairs, rubble, industrial clutter), payload conditions, and teleoperation modalities (direct, shared autonomy, VR/AR). Until such a dataset exists, deploying generic ML-based IDS on quadruped platforms can provide partial coverage, but detection performance for quadruped-specific signatures remains unquantified and should be treated as an unknown-risk assumption.

\subsubsection{LiDAR Spoofing Detection}
Sun et al. (2020) introduced CARLO, reducing LiDAR spoofing success from 80\% to 2.3\% through physical occlusion-law verification and shadow pattern analysis \cite{sun2020carlo}. Additional techniques include Shadow-Catcher (91.2\% TPR, 82.5\% TNR) and azimuth-based detection (100\% TPR, 0.02\% FPR) \cite{cao2023pra}.

\textbf{Quadruped-Specific Assessment:} These countermeasures were formulated and verified on autonomous vehicle platforms equipped with roof-mounted LiDAR and primarily flat road configurations. Quadrupeds exhibit three significant distinctions: (i) sensor positioning at lower and platform-dependent elevations, (ii) substantial sensor movement caused by body oscillations and footfall impacts, and (iii) functioning within complex 3D environments (stairs, handrails, pipes, scaffolding) that modify point-cloud statistics and occlusion/shadow configurations. Furthermore, documented computational delays of from 10 to 50 ms may be significant for high-frequency perception systems where state estimation and foothold planning are closely interconnected. Consequently, the principal deficiency lies not in the availability of algorithms but in the re-validation under quadrupedal operating conditions and the clear characterization of the latency-robustness tradeoff in legged locomotion.

\subsubsection{Passivity-Based Control for Delay Tolerance}
Passivity-based control guarantees stability under bounded delays by enforcing net energy dissipation. A standard passivity condition can be expressed as:
\begin{equation}
\int_0^T f(t)^{\top} v(t)\, dt \geq -E_0
\end{equation}
where $E_0$ bounds the initial stored energy.

Risiglione et al. (2021) exhibited Time-Domain Passivity Control (TDPC) for quadrupedal teleoperation, accommodating time-varying delays of up to 500 ms while ensuring stable operation, and secured third place in the ANA Avatar XPRIZE competition \cite{risiglione2021passivity}. To the best of our knowledge, this is the sole defense mechanism in our study validated on a physical quadruped during teleoperation. The extensive literature on bilateral teleoperation offers the foundational theoretical framework. Nuño et al. developed passivity-based controllers for nonlinear bilateral teleoperation \cite{nuno2011passivity}, whilst Haddadi and Hashtrudi-Zaad offered a tutorial on scattering transformation, wave variables, and energy-based methodologies \cite{haddadi2011passivity}. A recent analysis of force-feedback bilateral teleoperation highlights wave variables, TDPC, and four-channel designs as the predominant delay-compensation paradigms \cite{chen2024teleop}.

A crucial security detail is that passive layers can also become targets for attacks. Munteanu et al. \cite{munteanu2018bilateral} shown that hostile manipulation of master-side energy variables can lead the slave to deduce adequate available energy, resulting in unstable behavior that effectively circumvents passivity safeguards. Takanashi et al. \cite{takanashi2023encrypted} proposed an alternative approach: the application of homomorphic encryption to four-channel bilateral control, facilitating encrypted communication and controller execution while maintaining posture synchronization and force feedback—an effective countermeasure against FDIAs aimed at control parameters. Further advancements encompass passivity-shortage formulations that mitigate stringent passivity requirements \cite{passivityshortage2020}, energy-tank methodologies that alleviate force distortion caused by injected damping \cite{gao2024forcefeedback}, time-domain passivity controllers with delayed force feedback for bilateral tasks \cite{wang2015delayforce}, and sliding-mode controllers integrated with modified wave transformations for nonlinear systems \cite{hashemzadeh2015teleop}. Dash et al. \cite{dash2021pidpiper} introduced PID-Piper for real-time detection and compensation of compromised sensors and actuators during post-attack recovery. A survey on time-delay reduction emphasizes hybrid techniques that include user-intent modeling, movement prediction, and delay prediction as a promising approach \cite{teledelay2021survey}.

\textbf{Limitations:} Passivity-based designs are inherently conservative and may diminish responsiveness. During an assault or significant delay fluctuation, TDPC generally introduces damping, which operators may interpret as sluggish movement. The performance-security tradeoff must be clearly articulated in operator training and mission planning. Furthermore, the XPRIZE validation took place under controlled settings; systematic validation involving \emph{adversarial} delay injection and coordinated multi-layer attack scenarios is still an unresolved engineering necessity.

Wave-variable transformations provide formal guarantees under arbitrary \emph{constant} delays:
\begin{equation}
u = \frac{1}{\sqrt{2b}}(f + b\dot{x}), \quad v = \frac{1}{\sqrt{2b}}(f - b\dot{x})
\end{equation}
but assume delay characteristics that differ from the nonstationary jitter and burst loss patterns typical of wireless links \cite{hamdan2021teleop}.

\textbf{Formal Safety Guarantees via Control Barrier Functions:} While passivity-based approaches guarantee stability under bounded delays, they do not provide explicit safety guarantees (e.g., collision avoidance, joint limit satisfaction) under adversarial conditions. Control Barrier Functions (CBFs) offer a complementary framework for safety-critical control that directly encodes physical safety constraints. Given a safe set $\mathcal{S} = \{x \in \mathbb{R}^n : h(x) \geq 0\}$ defined by a continuously differentiable function $h$, a CBF enforces forward invariance of $\mathcal{S}$ by constraining admissible control inputs:
\begin{equation}
\dot{h}(x, u) + \alpha(h(x)) \geq 0
\end{equation}
where $\alpha$ is an extended class-$\mathcal{K}$ function. For quadruped teleoperation, relevant safety constraints include joint position and velocity limits, center-of-mass stability margins relative to the support polygon, ground reaction force bounds preventing slip, and minimum clearance distances from obstacles and personnel.

CBF-based safety filters can intercept potentially destabilizing operator commands---whether malicious or erroneous---and project them onto the nearest safe control input via quadratic programming:
\begin{equation}
u^* = \arg\min_{u \in \mathcal{U}} \|u - u_\text{des}\|^2 \quad \text{s.t.} \quad \dot{h}(x, u) + \alpha(h(x)) \geq 0
\end{equation}
where $u_\text{des}$ is the desired (possibly compromised) control input and $u^*$ is the minimally-modified safe input. This formulation provides a principled mechanism for bounding the physical consequences of control-layer attacks: even if an adversary successfully injects malicious commands, the CBF filter constrains achievable harm to the boundary of the safe set.

The integration of CBFs with passivity-based teleoperation remains an open research direction. The composition of passivity constraints (which regulate energy flow) and barrier constraints (which enforce state-space invariance) introduces potentially competing requirements: a passive controller may require energy dissipation that conflicts with the control authority needed to maintain barrier satisfaction, particularly during aggressive recovery maneuvers. Formal reconciliation of these frameworks---potentially through passivity-constrained CBF formulations or barrier-aware energy tanks---represents a promising avenue for quadruped teleoperation security.
\subsubsection{Visual Odometry Protection}
Xu et al. established that the observation of reprojection-error statistics, feature-descriptor stability, and motion-prediction residuals facilitates attack detection with an accuracy of 99.5\% and latency below 100 ms, without the need for GPU acceleration \cite{xu2025voattack}. The resultant computational footprint is consistent with standard quadruped onboard computers. Nonetheless, the approach has not been tested in legged locomotion scenarios, where body oscillation, rapid pitch and roll transients, and periodic foot-impact vibrations compromise visual-feature consistency and may exacerbate reprojection inaccuracy, even under benign settings. Quadruped deployment necessitates (i) gait-aware baselining of standard residual distributions across locomotion modes and (ii) explicit delineation of the sensitivity-false-alarm tradeoff under terrain-induced motion blur and occlusion.

\subsubsection{Lightweight Cryptography}
Table~\ref{tab:crypto_comparison} compares lightweight authenticated-encryption candidates appropriate for resource-constrained quadruped subsystems (e.g., peripheral microcontrollers, provisioning links, and low-power telemetry channels).

\begin{table}[htbp]
\caption{Lightweight Cryptography Performance Comparison}
\label{tab:crypto_comparison}
\centering
\footnotesize
\begin{adjustbox}{max width=\columnwidth}
\begin{tabular}{|l|c|c|c|c|c|}
\hline
\textbf{Algorithm} & \textbf{RAM} & \textbf{ROM} & \textbf{Throughput} & \textbf{Security} & \textbf{Ref.} \\
\hline
ASCON-128 & 940 B & 2.1 KB & 188 kbps & 128-bit & \cite{radhakrishnan2024lightweight} \\
SPECK-128 & 512 B & 8.8 KB & 824 kbps & 128-bit & \cite{radhakrishnan2024lightweight} \\
AES-128-GCM & 2.1 KB & 15 KB & 227 kbps & 128-bit & \cite{radhakrishnan2024lightweight} \\
ChaCha20-Poly & 1.8 KB & 12 KB & 312 kbps & 256-bit & \cite{radhakrishnan2024lightweight} \\
\hline
\end{tabular}
\end{adjustbox}

\vspace{2mm}
\scriptsize{\textbf{Note:} ASCON (NIST lightweight cryptography standard, 2023) provides authenticated encryption with associated data (AEAD) with RAM usage below 1\,KB \cite{radhakrishnan2024lightweight}. These primitives are standardized and well-characterized for IoT, but robotics-specific integration questions remain open: how key management, rotation, and provisioning are implemented at scale; how AEAD overhead interacts with high-rate control traffic; and how these ciphers are operationally integrated with ROS2/DDS security workflows rather than deployed as isolated point solutions.}
\end{table}

\subsection{Tier 3: Exploratory and Conceptual Defenses (TRL 1--3)}
These mechanisms represent promising research directions with foundations in adjacent domains, but they lack robotic---particularly quadruped---specific operational validation. They should be treated as medium-to-long-term priorities, not near-term deployable protections.

\subsubsection{Digital Twin-Based Anomaly Detection}
Digital twins—high-fidelity virtual duplicates that are synchronized with physical systems in real time—have emerged as a significant paradigm for cyber-physical security. Xu et al. (2021) introduced a fundamental anomaly-detection methodology for digital twin-based cyber-physical systems, integrating simulation consistency assessments with data-driven categorization to identify and attribute anomalies related to sensor malfunctions, model inconsistencies, and cyberattacks \cite{xu2021digitaltwin}. The LATTICE framework advanced this approach using curriculum learning to enhance robustness in training on CPS anomaly data \cite{lattice2023acm}. Erceylan et al. (2025) conducted an extensive examination of digital twins for sophisticated threat modeling in CPS cybersecurity, encompassing intrusion detection, simulation-based penetration testing, and security-enhancing digital twins (SEDTs) specifically developed to facilitate defensive monitoring and response \cite{erceylan2025dtcps}.

Wang et al. (2024) developed a digital-twin pipeline for anomaly detection in industrial robots, utilizing physics-informed hybrid convolutional autoencoders to compare predicted joint currents with measured values, thereby identifying deviations that suggest faults or attacks \cite{wang2024dtrobot}. Hnaien et al. (2025) illustrated unsupervised anomaly detection by integrating a high-fidelity digital twin of the St\"{a}ubli TX60 with deep autoencoders, facilitating real-time defect detection devoid of labeled training data \cite{hnaien2025dtrobot}. Arm et al. (2024) introduced a real-time "digital double" for legged systems that identifies terrain anomalies through discrepancies between simulated and actual quadruped movements, demonstrating that deviations in joint trajectories can serve as dependable indicators of unforeseen external disturbances \cite{arm2024digitaldouble}.

\textbf{Quadruped-Specific Outlook:} The primary assertion is \emph{cross-layer consistency}: when perception, state estimation, and control are integrated, attacks that remain below the threshold in any individual layer may still result in statistically significant discrepancies between simulation and reality. The primary obstacle is operational fidelity: precise real-time legged digital twins necessitate stringent calibration of contact models, actuator dynamics, and terrain interaction, which fluctuate significantly across different gaits and situations. Consequently, twin-based defenses are now classified as TRL~1--3 for quadruped security, with the immediate research goal being the development of resilient \emph{mismatch-aware} detectors that maintain reliability despite imperfections in the twin.

\textbf{Application to Quadruped Teleoperation Security:} The digital-twin paradigm is especially appealing for quadruped security due to legged locomotion producing high-dimensional, intricately structured behavioral traces, including gait phase timing, contact sequences, joint-torque patterns, and body oscillation modes. An assailant who injects erroneous joint instructions or distorts state estimations should create \emph{physically inconsistent} divergence between the projected trajectories of the twin and the robot's recorded proprioceptive/IMU feedback. The challenges are operational rather than conceptual: (1) ensuring temporal alignment and state synchronization at the precision necessary for rapid gait transitions (considering phase-dependent observability and contact discontinuities), (2) distinguishing adversarial anomalies from genuine exogenous disturbances (such as terrain compliance, slips, and impacts) that inherently generate significant residuals, and (3) adhering to computational and memory constraints on embedded robotic processors while maintaining real-time control deadlines. Current TRL: about 3 for quadruped-related security monitoring, with certain industrial digital-twin anomaly pipelines in related fields reported at TRL 5–6.

\subsubsection{Zero-Trust Architecture for Robotic Systems}
The conventional castle-and-moat model, which assumes implicit confidence for entities within a network perimeter, is inadequate for distributed robotic systems that encompass diverse devices, connections, and levels of privilege. Zero-Trust Architecture (ZTA), which is predicated on the principle of "never trust, always verify" as delineated in NIST SP 800-207 \cite{nist2020zt}, is ideally suited for ROS2-based quadruped implementations where lateral mobility among nodes poses a significant threat.

Alias Robotics (2020) introduced an initial application of Zero Trust Architecture (ZTA) principles to industrial robotic systems by developing a Robot Immune System—an Endpoint Protection Platform (EPP) that enforces trust decisions at the robot endpoint instead of depending exclusively on perimeter controls \cite{aliasrobotics2020zt}. Their architecture employs divided zones (e.g., field/control/process/DMZ/IT isolation) to limit blast radius, with the objective of preventing compromise transmission across robotic and plant networks. Deng et al. (2022) illustrated that SROS2 may exhibit vulnerabilities in authorization and information exfiltration under specific configurations and suggested countermeasures utilizing private broadcast encryption \cite{deng2022insecurity}, consistent with ZTA’s focus on explicit, cryptographically enforced identity and permissions. Yang et al. (2024) conducted a formal analysis of ROS2 communication vulnerabilities using state-transition systems and CIA characteristics, emphasizing that ROS2 implementations lacking supplementary protections are susceptible to node impersonation and parameter manipulation \cite{yang2024ros2formal}.

\textbf{Application to Quadruped Teleoperation:}An effective Zero Trust Architecture (ZTA) interpretation for teleoperated quadrupeds necessitates: (1) \emph{per-entity authentication} for each ROS2 participant and every publisher/subscriber relationship, (2) \emph{continuous re-authorization} (session freshness, ephemeral credentials, and revocation) instead of one-time access, (3) \emph{micro-segmentation} that segregates locomotion-critical control nodes from high-risk surfaces (perception ingestion, UI/VR clients, cloud relays), and (4) \emph{least-privilege policies} that inhibit compromised sensing or UI components from executing actuator-affecting commands. The primary engineering limitation is latency: these controls must adhere to the stringent timing constraints of gait-critical loops and control-path watchdogs, typically within single-digit milliseconds. Current TRL: around 3 for ROS2 quadruped teleoperation at scale, mostly attributable to integration and operational complexity rather than absent primitives.

\subsubsection{Moving Target Defense for Robotic Control Systems}
Moving Target Defense (MTD) mitigates the advantage of attackers by incorporating controlled stochastic variation into system configurations or control parameters, thereby diminishing the efficacy of reconnaissance and impairing the persistence of model-based attacks. Kanellopoulos and Vamvoudakis (2020) established an MTD control framework for CPS that stochastically alters parameters while enhancing unpredictability using entropy metrics, integrating proactive variation with reactive detection utilizing integral Bellman error \cite{kanellopoulos2020mtd}. Griffioen et al. suggested the use of MTD in CPS security with the introduction of time-varying control parameters, which diminish the attackers' confidence in the plant model, thus limiting the viability of covert false data injection \cite{griffioen2021mtd}. Giraldo et al. (2019) shown that MTD can reveal covert CPS attacks that elude traditional residual-based detectors \cite{giraldo2019mtd}. El-Mahdi et al. (2022) introduced an IoT-enabled MTD methodology employing signal replication and randomized selection to generate layered uncertainty, substantiating the notion on a real-time water system testbed \cite{elmahdi2022mtdiot}.

\textbf{Quadruped-Relevant Interpretation:} In teleoperated quadrupeds, the most justifiable MTD targets are \emph{non-performance-critical} parameters that influence an attacker's identifiability without destabilizing locomotion (e.g., randomized watermarking signals on specific telemetry channels, rotating cryptographic identities/keys, and regulated variation in non-critical estimator gains), as opposed to drastic alterations of fundamental gait or balance parameters that may jeopardize stability. The primary trade-off is evident: heightened unpredictability can directly elevate operator workload and diminish the closed-loop experience if it disrupts the transparency of teleoperation. Therefore, MTD designs tailored for quadrupeds must be \emph{phase-aware} (restricted during swing/stance transitions) and \emph{transparent-by-construction} (variations limited to remain beneath human-perceptual and controller-sensitivity thresholds), or the defense may pose a safety hazard.

\textbf{Application to Quadruped Teleoperation:} MTD is theoretically appealing for quadruped teleoperation as it can diminish an assailant's capacity to construct and sustain an accurate picture of the closed loop, while simultaneously escalating the expenses associated with reconnaissance and persistence. Three pertinent advantages are: (1) the incorporation of \emph{structured stochasticity} (such as watermarking or bounded parameter dithering) can complicate covert FDIA design by disrupting the attacker's system-identification process; (2) the \emph{dynamic reconfiguration} of communication pathways, credentials, and session parameters can diminish dwell time and impede lateral movement; and (3) the \emph{natural variability} of legged locomotion (terrain-induced disturbances, gait-dependent oscillations) may obscure minor defensive perturbations, allowing “security noise” to blend into pre-existing noisy dynamics. The primary risk is safety: any randomization must be shown constrained to avoid destabilizing gait limit cycles, disrupting contact timing, or compromising operator transparency. A secondary limitation is to timing: the overhead associated with switching and re-keying must adhere to real-time budgets and circumvent abrupt latencies that align with phase transitions (swing-to-stance, stair edge contact). In practice, MTD for quadrupeds should prioritize \emph{phase-aware, low-amplitude, non-critical-path} variations (such as telemetry watermarking, credential rotation, and micro-segmentation rebinds) rather than aggressive control-parameter switching. Current TRL: approximately 3 for generic CPS, and approximately 1–2 for robotic teleoperation deployments, particularly quadrupeds.

\subsubsection{Runtime Verification and Monitoring for Robotic Security}
Runtime verification (RV) offers lightweight formal assurance by observing actual executions in accordance with formally defined properties, identifying violations in real-time without the computational expense of comprehensive model checking. Huang et al. (2014) established ROSRV, a preliminary RV framework for ROS that intercepts and scrutinizes message flows in accordance with officially specified safety and security features \cite{huang2014rosrv}. Celik et al. (2023) expanded RV for ROS security by introducing modular monitors that can be tailored for certain threat classes, such as DoS, and may be incrementally added into current deployments \cite{celik2023rv}. Baumeister et al. (2025) presented an RTLola adaptor for ROS that facilitates stream-based real-time monitoring with minimal integration effort, exemplified in an unmanned aerial vehicle scenario for cross-validating detections against LiDAR-derived signals \cite{baumeister2025rtlola}. Santos et al. (2024) offered guidelines for practitioners on RV and field testing in ROS-based systems, incorporating a survey of robotics professionals that identified practical challenges, including property specification work and runtime overhead \cite{santos2024rvros}.

\textbf{Application to Quadruped Teleoperation:} RV is appropriately aligned with quadrupeds, as numerous safety- and security-related behaviors can be articulated as invariants and temporal constraints over streaming signals: joint torque/velocity envelopes, contact-consistency conditions, gait-phase admissibility, command validity assessments, and latency/jitter limits on operator-to-robot channels. Significantly, RV functions as a \emph{cross-layer tripwire}: a singular monitor may integrate physical signals (IMU/contact forces), control commands, and communication timing to identify multi-layer “sub-threshold” attacks that bypass standalone detectors. Specifications oriented towards quadrupeds can encapsulate properties such as: \emph{if the controller designates a swing phase, then the measured vertical contact force must stay below a predetermined threshold, except during brief impact intervals; if this condition is breached, initiate a secure fallback}—addressing both fault scenarios (unanticipated contact) and possible interference (falsified gait state, injected commands). The overhead is generally manageable when monitors are configured as constant-time stream checks; RTLola-style monitors are frequently noted to introduce latency ranging from sub-millisecond to a few milliseconds per evaluation. However, deployment must guarantee that monitors operate on a real-time-safe thread and do not interfere with locomotion-critical tasks. Current TRL: around 4 for ROS-based robotics overall, and approximately 2 for quadruped-specific teleoperation security, principally due to the necessity for validated quadruped-focused property libraries and phase-aware integration..

\subsubsection{Federated Learning for Distributed IDS}
Federated learning (FL) may facilitate the enhancement of intrusion-detection models across robotic fleets without the need to centralize sensitive operational data. Rather than transmitting raw telemetry, each robot does local training and shares solely model updates (e.g., gradients or weight deltas), which are then consolidated into a global model.

\textbf{Why it matters:} Deployments of quadrupeds in critical infrastructure and defense settings are frequently limited by data sovereignty, privacy, and export control regulations. FL is appealing because it maintains locality while simultaneously identifying population-level attack patterns that would not be regularly observed by any individual robot.

\textbf{Honest Assessment:}This is currently theoretical for quadruped IDS. The necessary prerequisites are absent: (1) a quadruped-specific IDS dataset that records gait-phase traffic signatures and attack labels, (2) baseline detection models validated in real locomotion scenarios (terrain, gaits, payloads), and (3) fleet-scale orchestration capable of accommodating intermittent connectivity and diverse hardware. Even with those prerequisites, FL presents further unresolved risks: \emph{poisoning attacks} (malicious clients compromise the global model), \emph{non-IID data} (each robot encounters distinct terrains and network conditions), and \emph{update leakage} (model updates may inadvertently disclose information about local data). Consequently, FL need to be regarded as a medium-to-long-term research trajectory, dependent on addressing Gap 1 (dataset formation) and instituting reliable aggregation and robustness procedures. A justifiable maturity assertion currently is \textbf{TRL 1--2} for quadruped IDS.

\subsubsection{Gait Signature Authentication and Locomotion-Aware Security}
Utilizing legged mobility as a security signal is conceptually intriguing due to the difficulty of perfect forgery and the extensive observability through proprioceptive channels (joint encoders, IMU, contact forces). The fundamental concept is to use gait as a physical "fingerprint" and to develop security methods that are specifically phase-aware (stance/swing timing, duty factor, contact sequence), instead of general network-centric protections.

Evidence from related fields substantiates plausibility. In human biometrics, gait-based identification has attained reliable authentication through inertial and radar-derived signatures, with deep learning models extracting unique micro-Doppler and IMU patterns \cite{akter2024legged}. In legged robotics, Shi et al. (2024) demonstrated that adversarial perturbations to reinforcement learning locomotion controllers generate discernible multi-modal signatures across proprioceptive and exteroceptive channels \cite{shi2024adversarial}, suggesting that gait variations generated by attacks may be quantifiable. Arm et al. (2024) showcased a real-time digital double that identifies terrain-induced gait abnormalities through motion-discrepancy analysis \cite{arm2024digitaldouble}, offering a definitive method for locomotion-aware anomaly detection.

\textbf{Honest Assessment:} This direction is promising but not yet implementable. Three feasibility inquiries remain unresolved: (1) \emph{stability and uniqueness}—the consistency of gait signatures for a specific robot across varying payloads, wear, battery conditions, and terrain, while maintaining discriminative capability among different robots; (2) \emph{separability}—the reliability of distinguishing attack-induced gait anomalies from legitimate disturbances (such as uneven terrain, slips, and contact loss) at acceptable false-positive rates; and (3) \emph{real-time compatibility}—the implementation of phase-aware authentication and policy enforcement without introducing latency or jitter into locomotion-critical processes. Addressing them necessitates systematic multi-terrain trials and meticulously crafted evaluation methodologies (cross-day, cross-payload, cross-terrain generalization). A cautious maturity statement is TRL 2–3.

\subsection{VR/AR Operator Protection}

\subsubsection{Cybersickness Detection and Mitigation}
Countermeasures for cybersickness seek to preserve operator functionality by identifying physiological and behavioral indicators (e.g., head motion patterns, gaze instability, latency jitter, or content-induced flicker) and initiating mitigations such as diminishing optic flow, limiting acceleration cues, or dynamically modifying the field of view \cite{yalcin2024cybersickness}. Yalcin et al. documented a 91.1\% F1-score utilizing bidirectional LSTM-based detection in controlled laboratory environments \cite{yalcin2024cybersickness}.

\textbf{Deployment Readiness:} \textbf{TRL 4.} Existing evidence predominantly pertains to seated VR experiences, rather than operators actively teleoperating robots under conditions of workload, stress, and mission restrictions. Two significant uncertainties prevail: (1) the extent to which the assessed cybersickness indicators correlate closely with teleoperation performance deterioration (reaction time, command accuracy, situational awareness), and (2) the degree to which mitigation strategies (e.g., field of view reduction) maintain sufficient perceptual fidelity for secure stabilization during complex maneuvers (stairs, debris, narrow walkways). A validation program for quadrupeds should measure operator performance measures in conjunction with physiological indications during representative locomotion activities.

\subsubsection{Fallback Display Modes}
Fallback display modes offer a regulated decline in functionality when the operator interface is believed to be compromised or when symptoms of cybersickness surpass established thresholds. Fallback options for candidates comprise: (i) reduced-motion "comfort" rendering featuring limited optic flow, (ii) a 2D schematic coupled with a stabilized horizon view, (iii) low-bandwidth telemetry overlays emphasizing balance-critical indicators (contact state, pitch/roll, center of mass margin), and (iv) audio and haptic alerts facilitating cautious recovery.

\textbf{Deployment Readiness:} \textbf{TRL 3--4.} The notion is clear; nevertheless, the significant challenge is in \emph{context-conditioned sufficiency}: a backup mechanism that is secure on level surfaces may prove hazardous on stairs or during navigation. Implementations for quadrupeds must be cognizant of gait and terrain, choosing fallbacks according to locomotion phase and stability margin, and should be accompanied by a controller-side "safe state" policy (e.g., controlled descent, brace posture, or step-to-stop) that is viable under restricted operator perception. Comprehensive investigations delineating the relationship between \emph{fallback mode} and recoverability in various locomotor scenarios are still lacking and are essential prior to asserting field readiness.


\subsection{Defense Coverage Assessment}
Table~\ref{tab:defense_effectiveness} maps defense mechanisms to the six-layer attack taxonomy and annotates each mechanism with Technology Readiness Level (TRL). This joint view enables a realistic appraisal of which protections are deployable today versus those that remain experimental or context-dependent.

\begin{table}[H]
\caption{Defense Mechanism Effectiveness and Maturity Matrix}
\label{tab:defense_effectiveness}
\centering
\footnotesize
\begin{adjustbox}{max width=\columnwidth}
\begin{tabular}{|p{2.35cm}|c|c|c|c|c|c|c|}
\hline
\textbf{Defense} & \textbf{L1} & \textbf{L2} & \textbf{L3} & \textbf{L4} & \textbf{L5} & \textbf{L6} & \textbf{TRL} \\
\hline
TLS/DTLS & -- & -- & $\bullet\bullet$ & $\bullet$ & -- & $\bullet$ & 9 \\
\hline
Authentication \& Access Control & -- & -- & $\bullet\bullet$ & $\bullet\bullet$ & -- & $\bullet\bullet$ & 9 \\
\hline
FHSS & -- & -- & $\bullet\bullet$ & -- & -- & -- & 8 \\
\hline
Multi-Link Redundancy & -- & -- & $\bullet\bullet$ & -- & -- & $\bullet$ & 7 \\
\hline
ML-Based IDS & $\bullet$ & -- & $\bullet\bullet$ & $\bullet\bullet$ & $\bullet$ & $\bullet\bullet$ & 5 \\
\hline
Sensor Fusion Consistency Checks & $\bullet\bullet$ & -- & -- & $\bullet$ & $\bullet\bullet$ & -- & 5 \\
\hline
TDPC / Passivity Layer & -- & -- & $\bullet$ & $\bullet\bullet$ & -- & $\bullet$ & 6 \\
\hline
VR/AR Operator Safety & $\bullet$ & $\bullet\bullet$ & -- & -- & -- & -- & 4 \\
\hline
Digital Twin Monitoring & $\bullet$ & -- & $\bullet$ & $\bullet\bullet$ & $\bullet$ & $\bullet$ & 3 \\
\hline
Zero-Trust Architecture & -- & -- & $\bullet\bullet$ & $\bullet$ & -- & $\bullet\bullet$ & 3 \\
\hline
Moving Target Defense & -- & -- & $\bullet$ & $\bullet\bullet$ & -- & $\bullet$ & 3 \\
\hline
Runtime Verification & $\bullet$ & $\bullet$ & $\bullet$ & $\bullet\bullet$ & $\bullet$ & $\bullet$ & 4 \\
\hline
\end{tabular}
\end{adjustbox}
\\[3pt]
\footnotesize{\textbf{Note:} $\bullet\bullet$ (strong): mechanism directly addresses $\geq$70\% of attack vectors in the layer with demonstrated effectiveness; $\bullet$ (partial): mechanism provides indirect mitigation or addresses $<$70\% of vectors; -- (none): mechanism is architecturally inapplicable to the layer. Coverage assessments derive from the attack taxonomy (Section~III) cross-referenced against mechanism capabilities documented in cited literature. L1--L6 correspond to the six-layer attack taxonomy. TRL indicates deployment readiness per Table~\ref{tab:trl_classification}.}
\end{table}

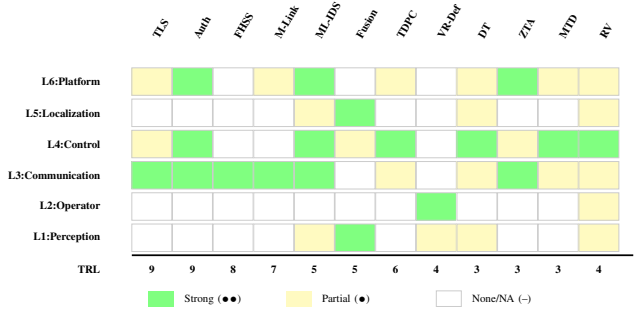
\begin{figure}[htbp]
\centering
\begin{adjustbox}{max width=\columnwidth}
\begin{tikzpicture}[
    cell/.style={minimum width=0.65cm, minimum height=0.45cm, font=\tiny, anchor=center},
    header/.style={font=\tiny\bfseries, anchor=center, minimum height=0.45cm},
    layerh/.style={font=\tiny\bfseries, anchor=east, text width=2.0cm, align=right}
]

\foreach \x/\name in {1/TLS, 2/Auth, 3/FHSS, 4/M-Link, 5/ML-IDS, 6/Fusion, 7/TDPC, 8/VR-Def, 9/DT, 10/ZTA, 11/MTD, 12/RV} {
    \node[header, rotate=60, anchor=west] at (\x*0.65+0.6, 0.4) {\name};
}

\foreach \y/\name in {0/L6:Platform, -1/L5:Localization, -2/L4:Control, -3/L3:Communication, -4/L2:Operator, -5/L1:Perception} {
    \node[layerh] at (0.55, \y*0.5-0.2) {\name};
}

\foreach \x/\trl in {1/9, 2/9, 3/8, 4/7, 5/5, 6/5, 7/6, 8/4, 9/3, 10/3, 11/3, 12/4} {
    \node[cell, font=\tiny\bfseries] at (\x*0.65+0.6, -3.2) {\trl};
}
\node[layerh] at (0.55, -3.2) {TRL};

\newcommand{\hmcell}[3]{%
  \fill[#3] (#1*0.65+0.6-0.32, #2-0.22) rectangle (#1*0.65+0.6+0.32, #2+0.22);
  \draw[gray!40] (#1*0.65+0.6-0.32, #2-0.22) rectangle (#1*0.65+0.6+0.32, #2+0.22);
}



\foreach \x/\col in {1/yellow!30,2/green!50,3/white,4/yellow!30,5/green!50,6/white,7/yellow!30,8/white,9/yellow!30,10/green!50,11/yellow!30,12/yellow!30} {
  \hmcell{\x}{-0.2}{\col}
}

\foreach \x/\col in {1/white,2/white,3/white,4/white,5/yellow!30,6/green!50,7/white,8/white,9/yellow!30,10/white,11/white,12/yellow!30} {
  \hmcell{\x}{-0.7}{\col}
}

\foreach \x/\col in {1/yellow!30,2/green!50,3/white,4/white,5/green!50,6/yellow!30,7/green!50,8/white,9/green!50,10/yellow!30,11/green!50,12/green!50} {
  \hmcell{\x}{-1.2}{\col}
}

\foreach \x/\col in {1/green!50,2/green!50,3/green!50,4/green!50,5/green!50,6/white,7/yellow!30,8/white,9/yellow!30,10/green!50,11/yellow!30,12/yellow!30} {
  \hmcell{\x}{-1.7}{\col}
}

\foreach \x/\col in {1/white,2/white,3/white,4/white,5/white,6/white,7/white,8/green!50,9/white,10/white,11/white,12/yellow!30} {
  \hmcell{\x}{-2.2}{\col}
}

\foreach \x/\col in {1/white,2/white,3/white,4/white,5/yellow!30,6/green!50,7/white,8/yellow!30,9/yellow!30,10/white,11/white,12/yellow!30} {
  \hmcell{\x}{-2.7}{\col}
}

\draw[thick] (0.93, -2.98) -- (8.95, -2.98);

\fill[green!50] (1.2,-3.8) rectangle (1.6,-3.55);
\node[font=\tiny, anchor=west] at (1.65,-3.68) {Strong ($\bullet\bullet$)};
\fill[yellow!30] (3.4,-3.8) rectangle (3.8,-3.55);
\node[font=\tiny, anchor=west] at (3.85,-3.68) {Partial ($\bullet$)};
\fill[white] (5.8,-3.8) rectangle (6.2,-3.55);
\draw[gray!40] (5.8,-3.8) rectangle (6.2,-3.55);
\node[font=\tiny, anchor=west] at (6.25,-3.68) {None/NA (--)}; 

\end{tikzpicture}
\end{adjustbox}
\caption{Defense coverage heatmap across attack layers (L1--L6) and defense mechanisms. Cell colors encode the same effectiveness levels as Table~\ref{tab:defense_effectiveness}: green denotes strong coverage ($\bullet\bullet$), yellow denotes partial coverage ($\bullet$), and white denotes no coverage or non-applicability (--).}
\label{fig:defense_heatmap}
\end{figure}

Fig.~\ref{fig:defense_heatmap} visualizes the same matrix, emphasizing the maturity gap between network/control protections and the comparatively immature defenses for perception-, operator-, and localization-facing threats.

\textbf{Critical Observation on Layer 5 (Localization) Deficiency:} The defense coverage matrix reveals a systematic gap in localization-layer protection. Unlike autonomous vehicles where GPS/GNSS defense has received substantial research attention~\cite{gps2024pmc}, quadruped deployments frequently operate in GPS-denied environments (indoor facilities, urban canyons, subterranean spaces) where alternative localization modalities---visual odometry, LiDAR SLAM, and leg odometry---become primary. These modalities lack the mature defense ecosystem developed for GNSS, and their quadruped-specific vulnerabilities (gait-induced drift, contact-dependent observability, terrain-coupled uncertainty) remain largely uncharacterized. The only partial coverage in Layer~5 derives from sensor fusion consistency checks originally developed for autonomous vehicles, whose transferability to legged platforms with fundamentally different motion profiles is unvalidated. This gap is particularly consequential because localization corruption propagates bidirectionally: upstream to operator situational awareness (via corrupted map displays) and downstream to foothold planning and trajectory execution.

Three observations follow from this assessment. First, the matrix exhibits a clear maturity gradient: defenses at TRL $\geq$ 7 predominantly protect Layers 3--4 (communication and command/control integrity), whereas Layers 1, 2, and 5 (perception, operator interface, and localization) rely mainly on TRL 3--5 techniques whose robustness under quadruped-specific dynamics and teleoperation constraints remains insufficiently validated. Consequently, the practical protection envelope of a deployed quadruped system is narrower than a layer-agnostic checklist would suggest, because only the higher-TRL subset is immediately operationalizable with predictable behavior.

Second, Layer 2 (VR/AR operator interface) constitutes the most pronounced coverage deficit. The only meaningful protections in this layer currently correspond to laboratory-validated monitoring and limited runtime checks (TRL $\leq$ 4), with minimal evidence of effectiveness under sustained teleoperation workloads. Given the tight coupling between operator perception and control decisions, this gap is not merely a cybersecurity issue but a direct contributor to cyber-physical safety risk.

Third, the absence of any single mechanism providing uniformly strong coverage across all layers confirms that quadruped security is inherently defense-in-depth. However, comprehensive coverage today typically requires combining field-proven controls (e.g., TLS, authentication, spectrum resilience) with immature components (e.g., perception and operator-layer detectors), introducing reliability uncertainty and potential false-positive-induced operational hazards. Until Tier-2 mechanisms mature, residual risk in perception-, operator-, and localization-facing layers must be explicitly managed through operational constraints (terrain policies, gait-aware fail-safes, and supervised modes) rather than assumed to be eliminated by technical controls alone.

\subsection{Performance Overhead Analysis}

Table~\ref{tab:overhead_analysis} delineates typical performance overheads for security systems in relation to the time limitations of quadruped teleoperation. The key interpretation is not the absolute numbers alone, but whether the overheads are \emph{validated on operational robotic systems} (higher TRL) versus extrapolated from adjacent domains (lower TRL). The differentiation in maturity is significant as legged teleoperation enforces stricter and more state-dependent latency constraints compared to conventional mobile robots or IoT implementations.

\begin{table}[htbp]
\caption{Security Mechanism Performance Overhead}
\label{tab:overhead_analysis}
\centering
\footnotesize
\begin{adjustbox}{max width=\columnwidth}
\begin{tabular}{|l|c|c|c|c|c|}
\hline
\textbf{Mechanism} & \textbf{Latency} & \textbf{BW} & \textbf{CPU} & \textbf{TRL} & \textbf{Ref.} \\
\hline
TLS 1.3 (handshake) & 100--150ms & +2\% & +3\% & 9 & \cite{radhakrishnan2024lightweight} \\
DTLS per-packet & 1--3ms & +5\% & +5\% & 9 & \cite{radhakrishnan2024lightweight} \\
ASCON encryption & $<$1ms & +8\% & +2\% & 6 & \cite{radhakrishnan2024lightweight} \\
ML IDS inference & 8--15ms & -- & +8\% & 5 & \cite{holdbrook2024ids} \\
Freq. hopping & 5--20ms & -10\% & +2\% & 8 & \cite{pirayesh2022jamming} \\
Multi-sensor fusion & 10--50ms & -- & +15\% & 5 & \cite{sun2020carlo} \\
Digital twin sync & 5--20ms* & +3\% & +12\%* & 3 & \cite{xu2021digitaltwin} \\
Runtime verif. & $<$5ms & -- & +3\% & 4 & \cite{huang2014rosrv} \\
\hline
\end{tabular}
\end{adjustbox}

\vspace{2mm}
\scriptsize{\textbf{Note:} *Estimated from IoT/CPS benchmarks and simulation studies; not measured on quadruped robotic compute stacks. TRL enables distinguishing operationally validated measurements (TRL $\geq$7) from laboratory/theoretical estimates (TRL $<$7).}
\end{table}

Practical Deployment Suggestion: For mission-critical quadruped teleoperation, a robust baseline includes DTLS for control and video telemetry transfer, strong authentication (JWT/X.509) for operator and client identity, and FHSS or comparable spectrum agility for anti-jamming purposes. These collectively offer strong Layer 3-4 security utilizing field-proven technology (TRL 7-9), generally introducing approximately 7-25ms of end-to-end latency and around 10\% CPU overhead when used with session resumption and efficient cipher suites. Enhancing coverage in Layers 1 and 6 using a machine learning-based Intrusion Detection System can augment attack visibility; nevertheless, it incurs an extra inference delay of 8 to 15 milliseconds and around 8\% CPU load, with the significant caveat that detection efficacy under quadruped-specific traffic and gait-dependent dynamics has not been proven.
\section{Comparative Analysis}

\subsection{Cross-Domain Security Comparison}

Table~\ref{tab:cross_domain_comparison} situates quadruped teleoperation security inside the wider robotics landscape by contrasting four principal domains across fundamental security parameters. The objective is not to assert uniformity across each domain, but to emphasize areas where institutional pressure (regulation, liability, certification) has historically elevated security maturity—and where it has not.

\begin{table}[htbp]
\caption{Cross-Domain Security Comparison}
\label{tab:cross_domain_comparison}
\centering
\footnotesize
\begin{adjustbox}{max width=\columnwidth}
\begin{tabular}{|p{2.2cm}|p{2.35cm}|p{2.35cm}|p{2.35cm}|p{2.35cm}|}
\hline
\textbf{Aspect} & \textbf{Quadrupeds} & \textbf{UAVs} & \textbf{Industrial Robots} & \textbf{Autonomous Vehicles} \\
\hline
Typical encryption posture & Variable; often absent or partially enabled in research deployments & Common for command links; vendor-specific stacks; increasing use of strong crypto & Common within modern industrial stacks (e.g., OPC UA security profiles) & Broad use of TLS 1.3+ and authenticated links in modern architectures \\
\hline
Typical authentication posture & Variable; default credentials and weak provisioning still observed on some platforms & Proprietary pairing / vendor account models; mixed rigor across COTS vs defense UAVs & PKI/role-based access common in industrial security programs & Strong identity anchoring (TPM/HSM usage common; certificate-based ecosystems) \\
\hline
Dominant threat pressure & Jamming, command injection, operator deception, supply-chain exposure & GNSS spoofing, jamming, link takeover, C2 compromise & PLC/SCADA compromise, safety interlock bypass, OT lateral movement & V2X misuse, sensor spoofing, ECU compromise, OTA/update channel abuse \\
\hline
Commercial IDS availability & \textbf{Sparse / non-specialized}; mostly generic IT/OT IDS adaptations & Commercial monitoring exists for many UAV ecosystems; mixed depth & Mature OT security market and monitoring toolchains & Extensive ecosystem (vehicle SOC, fleet telemetry, anomaly monitoring) \\
\hline
Regulatory / compliance drivers & \textbf{None domain-specific}; typically internal policy only & FAA-driven compliance pressures (e.g., Remote ID) and safety certification in specific contexts & IEC 62443 and related OT standards shape procurement and certification & UNECE WP.29 (R155/R156) and homologation pressure drive security processes \\
\hline
Research maturity & \textbf{Low} and fragmented; limited quadruped-teleop focus & High; long-standing EW/jamming/spoofing literature & Medium; strong OT security literature, less robotics-specific nuance & Very high; large cross-disciplinary security community \\
\hline
Failure consequence profile & Critical: proximate-to-human, kinetic energy, unstable fall modes & High: airspace risk, payload risk; many safe modes exist & High: physical damage and safety risk, mitigated by safety cells/interlocks & Critical: public safety, large-scale liability and recall risk \\
\hline
Standards and certification culture & \textbf{Absent} for legged-teleop security; ad hoc vendor practices & Active and evolving & Established and procurement-driven & Emerging but rapidly institutionalizing \\
\hline
\end{tabular}
\end{adjustbox}

\vspace{2mm}
\scriptsize{\textbf{Note:} Quadruped teleoperation exhibits the weakest external ``security forcing function'' (regulation, certification, liability norms) among major robotic domains, despite consequence severity comparable to autonomous vehicles. This mismatch helps explain why deployable defenses and standardized assurance practices lag behind adjacent domains \cite{wanguavswarm2024, botta2023survey, chi2025adversarial}.}
\end{table}

The comparison reveals a structural disparity: sectors engaging with regulated public space (autonomous vehicles, UAVs) or established operational technology (industrial robotics) face persistent external pressure that progressively elevates security standards. The security of autonomous vehicles, for example, is enhanced by significant academic investment, regulatory progress (UNECE WP.29), and direct liability from manufacturers \cite{chi2025adversarial}. UAV ecosystems similarly adopt established GNSS/jamming threat models and regulatory compliance demands that have prompted commercial fortification and surveillance. Industrial robots, traditionally considered "air-gapped," is progressively integrating with OT security programs influenced by IEC 62443, resulting in more defined security requirements and assurance driven by procurement.

Quadruped teleoperation exists in a security void, lacking a specific regulatory framework, standardized certification process, and facing no market repercussions for inadequate security measures in non-defense applications. Consequently, security posture fluctuates significantly across platforms and deployments, transferring the responsibility of assurance from institutions to individual integrators—frequently lacking the competence or resources to implement comprehensive controls across.

A more nuanced yet operationally critical distinction is the \emph{failure mode}. Unmanned Aerial Vehicles (UAVs) are capable of hovering or performing return-to-home maneuvers; wheeled robots often halt without compromising stability; industrial robotic arms function within safety enclosures using built emergency stop protocols. Quadrupeds do not possess a comparable secure default condition during dynamic movement. An emergency stop while navigating stairs or transitioning across terrains can lead to an uncontrolled fall instead of a safe halt. Thus, security measures adapted from other domains may be \emph{functionally unsafe} if they cause sudden interruptions or excessive latency during critical periods of gait, despite being cryptographically secure.

\subsection{Maturity Gap Analysis}

Section IV's TRL stratification demonstrates a systematic maturity disparity between attack tiers. Communication defenses (Layer 3) leverage decades of study on network security and consistently achieve TRL~7--9. Control-plane safeguards (Layer 4) are somewhat mitigated using existing mechanisms (encryption, authentication) and stability-preserving teleoperation control designs (e.g., passivity layers), resulting in TRL ranges from 6 to 9, contingent upon the implementation and validation circumstances.

Nonetheless, the layers most closely associated with human safety and situational awareness are still underdeveloped. Perception-layer defenses (Layer 1) predominantly maintain a Technology Readiness Level of 4 to 5, as most assessments were performed on autonomous vehicles featuring varying sensor configurations, motion trajectories, and ambient structures. Protection for VR/AR operators (Layer~2) generally exhibits a Technology Readiness Level (TRL) of approximately 3 to 4, predominantly supported by laboratory research rather than real-time robot teleoperation in hostile environments. Ultimately, quadruped-specific innovations—such as gait-phase-aware security protocols and locomotion signature authentication—remain at TRL~1--2, indicating an absence of datasets, benchmarks, and field experiments.

This maturity distribution is not merely an intellectual nuisance; it constitutes the primary practical limitation. Currently, effective security for real deployments is primarily attainable for Layers 3 and 4, whereas residual risks in Layers 1, 2, and 5 must be mitigated through operational controls—such as mission constraints, environmental conditioning, operator training, and conservative fallback strategies—until higher Technology Readiness Level (TRL) defenses are developed.
To quantify the maturity deficit, we define the \textit{layer-specific security readiness index} (SRI) as the ratio of the maximum available TRL for defenses addressing that layer to the minimum TRL required for operational deployment (TRL 7):
\begin{equation}
\text{SRI}_\ell = \frac{\max_{d \in \mathcal{D}_\ell} \text{TRL}(d)}{7}
\end{equation}
where $\mathcal{D}_\ell$ denotes the set of defense mechanisms applicable to layer $\ell$. Layers with SRI $\geq 1.0$ are considered deployment-ready; layers with SRI $< 1.0$ exhibit a maturity deficit proportional to $(1 - \text{SRI}_\ell)$. Applying this metric: Layer 3 (Communication) achieves SRI = 1.29, Layer 4 (Control) achieves SRI = 1.29, and Layer 6 (Platform) achieves SRI = 1.29---all deployment-ready. In contrast, Layer 1 (Perception) achieves SRI = 0.71, Layer 2 (Operator) achieves SRI = 0.57, and Layer 5 (Localization) achieves SRI = 0.71---all exhibiting significant maturity deficits. The aggregate system SRI, computed as the harmonic mean across layers, is 0.88, indicating that the overall quadruped teleoperation security posture remains below deployment-ready maturity when all attack surfaces are considered.

\begin{figure}[htbp]
\centering
\begin{tikzpicture}
\begin{axis}[
    ybar,
    bar width=12pt,
    ylabel={Maximum Defense TRL},
    symbolic x coords={L1, L2, L3, L4, L5, L6},
    xtick=data,
    xlabel={Attack Layer},
    ymin=0, ymax=10,
    ytick={0,2,4,6,8,10},
    extra y ticks={3,7},
    extra y tick labels={},
    extra y tick style={grid=major, grid style={dashed, red!50}},
    nodes near coords,
    every node near coord/.append style={font=\scriptsize},
    width=0.45\textwidth,
    height=0.3\textwidth,
    enlarge x limits=0.15,
]
\addplot[fill=blue!40] coordinates {
    (L1,5) (L2,4) (L3,9) (L4,9) (L5,5) (L6,9)
};
\end{axis}
\node[font=\tiny, red] at (1.1,2.7) {Lab};
\node[font=\tiny, green!60!black] at (3.6,3.7) {Field};
\end{tikzpicture}
\caption{Maximum defense maturity (TRL) by attack layer. Dashed lines indicate TRL 3 (proof-of-concept) and TRL 7 (field-deployable) thresholds. Layers 1, 2, and 5 lack field-deployable defenses entirely.}
\label{fig:maturity_gap}
\end{figure}
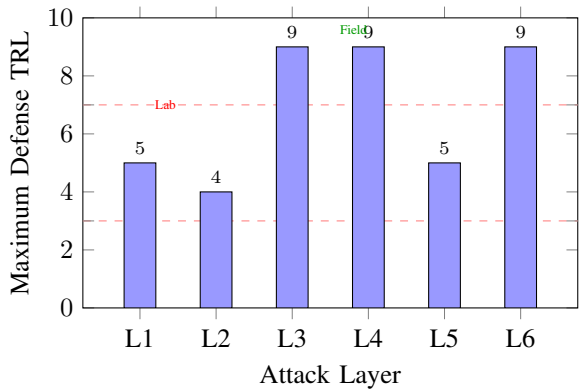

The maturity gap has a direct operational impact: organizations utilizing quadrupeds can enhance communication and control channels through established, field-tested mechanisms, but possess limited defenses of similar maturity against perception manipulation, operator-targeting attacks, or localization spoofing. Consequently, a credible security posture cannot rely solely on "coverage"; it must be founded on what is genuinely implementable. The defense strategy should consequently prioritize high-TRL protections as the foundation (secure transport, robust authentication, anti-jamming resilience, and fortified control-plane integrity), while explicitly addressing residual risk in lower-maturity layers through conservative operational constraints, context-aware fail-safe behaviors, and continuous monitoring, rather than presuming that laboratory-stage perception or VR/AR defenses will consistently translate to legged teleoperation in practical applications.

\subsection{Investment–Consequence Disparity}

The cross-domain comparison and maturity-gap analysis reveal what we refer to as the \textit{investment–consequence disparity}: a structural incongruity between (i) the severity of potential outcomes from successful attacks and (ii) the extent of research, tooling, and engineering resources allocated to mitigating those outcomes. The gap is particularly evident in Layers 1 (Perception) and 2 (VR/AR), where consequence severities are assessed as High to Critical (Table~\ref{tab:consequence_severity}), yet the most advanced defenses are confined to TRL 4–5.

A tangible illustration of this disparity is the juxtaposition with driverless automobiles. Perception security for autonomous vehicles, vulnerable to similar attack vectors from LiDAR and cameras, has resulted in several articles, established benchmarking procedures, and several commercial-grade detection and validation pipelines \cite{chi2025adversarial, kim2024lidar}. In contrast, quadruped perception security is predominantly pre-paradigmatic: despite the complexities introduced by gait-phase-dependent vulnerability windows and narrower recovery margins, the field lacks a widely accepted quadruped-specific dataset, a broadly validated perception-attack detection framework, and virtually no published end-to-end validation on a legged platform under realistic teleoperation conditions.

The most likely root problem is economic rather than technical. The autonomous vehicle sector---which attracts substantially greater commercial and regulatory security investment than quadruped robotics---may distribute significant security research and development costs over extensive fleets and liability-sensitive applications, but the commercial quadruped market is smaller and more fragmented. Quadrupeds are being deployed in military, public safety, and critical infrastructure contexts where the severity of consequences is disproportionately large compared to unit volumes. This mismatch suggests that market forces alone are improbable to bridge the gap. It instead encourages focused public research funding, collaborative testbeds, and industry consortia aimed at high-stakes layers (Perception and VR/AR), with specific deliverables including benchmark datasets, reference implementations, and reproducible validation protocols that can elevate defenses from TRL 4–5 to deployable maturity.

While precise quantification of the investment gap falls outside 
the scope of this survey, the qualitative disparity is evident: 
autonomous vehicle security has benefited from decades of 
dedicated research programs, industry consortia, and regulatory 
mandates, whereas quadruped security research remains in its 
infancy with no equivalent institutional infrastructure.
\section{Open Challenges and Research Gaps}
\label{sec:gaps}
The research gaps identified in this section are prioritized according to a composite scoring framework integrating three analytical dimensions. The first dimension, \textit{consequence severity}, derives from the attack-to-consequence mapping established in Section~\ref{sec:impact_analysis}, employing the maximum severity rating across affected impact categories for each gap. The second dimension, \textit{maturity deficit}, quantifies the difference between the required Technology Readiness Level for operational deployment (TRL 7) and the current maximum TRL documented for applicable defense mechanisms. The third dimension, \textit{research tractability}, provides a qualitative assessment of near-term progress feasibility based on prerequisite infrastructure availability, including benchmark datasets, experimental testbeds, and theoretical foundations.

Table~\ref{tab:gap_priority} summarizes the prioritization scoring. This framework enables systematic resource allocation while accommodating future refinement as research directions evolve or additional gaps emerge from ongoing threat landscape analysis.

\begin{table}[htbp]
\caption{Research Gap Prioritization Scoring}
\label{tab:gap_priority}
\centering
\footnotesize
\begin{adjustbox}{max width=\columnwidth}
\begin{tabular}{|l|c|c|c|c|}
\hline
\textbf{Research Gap} & \textbf{Severity} & \textbf{Deficit} & \textbf{Tract.} & \textbf{Priority} \\
\hline
Gap 1: Gait-Phase-Aware IDS & Critical & 2 & High & \textbf{High} \\
Gap 2: FDIA Detection & Critical & 2 & Medium & \textbf{High} \\
Gap 3: Recovery Control & Critical & 4 & Medium & \textbf{High} \\
Gap 4: Perception Defense & High & 2 & Medium & Medium-High \\
Gap 5: Operator Monitoring & High & 3 & Low & Medium \\
Gap 6: Locomotion-Aware Comm & Medium & 2 & High & Medium \\
Gap 7: Cross-Layer Correlation & Variable & 4 & Low & Medium-Low \\
Gap 8: Fleet-Level Security & High & 3 & Medium & Medium \\
\hline
\end{tabular}
\end{adjustbox}

\vspace{2mm}
\scriptsize{\textbf{Note:} Severity derives from maximum consequence rating across affected impact categories (Table~\ref{tab:consequence_severity}). Deficit quantifies the gap between required operational TRL (7) and current maximum documented TRL. Tractability reflects near-term feasibility based on prerequisite infrastructure availability. Priority integrates all three dimensions, with dependencies explicitly noted (Gap~7 contingent on Gaps~1--4; Gap~8 contingent on Gaps~1 and~6).}
\end{table}

Gaps exhibiting Critical consequence severity combined with addressable maturity deficits receive highest priority designation. Cross-layer correlation mechanisms are explicitly contingent on progress in single-layer detection capabilities, as effective multi-layer anomaly fusion requires functional baseline detectors as input signals. This dependency structure implies that the prioritization ordering reflects not merely independent importance rankings but also prerequisite relationships that constrain feasible research sequencing.
The maturity gap analysis and cross-domain comparison in the previous section expose a fundamental tension: quadruped robots encounter consequence severities comparable to those of autonomous vehicles, yet receive substantially less security research investment. More critically, the existing defense landscape reveals that while communication-layer protections (TLS/DTLS, authentication, FHSS) have reached field-deployable maturity (TRL~7--9), the detection, identification, and mitigation mechanisms that must operate \emph{within} the teleoperation control loop---where cyber disruptions translate into physical instability---remain at TRL~1--5, with minimal quadruped-specific validation.

This section synthesizes the preceding analysis into eight prioritized research gaps. Unlike infrastructure-level or policy-oriented recommendations, these gaps target the core algorithmic, control-theoretic, and signal-processing challenges that must be resolved to secure the operator--robot communication and control chain against the threat classes identified in our taxonomy. Each gap is ranked by the severity of consequences (Section~\ref{sec:impact_analysis}), the current maturity deficit (Section~IV), and the feasibility of near-term research progress.

\subsection{Gap 1 (High Priority): Gait-Phase-Aware Intrusion Detection Algorithms}

Current ML-based intrusion detection systems achieve 97--99\% accuracy on general ROS2/IoT traffic (Table~\ref{tab:ids_performance}), but these results are conditioned on datasets where traffic patterns are approximately stationary---an assumption that fundamentally breaks down for quadruped teleoperation. Legged locomotion generates inherently \emph{non-stationary, multi-rate, and phase-coupled} communication patterns: periodic gait cycles modulate the timing and size of joint-state publications at 100--1000~Hz; terrain transitions (flat$\rightarrow$stairs, stance$\rightarrow$slip recovery) produce abrupt changes in message burstiness and control-loop bandwidth demand; and teleoperation mode switches (direct control$\rightarrow$shared autonomy$\rightarrow$VR immersive) alter the statistical profile of the entire command-feedback pipeline. Existing IDS datasets (ROSIDS23 \cite{rosids23}, ROSPaCe \cite{puccetti2024rospace}) contain none of these signatures, and generic anomaly detectors trained on such data are likely to produce unacceptable false-positive rates when exposed to normal quadruped dynamics---or, worse, to miss attacks whose signatures are masked by legitimate locomotion transients.

The algorithmic challenge is twofold. First, the detector must maintain a \emph{phase-conditioned baseline} that adapts to the robot's current locomotion state (gait mode, terrain class, contact configuration) rather than relying on a single global traffic model. This requires explicit coupling between the IDS feature space and the locomotion controller's state machine---a design pattern absent from all published ROS2 IDS architectures \cite{holdbrook2024ids, martin2025rips, puccetti2024rospace}. Second, the detector must achieve \emph{sub-gait-cycle detection latency}: for a trot at 2~Hz, each gait cycle spans approximately 500~ms, and attack detection must occur within a fraction of this window to enable meaningful intervention before the cascade reaches the stability boundary. Standard batch-inference IDS pipelines operating at 8--15~ms per inference \cite{holdbrook2024ids} are fast enough in principle, but their detection \emph{delay} (time from attack onset to confident alert) depends on the number of observations required for statistical confidence---a quantity that is highly sensitive to the non-stationarity of the baseline and has not been characterized for locomotion-coupled traffic.
\textbf{Success Criteria for Gap 1:} A viable gait-phase-aware IDS must satisfy the following quantitative thresholds to be considered operationally effective: (i) detection latency below 250\,ms (approximately half a gait cycle at 2\,Hz trot frequency), enabling intervention before cascading instability; (ii) false positive rate below 1\% per gait phase during normal terrain transitions, avoiding alert fatigue and unnecessary mode switches; (iii) true positive rate exceeding 90\% for Layer~4 (control signal) attacks, which pose the most immediate stability threat; (iv) computational overhead below 10\% of available CPU cycles on representative embedded platforms (e.g., NVIDIA Jetson Orin), ensuring compatibility with concurrent locomotion control; and (v) graceful degradation under partial observability, maintaining detection capability when individual sensor streams are unavailable or corrupted.

\textbf{Inherent Limitations:} Gait-phase-aware IDS faces fundamental tradeoffs that constrain achievable performance. First, the non-stationarity of locomotion-coupled traffic implies that any fixed baseline will generate elevated false positives during legitimate mode transitions; adaptive baselines introduce detection latency proportional to the adaptation time constant. Second, adversaries with gait-phase knowledge can time attacks to coincide with high-variance phases (terrain transitions, gait switches) where detection thresholds must be relaxed, creating \textit{phase-dependent vulnerability windows} that cannot be fully eliminated without unacceptable false-positive rates. Third, the absence of ground-truth attack labels for quadruped traffic precludes supervised learning approaches until annotated datasets become available, limiting current methods to unsupervised anomaly detection with inherently higher false-positive rates.

\subsection{Gap 2 (High Priority): Real-Time Detection and Mitigation of False Data Injection in Teleoperation Control Loops}

False data injection attacks (FDIA) represent a uniquely dangerous threat to quadruped teleoperation because they target the mathematical structure of the feedback loop rather than the communication channel itself. The fundamental detection challenge can be formalized as follows: consider a discrete-time teleoperation system with state dynamics $x_{k+1} = Ax_k + Bu_k + w_k$ and measurement equation $y_k = Cx_k + v_k$, where $w_k$ and $v_k$ denote process and measurement noise respectively. An FDIA introduces a malicious signal $a_k$ such that the compromised measurement becomes $\tilde{y}_k = y_k + a_k$. The attacker's objective is to design $a_k$ such that the innovation sequence $r_k = \tilde{y}_k - C\hat{x}_{k|k-1}$ remains statistically indistinguishable from the nominal case, thereby evading $\chi^2$-based detectors that monitor the quadratic form $g_k = r_k^\top S_k^{-1} r_k$ against threshold $\tau$.

For bilateral teleoperation with nonlinear manipulator dynamics, Kwon and Ueda~\cite{kwon2024fdia} established that perfectly undetectable FDIA exists when the attack signal satisfies specific Lie-algebraic conditions on the system's observability codistribution. Their experimental validation across a US--Japan teleoperation link demonstrated that such attacks can persist under realistic network conditions, confirming theoretical predictions. Dong et al. \cite{dong2020fdia} demonstrated that carefully designed injection signals can destabilize joint velocity regulation while maintaining sufficient residual consistency to evade conventional $\chi^2$ detectors. More critically, Kwon and Ueda \cite{kwon2024fdia} established Lie-group-based conditions for \emph{perfectly undetectable} FDIA in bilateral teleoperation with second-order nonlinear dynamics, experimentally confirming feasibility across a US--Japan teleoperation link. Kim et al. \cite{liegroup2024fdia} further showed that such attacks can succeed even when the control pipeline employs encryption, exploiting malleability properties of certain cryptographic schemes.

For quadrupeds, the consequence of undetected FDIA is amplified by the tight coupling between command timing and dynamic stability. An FDIA that introduces structured perturbations to swing-timing signals or touchdown-event estimates---perturbations that may appear as normal sensor noise or minor calibration drift---can progressively degrade balance margins over several gait cycles until the robot enters an irrecoverable state. Standard residual-based detectors achieve only 35--75\% detection rates at 5\% false-alarm rates in typical robotic environments \cite{dieber2020penetration}, and this performance degrades further when the attacker has partial knowledge of the system model.

The core research challenge is developing detection mechanisms that exploit the \emph{physical constraints specific to legged locomotion}---constraints that the attacker cannot simultaneously satisfy while achieving destabilization. Quadruped dynamics impose several such invariants: (i) contact-consistent force/torque relationships at each foot (ground reaction forces must be consistent with the robot's mass, acceleration, and contact geometry), (ii) gait-phase sequencing constraints (the temporal ordering of swing-to-stance transitions across legs must satisfy kinematic feasibility), (iii) energy conservation bounds (the total mechanical energy must evolve consistently with actuator work and dissipation), and (iv) inter-leg coordination symmetry (gait patterns impose predictable phase relationships between legs that FDIA targeting a single leg would violate).

\subsection{Gap 3 (High Priority): Stability-Preserving Recovery Control Under Cyber Attack}

A critical and largely unaddressed deficiency in the current defense landscape is the absence of \emph{recovery mechanisms} that maintain dynamic stability when an attack is detected. The recovery control problem can be formulated within a switched-system framework: let $\mathcal{M} = \{m_1, \ldots, m_N\}$ denote a finite set of locomotion modes (e.g., trot, walk, stance, controlled descent), each characterized by hybrid dynamics $\dot{x} = f_{m_i}(x, u)$ with guard conditions $\mathcal{G}_{ij}$ governing transitions. Upon attack detection at time $t_d$, the recovery controller must identify a mode sequence $\sigma: [t_d, t_d + T] \rightarrow \mathcal{M}$ and corresponding control policy $\pi_\sigma$ such that the system trajectory $x(t)$ remains within a safe invariant set $\mathcal{S}$ and converges to a stable equilibrium $x^* \in \mathcal{S}$ within horizon $T$.

The challenge is compounded by the hybrid contact dynamics of legged locomotion: each mode $m_i$ corresponds to a distinct contact configuration with associated kinematic constraints and force closure conditions. A recovery primitive that is valid in stance phase (all feet grounded) may be infeasible during swing phase when the support polygon is reduced. Consequently, recovery controller synthesis must be \emph{gait-phase-aware}, with mode-dependent control barrier functions (CBFs) that encode phase-specific safety constraints:
\begin{equation}
\dot{h}_{m_i}(x, u) + \alpha(h_{m_i}(x)) \geq 0, \quad \forall x \in \partial\mathcal{S}_{m_i}
\end{equation}
where $h_{m_i}$ is the barrier function for mode $m_i$, $\mathcal{S}_{m_i}$ is the mode-dependent safe set, and $\alpha$ is an extended class-$\mathcal{K}$ function. The synthesis problem is to find a switching policy and associated CBF-constrained controllers that guarantee forward invariance of $\bigcup_i \mathcal{S}_{m_i}$ under adversarial disturbances bounded by the attack model. The standard robotic response to a detected anomaly---emergency stop (e-stop)---is predicated on the assumption of static or quasi-static stability. For a quadruped traversing stairs, narrow ledges, or uneven terrain, an abrupt actuator shutdown can cause an \emph{uncontrolled fall}, transforming the defense mechanism itself into a safety hazard. Zhang et al. \cite{zhang2024fall} demonstrated that unmitigated falls produce impact velocities approximately 3$\times$ higher than controlled descent, confirming that the manner of failure termination directly determines consequence severity.

This gap sits at the intersection of control theory and cybersecurity: the challenge is not merely detecting an attack but transitioning the robot to a \emph{dynamically stable safe state} that is compatible with the current contact configuration, terrain geometry, and gait phase---all within the sub-second window between detection and physical consequence. Existing passivity-based teleoperation controllers (TDPC) \cite{risiglione2021passivity} provide delay tolerance up to 500~ms but were not designed to handle adversarial control-signal corruption; they guarantee stability under bounded delay but not under malicious command injection or perception manipulation. PID-Piper \cite{dash2021pidpiper} addresses post-attack recovery for compromised sensors and actuators but was validated on wheeled/aerial platforms where static stability permits graceful degradation.

\subsection{Gap 4 (Medium--High Priority): Perception Attack Detection Adapted for Legged Platform Dynamics}

LiDAR and camera spoofing defenses (e.g., CARLO \cite{sun2020carlo}, Shadow-Catcher \cite{cao2023pra}) were developed and validated exclusively on autonomous vehicle platforms with roof-mounted sensors operating in predominantly planar road environments. Quadruped platforms present three fundamental differences that invalidate direct transfer of these methods.

First, \textbf{sensor placement and motion profile}: quadruped-mounted sensors experience substantial platform oscillation (pitch/roll excursions of $\pm$5--15$^\circ$ during trot), periodic vertical displacement from body bounce, and impulsive vibration at foot-ground contact. These dynamics alter point-cloud statistics, shadow geometries, and occlusion patterns in ways that are \emph{normal} for legged locomotion but would trigger false alarms in AV-calibrated detectors. Second, \textbf{operating environment geometry}: quadrupeds navigate complex 3D environments (staircases, pipe racks, scaffolding, narrow corridors) where occlusion patterns, reflective surfaces, and multi-path interference differ fundamentally from road scenarios. Third, \textbf{consequence coupling}: for a wheeled vehicle, a brief perception error typically results in a braking event; for a quadruped, a centimeter-scale depth error during stair climbing directly affects foothold selection and can trigger a balance-loss cascade within a single gait cycle ($<$500~ms).

The detection challenge is therefore not merely computational (adapting algorithms to new data) but \emph{signal-theoretic}: distinguishing adversarial perception manipulation from the rich, structured noise inherent in legged locomotion. Hallyburton et al. \cite{hallyburton2022cameralidar} demonstrated that camera--LiDAR fusion provides inherent resilience against naive spoofing ($<$5\% success), but sophisticated tracking attacks remain effective---suggesting that fusion helps but does not eliminate the threat without explicit adversarial modeling.

\subsection{Gap 5 (Medium Priority): Operator Degradation Detection and Adaptive Autonomy Transfer}

Attacks targeting the human operator---cybersickness induction, display content injection, latency manipulation, and HMD side-channel exploitation---constitute a distinct threat class because the ``compromised component'' is not a software module but the human-in-the-loop controller. Current cybersickness detection achieves 91.1\% F1-score using Bi-LSTM models \cite{yalcin2024cybersickness}, but this result was obtained in seated VR experiments, not during active teleoperation where head-motion patterns, cognitive load, and physiological baselines differ substantially. Moreover, Kundu et al. \cite{kundu2025cybersicknessattack} demonstrated that adversarial examples can degrade LSTM-based detection accuracy by 4.65$\times$ and Transformer-based detection by 5.94$\times$, indicating that detection systems themselves may be compromised.

The research challenge extends beyond symptom detection to \emph{closed-loop response}: when operator impairment is detected, the system must autonomously transfer control authority from the degraded human to the robot's local autonomy stack---smoothly, safely, and without creating a new vulnerability window. This autonomy transfer must be (i) \emph{graduated} (proportional to detected impairment severity rather than binary), (ii) \emph{stability-preserving} (the transition itself must not destabilize the robot), (iii) \emph{attack-resistant} (an adversary should not be able to trigger autonomy transfer as a means of removing the human from the loop and then exploiting the less-capable autonomous fallback), and (iv) \emph{reversible} (the operator must be able to reclaim authority once the impairment is resolved).

\subsection{Gap 6 (Medium Priority): Locomotion-Aware Adaptive Communication Security Protocols}

Current communication security for quadruped teleoperation treats all traffic uniformly: the same encryption, QoS policy, and redundancy level is applied regardless of the robot's physical state. This one-size-fits-all approach creates two problems. First, during stability-critical locomotion phases (stair ascent, obstacle traversal, slip recovery), the \emph{consequence} of a communication disruption is drastically higher than during steady-state walking on flat ground---yet the communication subsystem allocates identical protection. Second, the overhead of security mechanisms (1--3~ms DTLS per-packet processing, 5--20~ms FHSS hop transitions) is \emph{constant}, even during phases where the control loop can tolerate no additional latency.

The fundamental research question is how to design communication protocols that are \emph{explicitly aware of locomotion state} and dynamically allocate security resources accordingly. This requires a formal framework linking physical stability margins to communication QoS requirements---a coupling that, to the best of our knowledge, has not been established for any legged robotic system.

\subsection{Gap 7 (Medium-Low Priority, Contingent): Cross-Layer Anomaly Correlation for Coordinated Attack Detection}

The cross-layer coordinated attack scenario analyzed in Section~III demonstrates that a capable adversary can engineer \emph{sub-threshold} manipulations across multiple layers---perception, operator interface, and control---that individually remain below single-layer detection thresholds but collectively produce a high-probability destabilization. This creates a detection problem that is fundamentally different from single-layer IDS: the signal of interest is not an anomaly in any one stream, but a \emph{statistical dependency} across streams that should be independent under normal operation.

Existing anomaly detection in robotics operates in layer-specific silos: network IDS monitors traffic patterns \cite{holdbrook2024ids, martin2025rips}, perception integrity checks validate sensor data \cite{sun2020carlo, xu2025voattack}, and control-loop monitors track state-estimation residuals \cite{dong2020fdia}. No published framework correlates weak signals across these layers to detect coordinated attacks. The digital-twin paradigm \cite{xu2021digitaltwin, erceylan2025dtcps} provides a conceptual foundation for cross-domain consistency checking, but existing implementations address single-layer anomalies (e.g., joint-trajectory deviations \cite{wang2024dtrobot}) rather than multi-layer correlation patterns.

This gap is rated Medium--Low (contingent) because meaningful cross-layer correlation 
explicitly depends on progress in Gaps~1--4; without  functional single-layer detectors as prerequisites,  correlation-based detection lacks the input signals  necessary for joint statistical testing. Accordingly, this gap is best pursued as an integration effort once  foundational detection capabilities mature.

\subsection{Gap~8: Fleet-Level Security and Wormable Exploit Containment}
\label{sec:gap8}
\label{sec:ble_attacks}

The demonstrated feasibility of wormable BLE exploits capable of lateral propagation across quadruped fleets~\cite{makris2025unipwn} introduces systemic risk that single-robot defenses cannot address. Current security architectures treat each robot as an isolated entity, lacking fleet-wide anomaly correlation, coordinated quarantine mechanisms, or cryptographic segmentation between fleet members. The absence of federated intrusion detection systems means that a single compromised robot can serve as a persistent beachhead for lateral movement across an entire operational deployment.

\textbf{Propagation Model:} Fleet-level exploit spread can be modeled as an epidemic process on a contact graph $G = (V, E)$ where vertices represent robots and edges represent communication proximity (BLE range, shared network). Let $S_i(t) \in \{0, 1\}$ denote the compromise state of robot $i$ at time $t$. The propagation dynamics follow:
\begin{equation}
P(S_i(t+\Delta t) = 1 \mid S_i(t) = 0) = 1 - \prod_{j \in \mathcal{N}(i)} (1 - \beta \cdot S_j(t) \cdot \Delta t)
\end{equation}
where $\beta$ is the transmission rate (successful exploit attempts per unit time) and $\mathcal{N}(i)$ is the set of robots within communication range of robot $i$. The UniPwn exploit demonstrated $\beta \approx 0.1$--$0.5$ attempts/second under favorable conditions~\cite{makris2025unipwn}.

\textbf{Containment Objective:} Given initial compromise at $t_0$, the defense objective is to minimize the expected number of compromised robots $\mathbb{E}[\sum_i S_i(t_0 + T)]$ within response window $T$. Effective containment requires: (i) detection latency $\tau_d < 1/\beta$ to identify compromise before secondary spread, (ii) isolation mechanisms that reduce effective $\beta$ by $>$90\% within $\tau_d + \tau_r$ (response time), and (iii) fleet-wide alerting that warns uncompromised robots to enter defensive posture.

\textbf{Success Criteria:} Containment of wormable exploits to $\leq$1 additional robot within 30~seconds of initial compromise; fleet-wide anomaly detection with $\geq$90\% true positive rate for coordinated attacks; isolation response time $\tau_r < 10$~seconds from detection to network segmentation.

\textbf{Priority:} Medium (contingent on Gaps~1 and~6).

\subsection{Evaluation Blueprint for Future Research}

To facilitate systematic progress on the identified gaps, Table~\ref{tab:benchmark_blueprint} provides an evaluation blueprint specifying observable signals, quantitative metrics, test conditions, and minimum instrumentation requirements for each major attack class. This blueprint is intended to promote methodological consistency across future studies and enable meaningful cross-study comparison of defense effectiveness.

\begin{table*}[htbp]
\caption{Evaluation Blueprint for Quadruped Teleoperation Security Research}
\label{tab:benchmark_blueprint}
\centering
\footnotesize
\begin{tabular}{|p{2.0cm}|p{2.8cm}|p{3.2cm}|p{3.0cm}|p{2.8cm}|}
\hline
\textbf{Attack Class} & \textbf{Observable Signals} & \textbf{Metrics} & \textbf{Test Conditions} & \textbf{Instrumentation} \\
\hline
LiDAR Spoofing & Point cloud anomalies, depth inconsistency, object hallucination & Detection rate (\%), FP/hour, time-to-detection (ms) & Flat/stairs/outdoor; trot/walk gaits & LiDAR + RGB camera + IMU \\
\hline
FDIA & State estimation residuals, joint torque deviation, gait timing errors & TTD (ms), stability margin reduction (\%), fall rate & Multiple gaits, varied terrain & Joint encoders + IMU + F/T sensors \\
\hline
Cybersickness & Head motion patterns, gaze instability, HRV changes & Symptom onset (min), SSQ score, task error rate & VR interface, 30+ min sessions & Eye tracker + head IMU + HRV monitor \\
\hline
Jamming/DoS & Link quality (RSSI), packet loss, latency spike & Recovery time (ms), control continuity (\%), fall rate & Dynamic locomotion, stairs, multi-link & RF spectrum analyzer + robot telemetry \\
\hline
GPS Spoofing & Position jumps, IMU/GPS inconsistency, heading drift & Detection accuracy (\%), max deviation (m) & Outdoor, dynamic motion & Multi-GNSS + IMU + VO \\
\hline
Command Injection & Unauthorized command detection, trajectory deviation & Detection rate (\%), response latency (ms) & Teleoperation active, varied autonomy & Command logger + state monitor \\
\hline
\end{tabular}

\vspace{2mm}
\scriptsize{\textbf{Note:} TTD = Time-to-Detection; FP = False Positive; SSQ = Simulator Sickness Questionnaire; F/T = Force/Torque; HRV = Heart Rate Variability. Metrics should be reported with confidence intervals across multiple trials and terrain conditions.}
\end{table*}


\section{Future Research Directions}
\label{sec:future}

Building on the eight prioritized research gaps identified in 
Section~\ref{sec:gaps}, this section presents concrete research 
directions organized along three temporal horizons. Each direction 
is explicitly linked to its originating gap(s) to maintain 
traceability between identified deficiencies and proposed solutions.
Near-term directions (1--3 years) target engineering validation 
of existing techniques under quadruped-specific conditions, where 
the primary constraint is not algorithmic novelty but domain-specific 
testing. Medium-term directions (3--5 years) require novel algorithmic 
contributions specifically designed for---not merely adapted 
to---the hybrid, phase-dependent dynamics of legged locomotion. 
Long-term directions (5--10 years) address foundational theory, 
formal assurance, and systemic infrastructure that will underpin 
next-generation secure teleoperation architectures.

\subsection{Near-Term (1--3 Years): Engineering and Validation}

The most impactful near-term work involves validating and adapting 
existing security mechanisms under quadruped-specific operating 
conditions---particularly the non-stationary communication patterns, 
sub-second stability-critical windows, and gait-phase-dependent 
failure modes that distinguish legged teleoperation from other 
robotic domains. The primary constraint across all near-term 
directions is not algorithmic novelty but domain-specific 
validation: many promising techniques exist at TRL~4--6 but have 
never been tested on a legged platform during dynamic locomotion.

\subsubsection{Quadruped IDS Benchmark Dataset and Baseline 
Detectors \textnormal{(enables Gap~1)}}

As the essential prerequisite for all detection-oriented research, 
a multi-institutional data collection campaign should instrument 
representative quadruped platforms (e.g., Unitree Go2, Boston 
Dynamics Spot) across gait modes (walk, trot, bound), terrain 
classes (flat, stairs, rubble, industrial clutter), and 
teleoperation modalities (2D video, stereo, VR, shared autonomy). 
Recordings must include synchronized network captures (pcap), 
ROS2/DDS graph traces (topic rates, QoS events, discovery 
traffic), and robot-state logs (IMU, joint states, contact forces, 
gait-phase labels). Annotated attack scenarios spanning all six 
taxonomic layers---LiDAR spoofing, FDIA, delay injection, replay, 
jamming, and privilege escalation---should be included, with 
precise temporal alignment to enable cross-layer correlation 
analysis. The ROSPaCe dataset~\cite{puccetti2024rospace} and 
ROSIDS23~\cite{rosids23} provide valuable ROS2 network traffic 
baselines, but neither captures the gait-phase-dependent traffic 
patterns unique to legged platforms. Upon release, baseline 
detection experiments should establish phase-agnostic 
vs.\ phase-conditioned IDS performance, quantifying the 
false-positive penalty of ignoring locomotion dynamics and 
providing a calibration target for Gap~1 algorithms. 
Complementary to the benchmark dataset, locomotion-state-aware 
feature engineering should develop IDS feature extractors that 
incorporate gait-phase indicators (swing/stance flags, duty 
factor, contact force thresholds) as conditioning variables, 
enabling the anomaly model to maintain separate or interpolated 
baselines for each locomotion regime. Existing network-based 
IDS frameworks for industrial and robotic 
systems~\cite{holdbrook2024ids, martin2025rips} provide 
architectural starting points, but require fundamental adaptation 
to accommodate the non-stationarity introduced by legged 
locomotion.

\subsubsection{Sub-200\,ms Communication Failover Validation 
\textnormal{(enables Gap~6)}}

Develop and validate a proactive multi-link transmission 
architecture (simultaneous WiFi + cellular with receiver-side 
selection and pre-established cryptographic sessions) targeting 
effective failover below 200\,ms under controlled RF interference. 
Critically, validation must be gait-phase-aware: measure fall 
probability and recovery-margin degradation during failover events 
initiated at swing-to-stance transitions versus mid-stance, on 
stairs versus flat terrain, establishing the first empirical link 
between communication-layer switching time and quadruped physical 
stability. Zhang et al.'s fall prediction and recovery 
analysis~\cite{zhang2024fall} provides the stability-margin 
framework against which communication failover timing must be 
benchmarked. This directly feeds Gap~6's stability-margin-driven 
QoS framework by providing the ground-truth timing constraints 
that adaptive protocols must satisfy. The jamming-aware path 
adaptation framework of Diller et al.~\cite{diller2023jamming} 
offers a complementary perspective, demonstrating that 
communication resilience can be integrated into robot motion 
planning rather than treated as a purely network-layer concern.

\subsubsection{FHSS Characterization Under Legged Locomotion 
\textnormal{(enables Gap~6)}}

Although FHSS is a mature anti-jamming primitive 
(TRL~8)~\cite{pirayesh2022jamming}, its interaction with quadruped 
teleoperation has not been characterized. Systematic experiments 
should delineate hopping-rate limits, resynchronization time 
distributions, and packet-loss bursts caused by motion-induced 
antenna masking, body-oscillation multipath, and terrain-dependent 
RF propagation in industrial environments. The deliverable is a 
set of locomotion-conditioned FHSS design guidelines that specify 
safe hopping rates and resynchronization budgets as a function of 
gait mode and terrain class.

\subsubsection{VR-to-Quadruped Security Testbed 
\textnormal{(enables Gaps~5, 1)}}

Construct a reproducible testbed spanning the full VR-to-quadruped 
teleoperation chain: HMD rendering and tracking, video compression 
and transport, command transmission, and robot control. The testbed 
must support controlled injection of operator-targeting attacks 
(latency jitter, optic-flow manipulation, depth-cue corruption, 
overlay injection, cybersickness-inducing 
stimuli~\cite{valluripally2021cybersickness, casey2021immersive}) 
while recording synchronized robot states, video timestamps, HMD 
pose streams, physiological indicators (eye tracking, head-motion 
dynamics), and operator command streams. The cybersickness testbed 
of De Pace et al.~\cite{csqvrtestbed2024} and the automatic 
detection pipeline of Yalcin et al.~\cite{yalcin2024cybersickness} 
provide methodological templates, but neither integrates physical 
robot teleoperation into the experimental loop. This 
infrastructure is the experimental foundation for Gap~5's 
operator-degradation detection research and provides labeled 
operator-state data for Gap~1's multi-rate IDS fusion 
architectures.

\subsubsection{Passivity-Based Control Under Adversarial 
Conditions \textnormal{(enables Gap~3)}}

Risiglione et al.'s TDPC 
framework~\cite{risiglione2021passivity} demonstrated delay 
tolerance up to 500\,ms during quadruped teleoperation, but was 
validated under benign (non-adversarial) delay conditions. 
Near-term experiments should systematically evaluate TDPC and 
wave-variable controllers~\cite{nuno2011passivity, 
haddadi2011passivity} under adversarial delay 
profiles---including delay injection with malicious jitter 
patterns, combined delay + FDIA~\cite{dong2020fdia}, and sudden 
communication loss during stair traversal---to establish the 
boundary between delay-tolerant and attack-vulnerable operating 
regimes. The passivity-shortage framework of Huang 
et al.~\cite{passivityshortage2020} and recent force-feedback 
bilateral teleoperation architectures~\cite{chen2024teleop, 
gao2024forcefeedback} offer alternative delay-compensation 
paradigms whose adversarial robustness remains untested. Results 
will directly inform Gap~3's recovery controller design by 
identifying which attack classes passivity alone can handle and 
where additional recovery primitives are required.

\subsection{Medium-Term (3--5 Years): Adapted Research}

Medium-term objectives require the development of novel algorithms 
and control architectures that are specifically designed for the 
hybrid, phase-dependent dynamics of legged locomotion. These 
directions build on near-term validation results and require 
closed-loop teleoperation experiments rather than isolated 
component evaluations.

\subsubsection{Phase-Conditioned Intrusion Detection Models 
\textnormal{(addresses Gap~1)}}

Using the benchmark dataset from near-term work, develop IDS 
architectures that maintain gait-phase-conditioned baselines, 
adapting anomaly thresholds to the robot's instantaneous 
locomotion state. Specific research questions include: 
(i)~whether phase-conditioned models reduce false-positive rates 
relative to phase-agnostic baselines during terrain transitions 
and gait switches, (ii)~optimal feature-fusion strategies for 
combining high-rate proprioceptive data (500--1000\,Hz) with 
medium-rate perception data (10--30\,Hz) and low-rate operator 
commands (1--10\,Hz), and (iii)~achievable detection latency as 
a function of the non-stationarity of the locomotion-induced 
traffic baseline. The ROBOCOP zero-shot 
framework~\cite{robocop2024iros} provides a starting point for 
handling previously unseen attack patterns, but requires 
extension to structured periodicity and contact discontinuities 
characteristic of legged systems. Network-based IDS 
approaches~\cite{holdbrook2024ids} and the ROS2-specific 
detection methodology of Guerrero-Higueras 
et al.~\cite{guerrero2018cyber} offer complementary detection 
paradigms whose integration with locomotion-state conditioning 
represents an open research challenge.

\subsubsection{Physics-Informed FDIA Detection for Legged 
Control Loops \textnormal{(addresses Gap~2)}}

Develop residual generators that incorporate rigid-body dynamics, 
contact models, and gait-phase constraints as structural priors 
for detecting false data injection attacks. The central research 
challenge is designing detectors that can identify 
Lie-group-class undetectable FDIAs~\cite{liegroup2024fdia, 
kwon2024fdia} by exploiting physical invariants that the attacker 
cannot simultaneously satisfy: contact-consistent force/torque 
relationships, gait-phase sequencing constraints, energy 
conservation bounds, and inter-leg coordination symmetry. The 
destabilization analysis of Dong et al.~\cite{dong2020fdia} 
established the foundational threat model for FDIA in bilateral 
teleoperation; extending this framework to legged systems 
requires incorporating the hybrid contact dynamics and 
gait-phase-dependent vulnerability windows that distinguish 
quadrupeds from the rigid-link manipulators considered in prior 
work. Complementary work should extend moving-target defense 
principles~\cite{kanellopoulos2020mtd, griffioen2021mtd} to 
inject low-amplitude, phase-aware watermarking signals into 
control commands that serve as authentication fingerprints 
without perturbing gait stability. Validation must include 
cross-modal consistency verification using proprioceptive and 
exteroceptive channels, leveraging the digital-twin anomaly 
detection paradigm~\cite{xu2021digitaltwin, 
arm2024digitaldouble} adapted to the non-smooth hybrid dynamics 
of legged contact. The curriculum-learning-based digital twin 
framework of Xu et al.~\cite{lattice2023acm} offers a promising 
methodology for progressively training anomaly detectors under 
increasingly sophisticated attack scenarios.

\subsubsection{Gait-Phase-Conditioned Recovery Controllers 
\textnormal{(addresses Gap~3)}}

Design and validate a library of phase-dependent recovery 
primitives that replace the unsafe default of emergency stop: 
controlled stance-widening from trot, three-point bracing from 
walk, sit-to-crouch from stair traversal, and controlled descent 
from elevated positions. Each primitive must be reachable from the 
corresponding gait phase within available stability margins and 
must not require operator input (since the operator channel may 
itself be compromised). The controller should select the 
appropriate primitive based on detected attack type, current gait 
phase, and estimated terrain geometry---drawing on the fall 
prediction and recovery framework of Zhang 
et al.~\cite{zhang2024fall} and the PID-Piper recovery 
methodology~\cite{dash2021pidpiper} which demonstrated autonomous 
recovery from physical attacks on robotic vehicles, albeit for 
wheeled rather than legged platforms. Formal safety analysis using 
Control Barrier Functions (CBFs) or reachability methods should 
establish guaranteed collision-avoidance and balance-maintenance 
envelopes under bounded adversarial perturbations, extending the 
ISS-based DoS tolerance framework of De Persis and 
Tesi~\cite{depersis2015dos} to the hybrid dynamics of legged 
contact and multi-channel teleoperation attack models. The 
targeted attack detection framework of Alemzadeh 
et al.~\cite{alemzadeh2016targeted}, originally developed for 
teleoperated surgical robots, provides a methodological template 
for dynamic model-based detection and mitigation that warrants 
adaptation to legged locomotion dynamics.
\textbf{Success Criteria for Gap 4:} Effective perception attack detection for quadrupeds must achieve: (i) detection latency below 100\,ms for point-cloud manipulation attacks, corresponding to approximately one-fifth of a typical gait cycle and enabling foothold replanning before commitment; (ii) false positive rate below 0.5\% during terrain transitions (flat$\rightarrow$stairs, hard$\rightarrow$compliant), which naturally produce high variance in perception signals; (iii) robustness to platform oscillation up to $\pm15^\circ$ pitch/roll without requiring explicit motion compensation as a preprocessing step; (iv) maintained detection performance under degraded visibility conditions (dust, fog, low light) that authentically reduce point-cloud density; and (v) computational overhead compatible with real-time perception pipelines, specifically below 20\,ms additional latency on embedded GPU platforms.

\subsubsection{Motion-Compensated Perception Attack Detection 
\textnormal{(addresses Gap~4)}}

Adapt LiDAR spoofing detection 
(CARLO~\cite{sun2020carlo}, Shadow-Catcher~\cite{cao2023pra}) 
and VO integrity 
monitoring~\cite{xu2025voattack, zhang2025voattack} for 
quadruped-specific operating conditions. This requires: 
(i)~explicit compensation for body oscillation 
($\pm 5$--$15^\circ$ pitch/roll during trot) and foot-impact 
vibration in point-cloud statistics and shadow/occlusion models, 
(ii)~terrain-context-aware detection thresholds that exploit 
geometric regularities of known terrain classes (stair edges, 
riser planes, pipe geometries), and (iii)~foothold-level 
integrity verification that detects perception corruption at the 
decision level---comparing planned footholds with proprioceptive 
feedback at contact (expected vs.\ actual contact timing, force 
magnitude, surface normal) to identify systematic discrepancies 
that signal terrain-model corruption. The next-generation LiDAR 
spoofing capabilities documented by Sato 
et al.~\cite{sato2024nextgenlidar} and the camera-LiDAR fusion 
vulnerabilities exposed by Hallyburton 
et al.~\cite{hallyburton2022cameralidar} underscore that 
perception attack detection cannot rely on any single sensor 
modality. Gait-mode-specific reprojection-error baselines and 
false-alarm budgets must be established for visual odometry 
protection under legged locomotion dynamics. The adversarial 
attacks on learning-based quadrupedal locomotion 
controllers~\cite{shi2024adversarial} demonstrate that 
perception-layer corruption can directly destabilize gait 
controllers, reinforcing the urgency of domain-specific detection 
mechanisms.

\subsubsection{Operator-State Monitoring and Graduated Autonomy 
Transfer \textnormal{(addresses Gap~5)}}

Transition cybersickness and operator-impairment detection from 
seated VR laboratory 
experiments~\cite{biswas2024cybersicknessreview, 
kourtesis2024csqvr} to active quadruped teleoperation, 
establishing quantitative correlations between physiological 
indicators (head-motion entropy, gaze 
stability~\cite{yalcin2024cybersickness}) and 
control-performance degradation metrics (reaction time, command 
smoothness, stability-margin violations). Develop graduated 
autonomy transfer protocols that implement a continuous spectrum 
of human--robot authority sharing---from full teleoperation 
through assisted modes (velocity limiting, obstacle avoidance 
overlays, gait-selection assistance) to full local 
autonomy---with transfer policies that are gait-phase-aware and 
resistant to adversarial spoofing of the transfer 
trigger~\cite{kundu2025cybersicknessattack}. The latency-induced 
cybersickness mechanisms documented by Stauffert 
et al.~\cite{stauffert2020latency} and the attack surface 
analysis of Casey et al.~\cite{casey2021immersive} establish that 
adversaries can deliberately induce operator impairment through 
targeted latency manipulation. The architecture must incorporate 
cross-layer correlation: simultaneous detection of operator 
impairment and communication anomalies should elevate the 
suspicion of coordinated attack rather than organic degradation.

\subsubsection{Locomotion-State-Driven Communication Protocol 
Adaptation \textnormal{(addresses Gap~6)}}

Develop the formal mapping from instantaneous stability margin 
(support polygon margin, divergent component of motion, 
capturability) to communication QoS parameters (maximum tolerable 
latency, packet-loss budget, required redundancy level). Implement 
and validate phase-aware anti-jamming coordination: during 
swing-to-stance transitions, hopping schedules should prioritize 
reliability; during stable stance, they can afford aggressive 
frequency diversity. The comprehensive jamming taxonomy of 
Pirayesh and Zeng~\cite{pirayesh2022jamming} provides the 
adversarial model against which adaptive protocols must be 
validated. Predictive multi-link management should anticipate 
communication challenges from locomotion planning (e.g., 
pre-activating secondary links before stairwell entry) rather 
than reacting to detected degradation. Benchmark NIST lightweight 
cryptography standard 
ASCON~\cite{radhakrishnan2024lightweight} and emerging 
post-quantum candidates~\cite{nistpqc2024, pqc2025embedded} 
specifically for quadruped control-loop traffic profiles (small 
payloads at high rates) under concurrent locomotion-control CPU 
load. The encrypted four-channel bilateral control framework of 
Takanashi et al.~\cite{takanashi2023encrypted} demonstrates that 
cryptographic overhead can be accommodated within bilateral 
teleoperation loops, but its interaction with gait-phase-dependent 
timing constraints remains unexplored.

\subsubsection{LLM Safety Verification for Physical Robot Control}
\label{sec:llm_safety}

\textit{Scope Note:} This direction addresses an emerging threat vector that crosscuts multiple gaps (Gaps~1, 3, and 4) but warrants dedicated treatment due to the rapid proliferation of LLM-based robot interfaces. While not assigned a dedicated gap number, the research challenges are substantive and increasingly urgent.

As natural language and foundation-model interfaces proliferate 
in robot teleoperation~\cite{cheng2024opentv}, developing 
semantic command verification and adversarial prompt defenses 
specific to physical robot control constitutes an emerging 
research need that cuts across multiple gaps but does not map 
cleanly to a single numbered deficiency. The RoboPAIR jailbreak 
achieving 100\% success on Unitree 
Go2~\cite{robson2024robopair} demonstrates that policy-only 
guardrails are insufficient; effective defenses require 
multi-layered checks combining intent validation, 
physical-constraint enforcement (joint limits, stability 
envelopes, proximity to humans), and action-level runtime 
monitoring that can reject commands regardless of their 
linguistic framing. This direction is positioned in the 
medium-term horizon because it requires both the detection 
infrastructure from Gap~1 and the physical-constraint models 
from Gap~3 to implement effective multi-layered verification. 
The broader secure robotics framework articulated by Pandit 
et al.~\cite{secureroboticsacm2025} situates LLM safety within 
the nexus of safety, trust, and cybersecurity for cyber-physical 
systems, reinforcing the need for holistic rather than 
component-level defenses.

\subsection{Long-Term (5--10 Years): Fundamental Research}

Long-term research trajectories address foundational challenges 
whose solutions will outlast current threat models and platform 
generations. These directions establish formal assurance 
frameworks, adaptive architectures, and systemic infrastructure 
that provide principled security guarantees rather than empirical 
patch-based defenses.

\subsubsection{Cross-Layer Anomaly Correlation and Coordinated 
Attack Detection \textnormal{(addresses Gap~7)}}

Building on validated single-layer detectors from Gaps~1--4, 
construct causal or graphical models (Bayesian networks, 
structural causal models, factor graphs) that represent expected 
statistical dependencies between perception outputs, 
communication metrics, operator behavior, and control-loop 
states. Under normal operation, these dependencies follow 
predictable patterns; coordinated attacks introduce anomalous 
conditional dependencies (e.g., perception degradation correlated 
with communication latency spikes correlated with control 
residual growth) that can be detected as structural deviations. 
Phase-conditioned coincidence testing should exploit the temporal 
signature of gait-phase-synchronized coordinated 
attacks---clustered weak anomalies at phase boundaries that 
random independent faults would not produce. Cross-layer 
digital-twin integration~\cite{xu2021digitaltwin, 
erceylan2025dtcps} should extend from single-domain consistency 
to full-pipeline prediction of expected perception, 
communication, operator, and control states, building on the 
physics-informed hybrid autoencoder framework of Wang 
et al.~\cite{wang2024dtrobot} and the unsupervised anomaly 
detection methodology of Hnaien 
et al.~\cite{hnaien2025dtrobot}. Validation requires 
standardized multi-layer red-team attack scenarios executed on 
physical quadruped platforms.

\textit{Dependency note:} This direction is rated as contingent 
because meaningful cross-layer correlation explicitly depends on 
progress in Gaps~1--4; without functional single-layer detectors 
as prerequisites, correlation-based detection lacks the input 
signals necessary for joint statistical testing. Accordingly, 
this direction is best pursued as an integration effort once 
foundational detection capabilities mature.

\subsubsection{Formal Verification of Teleoperation Security 
Properties}

Construct compositional formal models of the teleoperation 
pipeline---operator interface, transport layer, middleware, 
control architecture, and hybrid locomotion 
dynamics---and verify security-relevant properties (bounded 
damage under constrained adversary power, guaranteed safe-state 
reachability under attack, liveness of recovery protocols) 
through model checking and theorem proving. The hybrid automata 
framework is natural for legged systems with contact-switching 
dynamics, but the state-space explosion from combining cyber 
(protocol states, authentication states, IDS states) with 
physical (continuous dynamics, discrete contact modes) variables 
poses a fundamental scalability challenge. Compositional 
verification approaches that decompose the problem along trust 
boundaries (operator zone, communication zone, robot zone) while 
preserving cross-boundary invariants offer the most tractable 
path. The formal analysis methodology of Yang 
et al.~\cite{yang2024ros2formal} for ROS2 communication security 
and the runtime verification paradigm established by Celik 
et al.~\cite{celik2023rv} provide partial building blocks, but 
neither addresses the hybrid cyber-physical state space inherent 
in legged teleoperation.

\subsubsection{Locomotion-Aware Adaptive Security Policies}

Formulate security policies that are explicitly parameterized by 
physical context: terrain classification, gait phase, proximity 
to humans, stability margin, and mission criticality. This 
requires deep integration of security monitoring into locomotion 
planning and control, enabling defenses to intensify during 
critical phases (e.g., stair ascent near personnel) and relax 
during low-consequence phases to preserve performance. The 
theoretical challenge is ensuring that the security policy itself 
does not introduce destabilizing discontinuities---e.g., abruptly 
increasing encryption overhead at a gait-phase boundary must not 
create the very latency spike it aims to prevent. This direction 
synthesizes insights from all eight gaps into a unified, 
context-dependent security architecture.

\subsubsection{Moving Target Defense for Legged Hybrid Dynamics}

Implement CPS moving-target defense---stochastic switching of 
control parameters, entropy-maximizing credential rotation, 
randomized communication scheduling---in legged control systems 
while preserving the stability of hybrid limit-cycle dynamics. 
The foundational MTD frameworks of Kanellopoulos and 
Vamvoudakis~\cite{kanellopoulos2020mtd}, Griffioen 
et al.~\cite{griffioen2021mtd}, and Giraldo 
et al.~\cite{giraldo2019mtd} establish the control-theoretic 
basis for CPS moving-target defense, while the IoT-specific 
adaptation of El-Mahdi et al.~\cite{elmahdi2022mtdiot} 
demonstrates feasibility under resource constraints. The 
fundamental theoretical challenge for legged systems is proving 
that defensive variability does not disrupt the orbital stability 
of periodic gaits: perturbations must be bounded within the basin 
of attraction of each gait limit cycle, requiring tools from 
hybrid dynamical systems theory and contraction analysis. 
Phase-aware MTD that restricts switching to low-sensitivity gait 
phases offers a pragmatic intermediate approach, but formal 
guarantees require advances in the intersection of hybrid control 
theory and adversarial robustness.

\subsubsection{Post-Quantum Cryptography for Real-Time Robotic 
Control}

As NIST PQC standards 
(FIPS~203/204/205)~\cite{nistpqc2024} mature and quantum 
computing capabilities advance, benchmark PQC algorithms on 
robotic embedded processors representative of quadruped onboard 
computing (ARM Cortex-A class, NVIDIA Jetson) and formulate 
hybrid classical/PQC transition protocols for ROS2/DDS security 
frameworks. Timeline: 5--10 years, aligned with NIST IR~8547 
migration objectives~\cite{nistir8547}. Early benchmarking 
highlights significant computational challenges for 
resource-constrained platforms~\cite{pqc2025embedded}; 
quadruped-specific evaluation must assess not only throughput 
and mean latency but worst-case execution time under concurrent 
locomotion-control load, ensuring that PQC overhead does not 
violate real-time control deadlines during stability-critical 
gait phases. The documented DDS security 
vulnerabilities~\cite{deng2022insecurity, aliasrobotics2022dds} 
underscore that the cryptographic transition must address not 
only algorithm substitution but the broader middleware security 
architecture.

\subsubsection{Privacy-Preserving Distributed IDS via Federated 
Learning}

Enable fleet-wide intrusion detection model improvement without 
centralizing sensitive operational data, using federated learning 
to share model updates rather than raw telemetry across robot 
fleets. This direction is explicitly dependent on the 
availability of quadruped-specific IDS datasets (Gap~1) and 
validated baseline detectors. The wormable nature of the UniPwn 
BLE exploit~\cite{makris2025unipwn} and the fleet-wide 
implications of hardcoded cryptographic 
keys~\cite{unitreego1backdoor, unipwncve2025} demonstrate that 
fleet-level threat propagation is not theoretical but 
empirically documented, reinforcing the urgency of distributed 
detection capabilities. Research challenges include: robust 
aggregation under adversarial participants (Byzantine-resilient 
federated protocols), effective learning under non-IID data 
distributions (each robot encounters distinct terrains and threat 
environments), communication efficiency during training under 
constrained and intermittent connectivity, and model-update 
privacy (preventing gradient-based inference of operational 
patterns). A realistic maturity horizon is 7--10 years, 
contingent on the resolution of foundational Gaps~1--2.

\subsubsection{Zero-Trust Architecture and Runtime Verification 
for ROS2 Quadruped Deployments}

Develop and validate a comprehensive zero-trust framework for 
ROS2/DDS-based quadruped systems: per-node authentication with 
ephemeral credentials, continuous re-authorization, 
least-privilege topic permissions, and micro-segmentation 
isolating locomotion-critical control nodes from high-risk 
surfaces (perception ingestion, VR/UI clients, cloud relays). 
The NIST ZTA reference architecture~\cite{nist2020zt} and its 
robotics-specific adaptation by Mayoral-Vilches 
et al.~\cite{aliasrobotics2020zt} provide the conceptual 
foundation, while the SROS2 security 
framework~\cite{mayoral2022sros2} and the supply chain 
exploitation analysis of Severov 
et al.~\cite{sros2supplychain2025} delineate both the available 
tooling and its current limitations. Simultaneously, implement 
runtime verification monitors encoding quadruped safety 
invariants---phase-consistent contact sequences, joint 
torque/velocity envelopes, stability-margin bounds, and 
command-validity checks---using stream-based monitoring 
frameworks (e.g., RTLola~\cite{baumeister2025rtlola}) with 
verified sub-millisecond overhead. The ROSRV 
framework~\cite{huang2014rosrv} and its recent field-testing 
extensions~\cite{santos2024rvros} demonstrate that runtime 
verification can be integrated into ROS-based systems, but 
achieving the required temporal guarantees on embedded-class 
processors under concurrent locomotion-control load remains an 
open engineering challenge. The long-term goal is a formally 
verified security architecture where every node, message, and 
state transition is continuously authenticated and monitored 
against physically grounded safety specifications. The primary 
engineering challenge is achieving this without violating 
sub-10\,ms control-loop timing budgets on embedded-class 
processors---a constraint that current ZTA implementations for 
enterprise IT do not face.

\section{Note to Practitioners}
\label{sec:practitioners}

This section translates the survey findings into practical guidance for organizations deploying teleoperated quadruped robots. The recommendations below combine established cybersecurity best practices with quadruped-specific adaptations inferred from the consequence and maturity analyses presented in this survey. To support staged adoption, the guidance is organized by implementation timeline and organizational capability.

The platform-specific examples provided here reflect publicly documented vulnerabilities and vendor guidance available within the review window of this survey. Practitioners should validate all platform-specific controls against current firmware versions, software releases, and vendor advisories prior to deployment.

\subsection{Immediate Actions (Weeks 1--4)}

As an initial risk-reduction measure, organizations should harden platform-specific entry points that have been implicated in publicly documented attack chains. For Unitree platforms, Bluetooth Low Energy (BLE) provisioning services should be disabled when not actively used for WiFi credential configuration, since BLE access has been identified as an initial entry point in the UniPwn attack chain~\cite{makris2025unipwn}. For Ghost Robotics Vision 60 deployments, the MAVLink control channel should be isolated on a dedicated network segment, with explicit firewall rules restricting unauthorized tablet connections~\cite{incibe2025vision60}; where operationally feasible, VPN encapsulation can provide an interim mitigation until authenticated protocol variants are available. For Boston Dynamics Spot deployments, operators should verify that TLS certificate validation is enforced on all API connections and that default credentials have been rotated in accordance with vendor security guidance~\cite{boston2024security}.

Across all platforms, practitioners should perform an inventory of exposed network services (e.g., using \texttt{nmap} or equivalent enterprise scanning tools) and disable non-essential services, particularly legacy ROS1 endpoints, developer/debug interfaces, and unused remote access services. As a general hardening principle, a quadruped deployment should expose only explicitly justified network services, with each enabled service documented in the deployment security plan. Common unnecessary exposures include: SSH (disable if physical access suffices; if required, enforce key-based authentication only, disable password login, and restrict access via IP allowlisting or VPN), HTTP/HTTPS debug interfaces (ports 80/443/8080), and multicast DDS/RTPS discovery traffic (port ranges are vendor- and configuration-dependent) when operating on shared networks. Wireless networks supporting quadruped teleoperation should use WPA3 where supported by hardware and firmware, with WPA2-Enterprise as a minimum baseline for managed deployments. Under the threat model considered in this survey, open networks and WPA2-Personal networks using shared credentials are generally inadequate for safety-critical operation.

\subsection{Short-Term Hardening (Months 1--3)}

Communication channels should be upgraded to DTLS~1.2 or TLS~1.3, with session resumption enabled where available to reduce reconnection latency during link transitions. In addition, organizations should implement redundant communication paths (e.g., primary WiFi with cellular backup) and validate failover performance under controlled RF interference and congestion scenarios. As a practical target, failover behavior should be tested against sub-200\,ms recovery objectives where mission dynamics require it; however, acceptable thresholds should ultimately be derived from the platform's locomotion and stability constraints rather than from generic networking assumptions alone.

For ROS2-based deployments, DDS Security should be enabled with per-participant certificates and topic-level access control policies~\cite{deng2022insecurity, mayoral2022sros2}. Although configuration complexity can be significant, operating production robots with effectively unauthenticated and unencrypted middleware introduces a substantial and avoidable attack surface. Organizations lacking in-house PKI capability should consider external support (e.g., consultants or managed PKI services) rather than deferring DDS Security deployment indefinitely.

Operator training programs should explicitly include cybersecurity and cyber-physical incident awareness. Training should cover recognition of anomalous robot behavior that may indicate compromise, controlled mission-abort procedures when communication integrity is uncertain, and escalation paths for suspected security incidents. For VR/AR-based teleoperation, operators should also be briefed on cybersickness symptoms and encouraged to terminate sessions when disorientation occurs, without penalty or mission-failure stigma.

\subsection{Medium-Term Architecture (Months 3--12)}

Organizations should implement network segmentation that isolates quadruped control traffic from general enterprise IT networks, with explicit firewall policies and monitoring controls at segment boundaries. Intrusion detection systems (IDS) can improve visibility into abnormal network behavior, but practitioners should treat early deployments of machine learning-based IDS as advisory rather than authoritative, especially because quadruped-specific traffic baselines and labeled attack datasets remain immature. Elevated false-positive rates should be expected during initial deployment and managed through iterative threshold tuning and operator feedback loops.

To support both operational monitoring and future security improvements, organizations should establish baseline communication profiles that characterize normal traffic patterns across gait modes, terrain classes, and teleoperation modalities (e.g., joystick, tablet, VR/AR). At a minimum, baseline logging should capture timing and jitter characteristics, command rates, communication path transitions, and contextual metadata (e.g., mission mode, terrain type, and operator interface). These baselines can support anomaly detection even in the absence of quadruped-specific IDS models and can provide high-value data for internal validation or external research collaboration.

Supply-chain risk management should extend to firmware and software update processes. Where supported, firmware authenticity should be verified using cryptographic signature validation prior to deployment. Organizations should also maintain offline archives of known-good firmware and software packages to enable rollback in the event of compromise or unstable updates. Cloud-connected features (e.g., telemetry relay or fleet services) should be disabled unless operationally required; when enabled, outbound traffic should be explicitly monitored, documented, and reviewed to characterize telemetry flows and identify unexpected data exfiltration paths.

\subsection{Incident Response and Safety-Security Coordination}

Incident response procedures for quadruped systems should be designed as cyber-physical procedures, not merely IT procedures. When compromise is suspected, the immediate goal is to reduce risk while preserving locomotion safety. In many scenarios, abrupt communication termination or indiscriminate emergency-stop actions may increase fall risk, particularly during gait transitions, stair traversal, or unstable terrain contact. Response plans should therefore define controlled degraded modes (e.g., reduced-speed stance stabilization, supervised retreat, or safe posture transition) that can be executed before full isolation when conditions permit.

Organizations should predefine a minimum incident response playbook covering: (i) isolation of the affected network segment or communication path, (ii) preservation of logs and synchronized timestamps for forensic review, (iii) operator actions for transition to a safe or degraded mode, and (iv) escalation to engineering and security personnel for post-incident analysis. These procedures should be rehearsed periodically using tabletop exercises and controlled field tests, with explicit attention to the interaction between cybersecurity actions and locomotion safety constraints.

\subsection{Risk Acceptance Documentation}

For attack classes where mature mitigations are not yet available---particularly perception-layer attacks (approximately TRL~4--5) and operator-targeting threats (approximately TRL~3--4) in the taxonomy used in this survey---organizations should formally document residual risk acceptance decisions. Such documentation should include: (i) the relevant attack class and assessed consequence severity, (ii) available partial mitigations and known limitations, (iii) operational constraints used to reduce exposure (e.g., supervised operation, constrained environments, mission restrictions), and (iv) monitoring indicators or trigger conditions that will prompt reassessment.

Risk acceptance records should be reviewed not only after incidents, but also after major firmware/software updates, changes in mission profile, deployment to new environments, or periodic governance reviews (e.g., quarterly or per campaign/mission cycle). This process supports both governance objectives (demonstrating due diligence to leadership, auditors, and regulators) and operational objectives (ensuring field personnel understand which threats remain incompletely mitigated and what compensating procedures are expected).

Risk acceptance is not risk denial. It is the explicit recognition that some threats cannot yet be fully mitigated with available technology and must therefore be managed through a combination of technical controls, operational constraints, and informed human oversight.


\section{Conclusion}
\label{sec:conclusion}

This survey has presented a systematic analysis of cybersecurity 
threats to teleoperated quadruped robots, addressing security 
challenges arising from dynamic stability requirements, significant 
kinetic energy, gait-dependent vulnerability windows, and elevated 
operator cognitive load---characteristics that fundamentally 
distinguish quadrupeds from wheeled, aerial, and industrial 
robotic platforms.

\subsection{Principal Findings}

\textbf{Finding 1: Defense Maturity Asymmetry.} The Technology 
Readiness Level analysis reveals a pronounced maturity gradient 
across the six-layer attack taxonomy. Communication-layer defenses 
(TLS/DTLS, authentication, FHSS) have achieved field-deployment 
maturity (TRL 7--9), while perception-layer protections remain at 
TRL 4--5 with validation limited to autonomous vehicle contexts. 
VR/AR operator defenses are confined to TRL 3--4, representing the 
most significant coverage deficit given the tight coupling between 
operator perception and control decisions.

\textbf{Finding 2: Platform Security Heterogeneity.} Commercial 
quadruped platforms exhibit dramatic security disparities. Boston 
Dynamics Spot demonstrates industry-leading practices with per-device 
cryptographic keying and firmware integrity verification. Conversely, 
Unitree platforms have exhibited fleet-wide cryptographic weaknesses, 
while Ghost Robotics Vision 60---despite military deployment 
contexts---was disclosed to operate an unauthenticated control 
channel. This heterogeneity suggests that security posture is 
currently determined by vendor engineering decisions rather than 
industry standards or regulatory requirements.

\textbf{Finding 3: Cascading Failure Dynamics.} Attack-to-consequence 
analysis demonstrates that perception and control attacks can 
propagate to physical instability within sub-second timescales 
during dynamic locomotion phases, leaving minimal opportunity for 
operator intervention. Effective defense therefore requires automated 
detection and response mechanisms rather than operator-mediated 
incident handling.

\subsection{Key Quantitative Findings}

The analysis yields several quantifiable insights: (i) the defense maturity gap spans 2--4 TRL levels between communication-layer protections (TRL 7--9) and perception/operator-layer defenses (TRL 3--5); (ii) commercial platform security scores range from 2.2 (Unitree, Ghost Robotics) to 8.8 (Boston Dynamics Spot) on a 10-point composite scale, with sensitivity analysis revealing that alternative weighting schemes shift scores by up to $\pm$1.6 points; (iii) cascading failure pathways can propagate from initial compromise to physical instability within 100--500\,ms during dynamic locomotion phases; (iv) the aggregate security readiness index across all six attack layers is 0.88, indicating below-threshold maturity when the full attack surface is considered; and (v) eight prioritized research gaps require estimated 1--10 years to address, with near-term engineering validation (1--3 years) offering the highest return on research investment.

\subsection{Research Priorities}

The eight research gaps identified---gait-phase-aware intrusion detection (Gap~1), real-time FDIA detection exploiting locomotion physics (Gap~2), stability-preserving recovery control (Gap~3), motion-compensated perception attack detection (Gap~4), operator degradation monitoring with graduated autonomy transfer (Gap~5), locomotion-aware communication protocols (Gap~6), cross-layer anomaly correlation (Gap~7), and fleet-level security with wormable exploit containment (Gap~8)---represent the critical path toward closing the security deficit between accelerating deployment and infrastructure maturity. Gaps~1--3 are rated highest priority due to their Critical consequence severity and addressable maturity deficits; Gap~8 addresses an emerging threat vector with demonstrated real-world exploitation.

\subsection{Limitations and Future Work}

This survey is subject to several methodological constraints. Security 
assessments for ANYmal and DeepRobotics X30 are limited by proprietary 
documentation practices; the absence of disclosed vulnerabilities may 
reflect either robust engineering or restricted auditing access. 
Quantitative attack success rates are derived from original studies 
under controlled experimental conditions and require domain-specific 
validation before operational deployment assumptions. The field is 
evolving rapidly; specific taxonomy entries and maturity classifications 
will require revision as new threats emerge and defenses mature.

Future work should prioritize: (i) development of quadruped-specific 
IDS benchmark datasets capturing gait-phase-dependent traffic patterns, 
(ii) experimental validation of cross-domain defenses under legged 
locomotion dynamics, and (iii) establishment of industry security 
standards specific to teleoperated legged platforms.

\section*{Ethical Considerations}

This survey synthesizes publicly disclosed vulnerability information and peer-reviewed security research. No novel vulnerability discovery, exploitation, or penetration testing was conducted as part of this work. All cited CVEs were responsibly disclosed through established channels prior to our analysis. Platform-specific vulnerability details are presented at the level of abstraction necessary for defensive planning without providing exploitation recipes.

\textbf{Dual-Use Concerns:} We acknowledge that security research inherently carries dual-use potential: the same knowledge that enables defenders to protect systems can inform attackers seeking to exploit them. We have deliberately omitted implementation-level details (e.g., specific exploit code, exact protocol manipulation sequences, or step-by-step attack tutorials) while providing sufficient technical depth for practitioners to understand threat mechanisms and prioritize defenses. The attack taxonomy and consequence analysis are intended to motivate defensive investment, not to serve as an offensive playbook.

\textbf{Military and Law Enforcement Applications:} Quadruped robots are increasingly deployed in military reconnaissance, explosive ordnance disposal, and law enforcement contexts where security failures carry elevated consequences. We recognize that hardening these platforms may have implications for both defensive operations (protecting personnel) and offensive capabilities (enabling contested-environment deployment). This survey takes a platform-agnostic defensive perspective; we do not advocate for or against specific military applications, but note that organizations deploying quadrupeds in adversarial contexts should assume sophisticated threat actors and plan accordingly.

\textbf{Recommendations for Future Researchers:} Investigators building on this work should: (i) coordinate with affected vendors through responsible disclosure channels before publishing novel vulnerability findings, (ii) consider embargo periods that allow patch deployment before public disclosure, (iii) avoid publishing working exploit code except where necessary for reproducibility of defensive research, and (iv) clearly distinguish between demonstrated vulnerabilities and theoretical attack vectors. The goal of security research should be to raise the cost of attack while minimizing the risk that published findings enable harm.

\section*{Acknowledgments}
During the preparation of this manuscript, we used AI-assisted language tools (Claude and QuillBot) for grammar correction, language editing, and improving readability of the text. 


\balance

\end{document}